\documentclass[11pt]{article}
\usepackage[preprint]{acl}
\usepackage{times}
\usepackage{latexsym}
\usepackage[T1]{fontenc}
\usepackage[utf8]{inputenc}
\usepackage{microtype}
\usepackage{amsmath,amssymb}
\usepackage{algorithm}
\usepackage{algpseudocode}
\usepackage{graphicx}
\usepackage{placeins}
\usepackage{stfloats}
\usepackage{needspace}
\usepackage{booktabs,longtable,array}
\usepackage{xcolor}
\usepackage{listings}
\usepackage{colortbl}
\newcommand{\mask}{\textsc{mask}}

\title{ReMaRL: Learning to Unmask and Remask\\
in Frozen Diffusion Language Models}
\hypersetup{pdftitle={ReMaRL: Learning to Unmask and Remask in Frozen Diffusion Language Models}}
\definecolor{remarllink}{HTML}{157CA8} \hypersetup{colorlinks=true,allcolors=remarllink}
\author{
  Egor Shibaev$^{1,3}$ \quad
  Danila Biktimirov$^{2,3}$ \quad
  Ivan Klimov$^{1,3}$ \quad
  Artem Makoian$^{1,3}$ \\
  \bfseries 
  Viacheslav Meshchaninov$^{3,*}$ \quad
  Dmitry Vetrov$^{3,*}$ \\[4pt]
  \normalfont\normalsize
  $^{1}$JetBrains Research \quad $^{2}$JetBrains \quad
  $^{3}$Constructor University
}
\begin{document}
\raggedbottom
\maketitle
\begingroup
\renewcommand{\thefootnote}{\fnsymbol{footnote}}
\footnotetext[1]{Equal senior contribution.}
\endgroup
\begin{abstract}
Masked diffusion language models generate tokens in parallel, and the order in which tokens are unmasked affects final-answer accuracy.
Several methods learn unmasking policies from task rewards, but these policies cannot remask earlier predictions.
Remasking lets the model correct those predictions using newly generated context.
Heuristic rules guide both unmasking and remasking, while supervised remaskers learn to detect token errors.
Neither approach directly optimizes these decisions for final-answer correctness.
To address this gap, we introduce ReMaRL (Remasking with Reinforcement Learning), a lightweight policy that learns which tokens to unmask, retain, or remask.
We train this roughly 0.3M-parameter policy with Group Relative Policy Optimization (GRPO) from final-answer rewards while keeping the diffusion model frozen.
With LLaDA-8B-Instruct, ReMaRL achieves the highest scores among the compared unmasking and remasking policies within the tested inference budgets of 4--32 model calls on math and 8--16 calls on code.\footnote{%
\href{https://github.com/EgorShibaev/remarl}{Code}
and
\href{https://drive.google.com/drive/u/0/folders/1UFagh30vsvA-3jK-3sAf_iREaiQ3mxki}{checkpoints}
are publicly available.}
\end{abstract}
\section{Introduction}

Masked diffusion language models generate text by predicting tokens
in parallel and choosing which positions to unmask at each
step~\citep{mdlm,llada}. These choices affect generation quality,
especially when only a few model calls are available and many
positions must be filled at once.

Several methods use heuristics to choose which tokens to unmask or, when the method allows it, remask~\citep{llada,p2,remdm}.
Learning unmasking policies with reinforcement learning can improve over these rules while keeping the diffusion model frozen~\citep{dcolt,apple,upo,sas}.
These policies choose which masked positions to fill, but leave tokens unchanged once unmasked.

When several positions are unmasked in parallel, each token is predicted from the same context without conditioning on the other
tokens being unmasked. If those tokens depend on one another, this
factorization can produce predictions that are individually plausible
but inconsistent together~\citep{fastdllm}. Remasking one of them
lets the model predict it again with the others in context. Learned remasking modules, including G-Star~\citep{gstar} and BackPlay~\citep{backplay}, use token-level supervision derived from reference text to identify tokens to remask. These modules learn to detect token errors as a proxy for improving generation quality, without directly optimizing final-answer correctness.

To address this gap, we propose ReMaRL (Remasking with
Reinforcement Learning), a lightweight policy that learns
which positions to unmask and which earlier tokens to remask.
We build it on P2$'$, our stochastic adaptation of
P2~\citep{p2}, by learning a correction to its
confidence-based position scores.
We train this correction with Group Relative Policy
Optimization (GRPO; \citealp{grpo}) using final-answer
rewards, while keeping the diffusion model frozen. At each non-final step, a schedule sets the total number of positions that will be unmasked after the update, and ReMaRL chooses those positions. Following the P2 update rule~\citep{p2}, selected
masked positions receive the model's predictions, selected unmasked
positions keep their tokens, and unmasked positions outside the set
are remasked.

Our contributions are as follows:
\begin{itemize}
\item We introduce ReMaRL, a lightweight policy that learns
which positions to unmask, retain, or remask from final-answer
rewards while keeping the diffusion model frozen.

\item We show that, with frozen
LLaDA-8B-Instruct~\citep{llada}, ReMaRL achieves the highest
scores in our comparison within the tested inference budgets
of 4--32 calls on math and 8--16 calls on code.

\item We also develop P2$'$, the confidence-based remasking
rule underlying ReMaRL. Tuning its position temperature yields a strong decoding method without policy training:
at a budget of 16 model calls, it ranks second only to
ReMaRL on all four benchmarks among the compared methods.

    \item We conduct analyses and ablations showing that the
    policy's components contribute to accuracy and that trained
    policies can benefit from more calls at inference. We also
    demonstrate data efficiency and applicability to a second
    frozen diffusion model, Dream.
\end{itemize}
\section{Background}
\label{sec:background}

\subsection{Masked diffusion decoding}

Masked diffusion language models learn to predict missing tokens
from partially masked sequences~\citep{mdlm,llada}. To generate an
answer, decoding starts with a sequence of mask tokens. Each model call predicts tokens for the currently masked positions. A decoding rule determines how many positions to unmask and selects them. The number can follow a fixed schedule or depend on the model's predictions~\citep{llada,eb-sampler,fastdllm}. For example, confidence-based decoding fills a prescribed number of positions whose proposed tokens have the highest model probabilities~\citep{llada}.

Let $y$ denote the prompt and $L$ the answer length. We write
$x^{t-1}$ for the partially generated answer before model call $t$.
Each position contains either a token from vocabulary $\mathcal V$ or the mask symbol $\mask$:
\[
 x^{t-1}\in(\mathcal V\cup\{\mask\})^L,
 \qquad x^0=\mask^L.
\]
Given the prompt and current answer, the model returns a token
distribution $p_{\theta,i}(\cdot\mid y,x^{t-1})$ at each position $i$.
The greedy token proposal is
\begin{equation}
 \hat{x}_i^t=\arg\max_{v\in\mathcal V}
 p_{\theta,i}(v\mid y,x^{t-1}).
 \label{eq:proposal}
\end{equation}
The decoding rule uses these predictions to update the answer from
$x^{t-1}$ to $x^t$. We denote the total number of model calls by $T$,
also called the number of function evaluations (NFE).

Full-diffusion decoding can unmask positions across the entire
answer at each call. Semi-autoregressive decoding processes blocks from left to right,
unmasking tokens only within the current block~\citep{llada}.

\subsection{P2 decoding}

In standard masked diffusion decoding, each token remains unchanged
once it is unmasked. Remasking allows the model to revise earlier
predictions by masking selected tokens and predicting them
again~\citep{maskpredict,p2,remdm,remedi,gstar}.

P2~\citep{p2} uses a planner to choose which positions to unmask or
remask. It considers self-planning, external, and learned planners.
Here we describe its greedy self-planner with equal weighting of
masked and unmasked positions. It scores each position using the probability of its freshly proposed token, including positions that already contain a token:
\begin{equation}
 s_i^t=p_{\theta,i}(\hat{x}_i^t\mid y,x^{t-1}).
 \label{eq:p2-confidence}
\end{equation}
The self-planner defines $S_t$ as the set of the $k_t$ highest-scoring
positions among all $L$ answer positions. After the update, exactly
the positions in $S_t$ are unmasked. With a linear schedule,
$k_t=\lceil Lt/T\rceil$:
\begin{equation}
 S_t=\operatorname{TopK}(s^t,k_t).
 \label{eq:p2-selection}
\end{equation}
Given $S_t$, the P2 state transition is
\begin{equation}
 x_i^t =
 \begin{cases}
 x_i^{t-1}, & i\in S_t,\ x_i^{t-1}\ne\mask,\\
 \hat{x}_i^t, & i\in S_t,\ x_i^{t-1}=\mask,\\
 \mask, & i\notin S_t.
 \end{cases}
 \label{eq:transition}
\end{equation}
An unmasked token in $S_t$ is kept unchanged. If excluded from $S_t$, it is remasked and will be unmasked later. The number of unmasked positions increases with $t$, but the
selected sets are not necessarily nested. The last decoding step unmasks all remaining positions.

\subsection{Reward-based policy optimization}
\label{sec:grpo}

Group Relative Policy Optimization (GRPO; \citealp{grpo}) trains a
policy using rewards from groups of sampled outputs. A behavior
policy $\pi_{\phi_{\mathrm{old}}}$ generates $G$ trajectories for
each question $y$. With a terminal reward $R_{y,g}$ for trajectory
$g$, the group-normalized advantage is
\begin{equation}
 A_{y,g}=\frac{R_{y,g}-\bar R_y}{\sigma_y+\epsilon_A},
 \label{eq:advantage}
\end{equation}
where $\bar R_y$ and $\sigma_y$ are the within-group mean and standard
deviation, and $\epsilon_A>0$ stabilizes the denominator. All actions in
trajectory $g$ share its normalized advantage $A_{y,g}$.

For an action $a_t$ in state $\omega_t$, let
$r_t(\phi)=\pi_\phi(a_t\mid\omega_t)/
\pi_{\phi_{\mathrm{old}}}(a_t\mid\omega_t)$.
The PPO-style clipped surrogate objective is
\begin{equation}
 \ell(r,A)=\min\{rA,\operatorname{clip}(r,1-\epsilon,1+\epsilon)A\}.
 \label{eq:grpo-clipped}
\end{equation}
GRPO maximizes
\begin{equation}
 \mathbb E_{\mathrm{old}}\bigl[
 \ell(r_t(\phi),A_{y,g})
 -\beta D_{\mathrm{KL}}(\pi_\phi\|\pi_{\mathrm{ref}})\bigr].
 \label{eq:grpo-objective}
\end{equation}
The expectation averages over trajectories sampled from
$\pi_{\phi_{\mathrm{old}}}$ and their actions. The KL divergence is evaluated
at the corresponding states. Here $\pi_{\mathrm{ref}}$ is
a fixed reference policy and $\beta\geq 0$ controls the
KL penalty strength.

\section{Related Work}

\paragraph{Heuristic decoding rules.}
Many decoding methods for masked diffusion models use hand-designed
rules to choose which positions to unmask or remask. The basic rule unmasks the positions where the
model is most confident~\citep{llada}. Variants choose how many tokens to unmask based on confidence
thresholds or entropy bounds~\citep{fastdllm,eb-sampler}, wait until predictions are stable
across steps~\citep{klass}, avoid unmasking dependent tokens
together~\citep{dapd}, or reweight confidence with positional and
token-frequency priors~\citep{uncode}. These rules only reveal masked
tokens. \citet{two_temperatures} study how sampling temperatures affect sample diversity and the trade-off between sampling cost and pass@$k$.

A second line of work allows remasking: already unmasked tokens
can be remasked and predicted again. Mask-Predict~\citep{maskpredict}
refines non-autoregressive
translations by remasking low-confidence tokens and predicting
them again. Remasking methods select tokens using model
confidence~\citep{p2,rcr}, failed verification checks~\citep{wino},
sensitivity to context perturbations~\citep{core}, or a stochastic
remasking schedule~\citep{remdm}. Saber~\citep{saber} combines
adaptive unmasking with backtracking remasking for code generation.

\paragraph{Learned decoding with a frozen model.}
Several methods train a separate decoding module while keeping
the diffusion model frozen. \citet{apple} and \citet{upo} train unmasking policies with
reinforcement learning from task rewards. SAS~\citep{sas} uses GRPO with a
dense reward based on the likelihood of reference answers.
Adaptive order policies~\citep{adaptive-order} use the diffusion
model's token-prediction loss to learn which positions to
unmask. Learn2PD~\citep{learn2pd} and TraceLock~\citep{tracelock}
learn which predictions to accept from the model's own decoding
traces. These policies do not remask tokens.

Learned remasking modules use supervised training.
G-Star~\citep{gstar} and BackPlay~\citep{backplay} train
token-error predictors using reference data and use their
scores to choose which tokens to remask.

\paragraph{Adapting the diffusion model.}
A separate line of work fine-tunes the diffusion model itself with RL
or supervised training~\citep{dcolt,spg,mapg,remedi,pumba}.
MAPG~\citep{mapg} jointly optimizes token prediction and position
selection through the diffusion model's logits.
DCoLT~\citep{dcolt} uses GRPO with final-answer rewards to jointly train the diffusion model and a separate unmasking module. It also reports results from training only the
unmasking module while keeping the diffusion model frozen.

\paragraph{Concurrent work.}
TIPS~\citep{tips}, developed independently and concurrently,
also learns unmasking and remasking with a frozen LLaDA model.
Its probabilistic formulation of position selection differs
from ours, and its main training recipe uses supervised warmup
before RL (Appendix~\ref{app:tips}).
\section{Method}
\label{sec:method}
\begin{figure*}[!t]
 \centering
 \captionsetup{hypcap=false}
 \renewcommand{\ttdefault}{zi4}
 \lstset{
  basicstyle=\ttfamily\fontsize{7.6}{10.2}\selectfont,
  columns=fullflexible,
  keepspaces=true,
  aboveskip=2pt,
  belowskip=0pt,
  morecomment=[l]{\#},
  commentstyle=\color[HTML]{7A7A7A},
  morekeywords={for,in,if,while,continue,break,return,MASK},
  keywordstyle=\color[HTML]{1A3FD1},
  emph={where,argmax,ceil,top_k,gumbel,init,next_question,len,correct,
   sum,mean,std,append,plackett_luce_likelihood,update,clipped_objective},
  emphstyle=\color[HTML]{2B6CB0},
  emph={[2]frozen_model,mlp,decode},
  emphstyle={[2]\color[HTML]{6B46C1}},
 }
\begin{minipage}[t]{.49\textwidth}
  \vspace{0pt}
  \input{floats/inference-algorithm}
 \end{minipage}\hfill%
\begin{minipage}[t]{.49\textwidth}
  \vspace{0pt}
  \input{floats/training-algorithm}
 \end{minipage}
\end{figure*}

ReMaRL learns which answer positions to leave unmasked after each
non-final model call.
The diffusion model supplies the token predictions and stays frozen
while we train the position policy.

We retain P2's unmasking and remasking transition and make position
selection stochastic to train the policy with reinforcement
learning. Starting from an adapted confidence rule, ReMaRL learns
residual corrections to the position scores using final-answer
rewards.

\subsection{\texorpdfstring{P2$'$}{P2-prime} decoding}
\label{sec:selection}

We introduce P2$'$, which modifies the greedy P2 self-planner
in two ways: current-token scoring and stochastic position selection.
It scores each position by the probability of the token that would
occupy it if selected:
\begin{equation}
 \tilde{s}_i^t =
 \left\{
 \begin{array}{@{}l@{\;}l@{}}
 s_i^t, & x_i^{t-1}=\mask,\\[2pt]
 p_{\theta,i}(x_i^{t-1}\mid y,x^{t-1}),
     & x_i^{t-1}\ne\mask.
 \end{array}
 \right.
 \label{eq:confidence}
\end{equation}
Here, $s_i^t$ is the confidence score used by the original P2
self-planner (Eq.~\ref{eq:p2-confidence}).
At an unmasked position, selection retains the current token even when
the model proposes a different one. Appendix~\ref{app:p2-scoring}
compares this rule with the original P2 self-planner.

At each non-final call, we divide each position score by the
temperature $\tau>0$ and add independent noise
$g_i^t\sim\mathrm{Gumbel}(0,1)$. We select the $k_t$ positions
with the largest values of
\begin{equation}
 \tilde{s}_i^t/\tau+g_i^t,
 \label{eq:position-sampling}
\end{equation}
and record their indices as $a_t=(a_{t,1},\ldots,a_{t,k_t})$, ordered by decreasing perturbed score.

This is equivalent to sampling positions one at a time without
replacement, with probabilities proportional to
$\exp(\tilde{s}_i^t/\tau)$ among the remaining positions.
This procedure is called Gumbel top-$k$~\citep{gumbel-topk}.
The ordered list follows a Plackett--Luce distribution:
\begin{equation}
 \operatorname{PL}(a_t;\tilde{s}^t/\tau)=
 \prod_{j=1}^{k_t}
 \frac{\exp(\tilde{s}_{a_{t,j}}^t/\tau)}
 {\displaystyle\sum_{i\notin a_{t,<j}}
  \exp(\tilde{s}_i^t/\tau)},
 \label{eq:pl}
\end{equation}
where $a_{t,<j}=\{a_{t,1},\ldots,a_{t,j-1}\}$ contains the
previously selected indices.

We form the set $S_t=\{a_{t,1},\ldots,a_{t,k_t}\}$ and use it
in the P2 state transition (Equation~\ref{eq:transition}).
The temperature controls the randomness of position selection ($\tau=0$ denotes greedy selection).

\subsection{Learning the position scores}

ReMaRL learns a correction to the confidence score at each position.
Given the position's feature vector $f_i^t$, the policy computes
\begin{align}
 z_i^t &= \tilde{s}_i^t+\Delta_\phi(f_i^t), \label{eq:residual}\\
 \Delta_\phi(f) &= \mathrm{MLP}_\phi(f)+w_{\mathrm{skip}}^\top f.
\end{align}
The policy shares its weights across positions.

The features combine confidence $\tilde{s}_i^t$, the mask indicator
$m_i^t=\mathbf{1}\{x_i^{t-1}=\mask\}$, a learned position embedding
$e_i$, and a projection of the diffusion model's final hidden state $h_i^t$:
\begin{align}
 u_i^t &= W_h\,\mathrm{LN}(h_i^t)+b_h,\\
 f_i^t &= [\tilde{s}_i^t;\,m_i^t;\,e_i;\,u_i^t]. \label{eq:features}
\end{align}
The model call that produces the token probabilities also supplies $h_i^t$.
We use $e_i\in\mathbb R^{16}$ and $u_i^t\in\mathbb R^{64}$.
The network $\mathrm{MLP}_\phi$ has two 64-unit hidden layers
with GELU activations. The
policy has about 0.3M trainable parameters.

We initialize the final MLP layer and the linear skip weights to zero.
The initial scores therefore equal $\tilde{s}_i^t$, and ReMaRL starts from
stochastic P2$'$ at the training temperature $\tau=1$.
Algorithm~\ref{alg:inference}
gives the decoding procedure.

\begin{figure*}[!t]
 \centering
 \includegraphics[width=\linewidth]{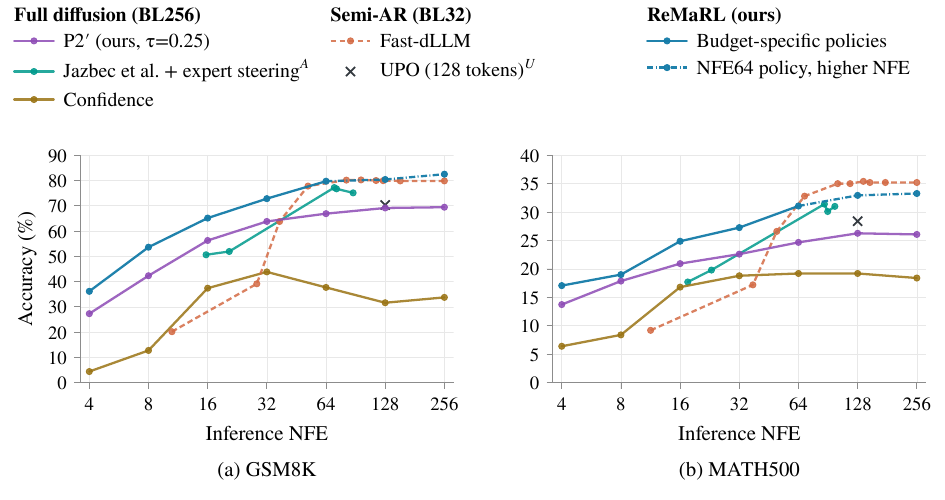}
 \caption{Math accuracy versus inference budget. The solid blue line connects ReMaRL policies trained for 32k groups
and evaluated at their respective training budgets.
 The dash-dot curve reuses the NFE64 checkpoint at larger
 budgets.
 Jazbec et al.\textsuperscript{A} and UPO\textsuperscript{U} results
are taken from their papers; UPO uses 128-token answers.}
 \label{fig:math}
\end{figure*}

\subsection{Training with GRPO}
\label{sec:training}

We train the position policy with GRPO~\citep{grpo} using
final-answer correctness as the reward. At each non-final call,
the state is $\omega_t=(y,x^{t-1},t)$. We sample $a_t$ using the learned scores $z^t$ as described in
Section~\ref{sec:selection} and apply the state transition
in Equation~\ref{eq:transition}.

The policy probability is
$\pi_\phi(a_t\mid\omega_t)=\operatorname{PL}(a_t;z^t/\tau)$,
where $\operatorname{PL}$ is defined in Equation~\ref{eq:pl}.

For each question $y$, the policy generates $G$ answers with
Algorithm~\ref{alg:inference}. The task's checker assigns each answer
a binary correctness reward $R_{y,g}$.
We compute the advantages in Equation~\ref{eq:advantage} using the population standard
deviation. Constant-reward groups have zero normalized advantages.
Following the dynamic sampling procedure in DAPO~\citep{dapo},
we discard these groups and sample new question groups until
the batch is full.

For each non-final decoding step, we form the likelihood ratio
\begin{equation}
 r_{y,g,t}(\phi)=
\frac{\pi_\phi(a_{y,g,t}\mid\omega_{y,g,t})}
     {\pi_{\phi_{\mathrm{old}}}(a_{y,g,t}\mid\omega_{y,g,t})}.
\end{equation}
Let $\mathcal D$ contain the non-final decisions from the accepted
groups. We minimize the negative GRPO surrogate:
\begin{equation}
 \mathcal L(\phi)
 =-\frac{1}{|\mathcal D|}
   \sum_{(y,g,t)\in\mathcal D}
   \ell\bigl(r_{y,g,t}(\phi),A_{y,g}\bigr).
 \label{eq:loss}
\end{equation}
We do not use a reference policy, so we omit the KL penalty.
For each batch, we take $E=4$ full-batch optimization steps,
then sample new answers with the updated policy
(Algorithm~\ref{alg:training}).

\begin{figure*}[!t]
 \centering
 \includegraphics[width=\linewidth]{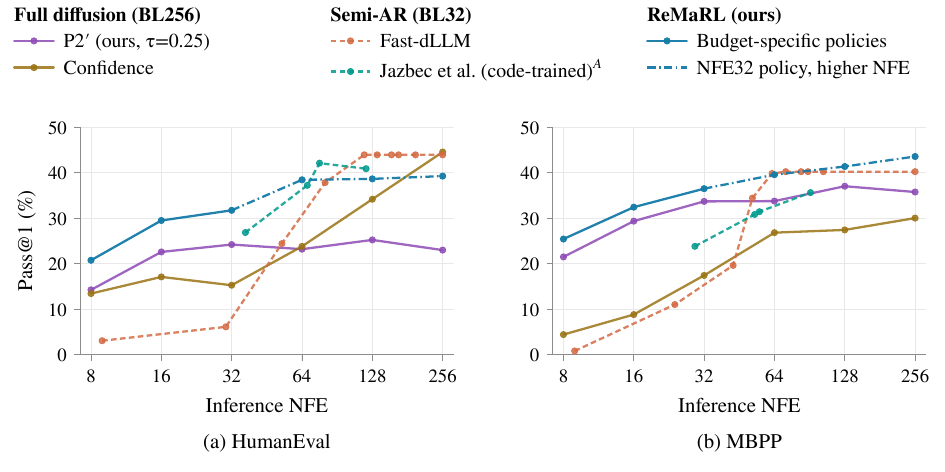}
 \caption{Code pass@1 versus inference budget for selected methods. The solid blue line connects three ReMaRL policies trained for 32k
groups, one at each budget of 8, 16, and 32 calls. Each is evaluated
at its training budget.
 The dash-dot ReMaRL curve reuses the NFE32 checkpoint
 at larger budgets.
 Full-diffusion and semi-autoregressive results are grouped in
 separate legend columns. Results marked \textsuperscript{A} are taken from \citet{apple}.}
 \label{fig:code}
\end{figure*}

\section{Experiments}
\label{sec:experiments}

\subsection{Setup}

\paragraph{Model and decoding.}
We use the same frozen LLaDA-8B-Instruct
checkpoint~\citep{llada} for the main experiments. ReMaRL generates
256-token answers with full-diffusion decoding. Our evaluations
follow the task-specific prompting and scoring of \citet{apple}.

\paragraph{Training data and budgets.}
We train a separate math policy for each fixed inference budget
of 4, 8, 16, 32, and 64 calls, using the 14,973-question GSM8K
and MATH mixture~\citep{gsm8k,math}. Separate code policies are
trained at 8, 16, and 32 calls on about 9.1k KodCode
problems~\citep{kodcode}. We screen code training problems for
reference-test compatibility and benchmark overlap
(Appendix~\ref{app:data}).

Main policies use 32,768 accepted question groups (32k).
Each batch contains $B=16$ informative question groups with $G=8$ sampled
answers per question. We use a learning rate of $10^{-4}$, a GRPO
clipping threshold of $\epsilon=0.2$, and gradient norm clipping at 1.
Appendix~\ref{app:implementation} gives implementation details.

\paragraph{Evaluation.}
We report accuracy on GSM8K and MATH500~\citep{math500}, and
pass@1 on HumanEval~\citep{humaneval} and MBPP~\citep{mbpp}. For stochastic configurations, we average scores over three inference seeds.

\paragraph{Compared methods.}
We compare ReMaRL with prior decoding methods and our P2$'$
planner at position temperatures $\tau\in\{0,0.25,1\}$.
We select $\tau=0.25$ separately for math and code using
training questions at eight calls and reuse it across
inference budgets (Section~\ref{sec:temperature}).
Figures~\ref{fig:math} and~\ref{fig:code} show a representative
subset of methods to keep the curves readable. Confidence decoding and Fast-dLLM represent prior heuristic
methods, while the published policies provide comparisons with
other learned methods.
Appendix~\ref{app:full} gives the full results for each benchmark.

We evaluate most prior methods ourselves and reuse published policy scores from \citet{apple,upo,sas}.
We also include DCoLT's published frozen-model ablation~\citep{dcolt}.
Appendix~\ref{app:tips} compares ReMaRL with concurrent TIPS.

\subsection{Main results}

\paragraph{Math.}
ReMaRL outperforms the compared full-diffusion methods on
both GSM8K and MATH500 at 4--64 model calls
(Figure~\ref{fig:math}; Appendix~\ref{app:full}).
At 128 and 256 calls, it continues to lead on GSM8K,
while PC-Sampler scores higher on MATH500.

ReMaRL also outperforms the compared semi-autoregressive
baselines on both benchmarks at 8--32 calls. At larger budgets,
it remains competitive with the strongest semi-autoregressive
methods and achieves the highest GSM8K accuracy at 256 calls.

Calibrated P2$'$ ranks second to ReMaRL on both math
benchmarks at 4--16 calls and on GSM8K at 32 calls,
outperforming the other compared methods at these budgets
without training a position policy.

\paragraph{Code.}
At inference budgets of 8 and 16 model calls, ReMaRL achieves the highest pass@1 on both HumanEval and MBPP among the compared methods (Figure~\ref{fig:code}; Appendix~\ref{app:full}).

At 16 calls, calibrated P2$'$ also ranks second to ReMaRL
on both code benchmarks. On MBPP, this ranking holds
at 8 and 32 calls as well.

At 128 and 256 calls, ReMaRL achieves the highest pass@1 on MBPP among all compared methods, while several full-diffusion and semi-autoregressive methods score higher on HumanEval.

\section{Analysis}
\begin{figure*}[!t]
 \centering
 \includegraphics[width=\linewidth]{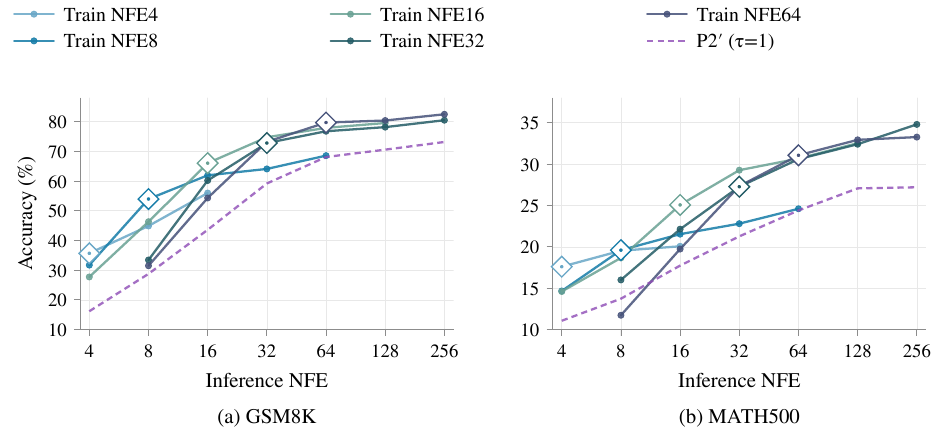}
 \caption{Math checkpoints trained for 32k groups, reevaluated at different inference budgets.
 Dot-centered diamonds mark matching training and inference budgets.}
 \label{fig:cross}
\end{figure*}

\subsection{Transfer across inference budgets}

To test whether a policy can be reused at a different inference
budget, we evaluate each math checkpoint at other inference budgets without further training (Figure~\ref{fig:cross}).

At a fixed inference budget, the policy trained for that budget
generally performs best. Policies trained with fewer
or more calls can still achieve similar accuracy at the same
inference budget, without further training. For every evaluated
checkpoint, mean accuracy increases with the number of inference
calls on both benchmarks.

\subsection{Ablations}

\begin{table}[t]
\centering
\small
\setlength{\tabcolsep}{3pt}
\caption{Math accuracy (\%) at eight calls after 16k accepted training
groups.}
\label{tab:ablations}
\begin{tabular}{@{}lrr@{}}
\toprule
Variant & GSM8K & MATH500 \\
\midrule
\textbf{ReMaRL} & \textbf{50.04} & \textbf{20.00} \\
No position embedding & 48.98 & 19.73 \\
No hidden states & 45.08 & 18.20 \\
No remasking & 38.26 & 15.00 \\
Position and mask biases only & 29.16 & 14.07 \\
Random init., no added confidence & 47.54 & 18.20 \\
Transformer policy & 45.41 & 17.80 \\
\bottomrule
\end{tabular}
\end{table}

We examine how ReMaRL's components, initialization, and architecture
affect math accuracy. We train each variant from scratch at eight calls for
16k accepted groups.

We remove position embeddings, hidden states, and remasking in
separate variants. In the bias-only variant, the correction to
token confidence depends only on position and whether it is masked.

In another variant, we randomly initialize the final MLP layer and linear skip and remove the confidence term $\tilde{s}_i^t$ from Equation~\ref{eq:residual}.

Following the use of a transformer policy in \citet{apple}, we test whether attention across answer positions improves accuracy. We replace the MLP with a transformer, keeping the same inputs, initialization, and training recipe.

Table~\ref{tab:ablations} shows that removing position embeddings has a small effect on accuracy, while removing hidden states or disabling remasking leads to larger accuracy drops.
Restricting the correction to position and mask-state biases gives
the lowest accuracy. Random initialization without the added confidence score also lowers accuracy compared with standard ReMaRL. 

The transformer has roughly twice as many parameters but achieves lower accuracy on both benchmarks at the same training budget. The diffusion model's hidden states already encode information about other answer positions, so the MLP can use this context when scoring each position. This may reduce the benefit of adding attention within the policy.

\subsection{Position temperature}
\label{sec:temperature}

\begin{figure*}[!h]
 \centering
 \includegraphics[width=\linewidth]{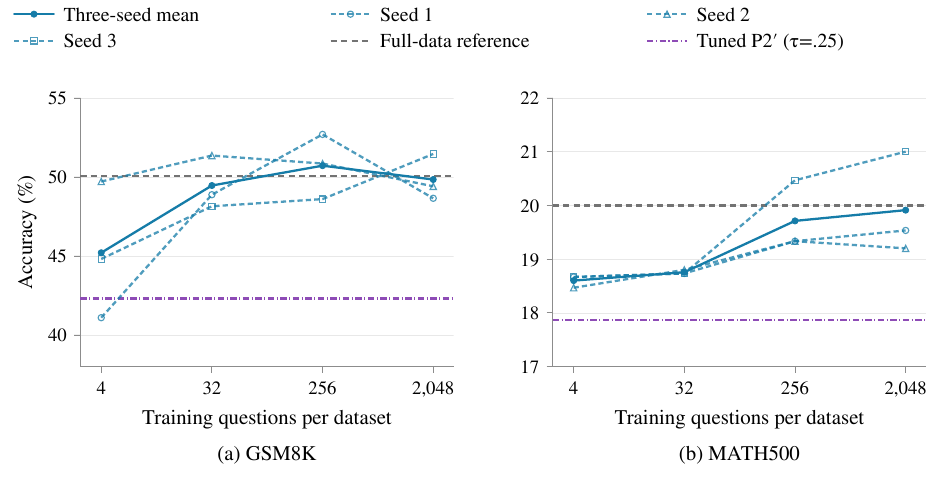}
 \caption{Training-data size at NFE8 and 16k training groups. Dashed blue
 curves track the three seeds across sizes; each seed controls both
 subset sampling and training. The darker curve is their mean.}
 \label{fig:data-size}
\end{figure*}

\begin{table*}[h]
\centering
\small
\setlength{\tabcolsep}{5pt}
\caption{P2$'$ position-temperature sweeps and ReMaRL on the
full benchmarks (\%). Here $\tau=0$ denotes greedy position
selection. Gray shading marks the temperature selected
separately for each domain on training questions at eight
calls and used unchanged at both budgets. Bold marks the
highest score in each column.}
\label{tab:temperature-sweep}
\begin{tabular}{@{}lrrrrrrrr@{}}
\toprule
& \multicolumn{4}{c}{Math}
& \multicolumn{4}{c}{Code} \\
\cmidrule(lr){2-5}\cmidrule(l){6-9}
& \multicolumn{2}{c}{8 calls}
& \multicolumn{2}{c}{16 calls}
& \multicolumn{2}{c}{8 calls}
& \multicolumn{2}{c}{16 calls} \\
\cmidrule(lr){2-3}\cmidrule(lr){4-5}
\cmidrule(lr){6-7}\cmidrule(l){8-9}
P2$'$ temperature
& GSM8K & MATH500 & GSM8K & MATH500
& HumanEval & MBPP & HumanEval & MBPP \\
\midrule
$0$
& 23.50 & 13.60 & 42.68 & 19.00
& 14.02 & 3.00 & 17.68 & 11.40 \\
$0.05$
& 29.19 & 14.80 & 41.57 & 16.87
& 13.01 & 4.60 & 18.70 & 9.33 \\
$0.1$
& 34.04 & 16.00 & 46.25 & 18.00
& 13.41 & 9.27 & 18.90 & 17.73 \\
$0.25$
& \cellcolor{black!12}42.30 & \cellcolor{black!12}17.87
& \cellcolor{black!12}56.25 & \cellcolor{black!12}20.93
& \cellcolor{black!12}14.23 & \cellcolor{black!12}21.47
& \cellcolor{black!12}22.56 & \cellcolor{black!12}29.33 \\
$0.5$
& 34.62 & 15.13 & 50.14 & 18.00
& 11.79 & 18.27 & 18.09 & 25.20 \\
$1$
& 28.76 & 13.73 & 43.59 & 17.73
& 11.59 & 15.73 & 14.84 & 23.33 \\
$2$
& 25.70 & 12.40 & 40.23 & 16.33
& 10.77 & 13.07 & 12.40 & 21.67 \\
\midrule
ReMaRL ($\tau=1$)
& \textbf{53.60} & \textbf{19.00}
& \textbf{65.13} & \textbf{24.87}
& \textbf{20.73} & \textbf{25.40}
& \textbf{29.47} & \textbf{32.40} \\
\bottomrule
\end{tabular}
\end{table*}

The residual policy can uniformly rescale confidence scores by a
positive factor, which is equivalent to changing the position
temperature. We test how much of ReMaRL's improvement can be
recovered by tuning this temperature in P2$'$
(Table~\ref{tab:temperature-sweep}). For each domain, we select
a temperature at eight calls using 256 training questions:
128 each from GSM8K and MATH for math, and 256 from KodCode
for code.

Temperature tuning improves P2$'$ and closes part of the gap
to ReMaRL. Training-based selection chooses $\tau=0.25$ in
both domains, which also achieves the highest P2$'$ test
accuracy among the tested temperatures at eight and sixteen
calls. ReMaRL outperforms calibrated P2$'$ on all four
benchmarks at both budgets.

We also vary the inference temperature of the trained NFE8 ReMaRL
policy. On GSM8K, accuracy peaks near the training temperature of
$\tau=1$ (Table~\ref{tab:remarl-temperature-sensitivity}), indicating
that the learned scores already have a suitable scale for sampling.

\subsection{Training-data size}
\label{sec:data-size}

We train NFE8 policies on 4, 32, 256, or 2,048 questions
per math dataset, using three training seeds and 16k accepted
groups per run. Figure~\ref{fig:data-size} compares them with
the policy trained on all 14,973 questions at the same
training budget and with temperature-calibrated P2$'$.

Even with only four training questions per dataset, ReMaRL’s mean accuracy across three runs exceeds that of temperature-calibrated P2$'$
on both benchmarks. With 32 questions per dataset, GSM8K accuracy
approaches that of the full-data policy. Increasing the subset to
256 questions per dataset also brings MATH500 accuracy close to
the full-data result. These results show that ReMaRL is data-efficient: a small training
subset is sufficient to approach full-data accuracy.

\subsection{Evaluation on Dream}
\label{sec:dream}

To test ReMaRL with another diffusion model, we train separate math
and code policies on frozen Dream-v0-Instruct-7B~\citep{dream}
at 8, 16, and 32 calls, using the same training recipe.
ReMaRL achieves the highest scores among the compared methods
at 8--32 calls on GSM8K and MATH500 and at 8--16 calls on
HumanEval and MBPP. The setup and full results are reported in
Appendix~\ref{app:dream}.

\subsection{Generation trajectories}
\label{sec:trajectories}

We analyze GSM8K trajectories to understand how ReMaRL
selects positions during generation. \mbox{ReMaRL} tends to
prioritize earlier answer positions, producing a pattern
similar to autoregressive generation. This behavior emerges
without an explicit left-to-right constraint or reward
for following this order.

ReMaRL also remasks fewer tokens than P2$'$ at
both $\tau=0.25$ and $\tau=1$.
Appendix~\ref{app:trajectories} presents the trajectory
analysis.

\section{Conclusion}

ReMaRL learns which positions to unmask and which unmasked tokens
to remask. We train only the 0.3M-parameter position policy, using
final-answer rewards while keeping the diffusion model frozen.
On both tested backbones, ReMaRL outperforms the compared baselines
at the tested budgets of 8--32 model calls on math and 8--16
calls on code. %

\Needspace{10\baselineskip}
\section{Limitations}

ReMaRL learns which positions to reveal, retain, or remask,
while the number of unmasked positions at each step and
the total inference budget are fixed in advance.
Our training recipe relies on task-specific correctness
rewards; extending it to open-ended generation requires
a suitable reward signal.

We evaluate two diffusion models on math and code tasks with
256-token answers. Longer answers and other task types remain
untested.

Training budgets count accepted question groups, so
resampling constant-reward groups adds rollout cost that
can differ across policies. At inference, NFE does not
capture sequence length, caching, or policy overhead,
and therefore does not establish wall-clock speedups.

\FloatBarrier
\bibliography{references}
\onecolumn
\appendix
\section{Full comparison tables}
\label{app:full}

\subsection{Math}

\begingroup
\definecolor{scorebar}{HTML}{D9ECF5}
\ifdefined\scorecellwidth\else\newlength{\scorecellwidth}\fi
\setlength{\scorecellwidth}{36pt}
\newcommand{\scorecell}[3]{%
 \makebox[\scorecellwidth][l]{%
  \rlap{\raisebox{-2.2pt}[0pt][0pt]{\color{scorebar}\rule{#1\scorecellwidth}{10pt}}}%
  \makebox[30pt][r]{#2}\makebox[6pt][l]{#3}}}
\newcommand{\emptycell}{\makebox[\scorecellwidth][c]{\textemdash}}

\fontsize{9.5}{10.5}\selectfont
\setlength{\LTcapwidth}{\textwidth}
\setlength{\LTpre}{3pt}
\setlength{\tabcolsep}{2.5pt}
\renewcommand{\arraystretch}{0.96}
\begin{longtable}{@{}>{\raggedright\arraybackslash}p{4.7cm} c *{7}{r}@{}}
\caption{GSM8K accuracy (\%). Adaptive cells show the best observed test score within each mean-NFE budget, with mean NFE below. Bars scale from zero to the maximum score in each column; bold marks that score. Unmarked: computed here; \textsuperscript{A}: Jazbec et al.; \textsuperscript{U}: UPO; \textsuperscript{S}: SAS; \textsuperscript{D}: DCoLT (\citealp{dcolt}, Table~4; published prompt/scorer). NR: block size unspecified. Dashes indicate unavailable results.}\label{tab:baseline-gsm8k}\\
\toprule
 & & \multicolumn{7}{c}{Mean-NFE budget}\\
Method & Blocks & \makebox[36pt][c]{4} & \makebox[36pt][c]{8} & \makebox[36pt][c]{16} & \makebox[36pt][c]{32} & \makebox[36pt][c]{64} & \makebox[36pt][c]{128} & \makebox[36pt][c]{256} \\
\midrule
\endfirsthead
\multicolumn{9}{l}{\textit{GSM8K comparison (continued)}}\\
\toprule
 & & \multicolumn{7}{c}{Mean-NFE budget}\\
Method & Blocks & \makebox[36pt][c]{4} & \makebox[36pt][c]{8} & \makebox[36pt][c]{16} & \makebox[36pt][c]{32} & \makebox[36pt][c]{64} & \makebox[36pt][c]{128} & \makebox[36pt][c]{256} \\
\midrule
\endhead
\bottomrule
\endfoot
\multicolumn{9}{l}{\textit{Fixed-budget decoding}}\\
\textbf{ReMaRL (ours)} & $1\times256$ & \scorecell{1.00000000}{\textbf{36.14}}{} & \scorecell{1.00000000}{\textbf{53.60}}{} & \scorecell{1.00000000}{\textbf{65.13}}{} & \scorecell{1.00000000}{\textbf{72.81}}{} & \scorecell{1.00000000}{\textbf{79.71}}{} & \emptycell & \emptycell \\
ReMaRL (ours), trained at NFE64 & $1\times256$ & \emptycell & \emptycell & \emptycell & \emptycell & \scorecell{1.00000000}{\textbf{79.71}}{} & \scorecell{0.99749138}{80.39}{} & \scorecell{1.00000000}{\textbf{82.49}}{} \\
P2$'$ (ours, $\tau=0.25$) & $1\times256$ & \scorecell{0.75521317}{27.29}{} & \scorecell{0.78926822}{42.30}{} & \scorecell{0.86373005}{56.25}{} & \scorecell{0.87605681}{63.79}{} & \scorecell{0.83890152}{66.87}{} & \scorecell{0.85765006}{69.12}{} & \scorecell{0.84157205}{69.42}{} \\
P2$'$ (ours, $\tau=1$) & $1\times256$ & \scorecell{0.44755245}{16.17}{} & \scorecell{0.53653937}{28.76}{} & \scorecell{0.66938300}{43.59}{} & \scorecell{0.81291218}{59.19}{} & \scorecell{0.85447051}{68.11}{} & \scorecell{0.87582314}{70.58}{} & \scorecell{0.88694853}{73.16}{} \\
P2$'$ (ours, greedy) & $1\times256$ & \scorecell{0.37132867}{13.42}{} & \scorecell{0.43847242}{23.50}{} & \scorecell{0.65541327}{42.68}{} & \scorecell{0.66539396}{48.45}{} & \scorecell{0.56404566}{44.96}{} & \scorecell{0.48165569}{38.82}{} & \scorecell{0.50643382}{41.77}{} \\
P2 (original self-planner) & $1\times256$ & \scorecell{0.15734266}{5.69}{} & \scorecell{0.31966054}{17.13}{} & \scorecell{0.61466822}{40.03}{} & \scorecell{0.68413745}{49.81}{} & \scorecell{0.55738744}{44.43}{} & \scorecell{0.49482596}{39.88}{} & \scorecell{0.51194853}{42.23}{} \\
Confidence & $1\times256$ & \scorecell{0.12167832}{4.40}{} & \scorecell{0.23762376}{12.74}{} & \scorecell{0.57392317}{37.38}{} & \scorecell{0.60187435}{43.82}{} & \scorecell{0.47273304}{37.68}{} & \scorecell{0.39228598}{31.61}{} & \scorecell{0.40900735}{33.74}{} \\
Confidence & $8\times32$ & \emptycell & \scorecell{0.38896747}{20.85}{} & \scorecell{0.48661234}{31.69}{} & \scorecell{0.61957654}{45.11}{} & \scorecell{0.89220038}{71.11}{} & \scorecell{0.95860771}{77.26}{} & \scorecell{0.96783088}{79.83}{} \\
Random & $1\times256$ & \scorecell{0.31398601}{11.35}{} & \scorecell{0.36916549}{19.79}{} & \scorecell{0.46100116}{30.02}{} & \scorecell{0.56056925}{40.81}{} & \scorecell{0.56943564}{45.39}{} & \scorecell{0.62119787}{50.06}{} & \scorecell{0.63541667}{52.41}{} \\
Random & $8\times32$ & \emptycell & \scorecell{0.38896747}{20.85}{} & \scorecell{0.37524253}{24.44}{} & \scorecell{0.40992711}{29.85}{} & \scorecell{0.58021560}{46.25}{} & \scorecell{0.78927563}{63.61}{} & \scorecell{0.87285539}{72.00}{} \\
Margin & $1\times256$ & \scorecell{0.27272727}{9.86}{} & \scorecell{0.37765205}{20.24}{} & \scorecell{0.59371362}{38.67}{} & \scorecell{0.62478306}{45.49}{} & \scorecell{0.50126823}{39.95}{} & \scorecell{0.38946378}{31.39}{} & \scorecell{0.41268382}{34.04}{} \\
Margin & $8\times32$ & \emptycell & \scorecell{0.38896747}{20.85}{} & \scorecell{0.47613504}{31.01}{} & \scorecell{0.65081569}{47.38}{} & \scorecell{0.89315155}{71.19}{} & \scorecell{0.95390405}{76.88}{} & \scorecell{0.94944853}{78.32}{} \\
ReMDM (UNCODE) & $1\times256$ & \emptycell & \emptycell & \scorecell{0.50989523}{33.21}{} & \scorecell{0.58729608}{42.76}{} & \scorecell{0.56214331}{44.81}{} & \scorecell{0.46754468}{37.68}{} & \scorecell{0.38327206}{31.61}{} \\
UNCODE/PC-Sampler & $1\times256$ & \scorecell{0.41538462}{15.01}{} & \scorecell{0.31400283}{16.83}{} & \scorecell{0.40745052}{26.54}{} & \scorecell{0.48212426}{35.10}{} & \scorecell{0.78376665}{62.47}{} & \scorecell{0.95390405}{76.88}{} & \scorecell{0.97610294}{80.52}{} \\
RCR & $1\times256$ & \scorecell{0.21188811}{7.66}{} & \scorecell{0.42857143}{22.97}{} & \scorecell{0.59604191}{38.82}{} & \scorecell{0.62165915}{45.26}{} & \scorecell{0.51648700}{41.17}{} & \scorecell{0.44967074}{36.24}{} & \scorecell{0.46047794}{37.98}{} \\
\midrule
\multicolumn{9}{l}{\textit{Adaptive decoding}}\\
Fast-dLLM & $1\times256$ & \begin{tabular}[t]{@{}c@{}}\scorecell{0.11118881}{4.02}{}\\[-2pt]{\fontsize{7.5}{9}\selectfont\color{black!65}(2.7)}\end{tabular} & \begin{tabular}[t]{@{}c@{}}\scorecell{0.07496464}{4.02}{}\\[-2pt]{\fontsize{7.5}{9}\selectfont\color{black!65}(2.7)}\end{tabular} & \begin{tabular}[t]{@{}c@{}}\scorecell{0.30966240}{20.17}{}\\[-2pt]{\fontsize{7.5}{9}\selectfont\color{black!65}(11.3)}\end{tabular} & \begin{tabular}[t]{@{}c@{}}\scorecell{0.51232211}{37.30}{}\\[-2pt]{\fontsize{7.5}{9}\selectfont\color{black!65}(22.6)}\end{tabular} & \begin{tabular}[t]{@{}c@{}}\scorecell{0.46797717}{37.30}{}\\[-2pt]{\fontsize{7.5}{9}\selectfont\color{black!65}(22.6)}\end{tabular} & \begin{tabular}[t]{@{}c@{}}\scorecell{0.46284102}{37.30}{}\\[-2pt]{\fontsize{7.5}{9}\selectfont\color{black!65}(22.6)}\end{tabular} & \begin{tabular}[t]{@{}c@{}}\scorecell{0.45220588}{37.30}{}\\[-2pt]{\fontsize{7.5}{9}\selectfont\color{black!65}(22.6)}\end{tabular} \\
Fast-dLLM & $8\times32$ & \emptycell & \emptycell & \begin{tabular}[t]{@{}c@{}}\scorecell{0.30966240}{20.17}{}\\[-2pt]{\fontsize{7.5}{9}\selectfont\color{black!65}(10.5)}\end{tabular} & \begin{tabular}[t]{@{}c@{}}\scorecell{0.53731343}{39.12}{}\\[-2pt]{\fontsize{7.5}{9}\selectfont\color{black!65}(28.5)}\end{tabular} & \begin{tabular}[t]{@{}c@{}}\scorecell{0.97590361}{77.79}{}\\[-2pt]{\fontsize{7.5}{9}\selectfont\color{black!65}(51.8)}\end{tabular} & \begin{tabular}[t]{@{}c@{}}\scorecell{0.99529633}{80.21}{}\\[-2pt]{\fontsize{7.5}{9}\selectfont\color{black!65}(96.4)}\end{tabular} & \begin{tabular}[t]{@{}c@{}}\scorecell{0.97242647}{80.21}{}\\[-2pt]{\fontsize{7.5}{9}\selectfont\color{black!65}(96.4)}\end{tabular} \\
EB-Sampler & $1\times256$ & \emptycell & \emptycell & \emptycell & \begin{tabular}[t]{@{}c@{}}\scorecell{0.47587643}{34.65}{}\\[-2pt]{\fontsize{7.5}{9}\selectfont\color{black!65}(21.7)}\end{tabular} & \begin{tabular}[t]{@{}c@{}}\scorecell{0.43468611}{34.65}{}\\[-2pt]{\fontsize{7.5}{9}\selectfont\color{black!65}(21.7)}\end{tabular} & \begin{tabular}[t]{@{}c@{}}\scorecell{0.42991533}{34.65}{}\\[-2pt]{\fontsize{7.5}{9}\selectfont\color{black!65}(21.7)}\end{tabular} & \begin{tabular}[t]{@{}c@{}}\scorecell{0.42003676}{34.65}{}\\[-2pt]{\fontsize{7.5}{9}\selectfont\color{black!65}(21.7)}\end{tabular} \\
EB-Sampler & $8\times32$ & \emptycell & \emptycell & \emptycell & \begin{tabular}[t]{@{}c@{}}\scorecell{0.66018743}{48.07}{}\\[-2pt]{\fontsize{7.5}{9}\selectfont\color{black!65}(30.7)}\end{tabular} & \begin{tabular}[t]{@{}c@{}}\scorecell{0.96639188}{77.03}{}\\[-2pt]{\fontsize{7.5}{9}\selectfont\color{black!65}(54.0)}\end{tabular} & \begin{tabular}[t]{@{}c@{}}\scorecell{1.00000000}{\textbf{80.59}}{}\\[-2pt]{\fontsize{7.5}{9}\selectfont\color{black!65}(93.2)}\end{tabular} & \begin{tabular}[t]{@{}c@{}}\scorecell{0.98897059}{81.58}{}\\[-2pt]{\fontsize{7.5}{9}\selectfont\color{black!65}(187.3)}\end{tabular} \\
Jazbec et al. & $1\times256$ & \emptycell & \emptycell & \begin{tabular}[t]{@{}c@{}}\scorecell{0.74164959}{48.30}{\textsuperscript{A}}\\[-2pt]{\fontsize{7.5}{9}\selectfont\color{black!65}(13.0)}\end{tabular} & \begin{tabular}[t]{@{}c@{}}\scorecell{0.75266782}{54.80}{\textsuperscript{A}}\\[-2pt]{\fontsize{7.5}{9}\selectfont\color{black!65}(27.1)}\end{tabular} & \begin{tabular}[t]{@{}c@{}}\scorecell{0.70382911}{56.10}{\textsuperscript{A}}\\[-2pt]{\fontsize{7.5}{9}\selectfont\color{black!65}(50.5)}\end{tabular} & \begin{tabular}[t]{@{}c@{}}\scorecell{0.76683161}{61.80}{\textsuperscript{A}}\\[-2pt]{\fontsize{7.5}{9}\selectfont\color{black!65}(93.3)}\end{tabular} & \begin{tabular}[t]{@{}c@{}}\scorecell{0.74921140}{61.80}{\textsuperscript{A}}\\[-2pt]{\fontsize{7.5}{9}\selectfont\color{black!65}(93.3)}\end{tabular} \\
Jazbec et al. + expert steering & $1\times256$ & \emptycell & \emptycell & \begin{tabular}[t]{@{}c@{}}\scorecell{0.77696624}{50.60}{\textsuperscript{A}}\\[-2pt]{\fontsize{7.5}{9}\selectfont\color{black!65}(15.7)}\end{tabular} & \begin{tabular}[t]{@{}c@{}}\scorecell{0.71283686}{51.90}{\textsuperscript{A}}\\[-2pt]{\fontsize{7.5}{9}\selectfont\color{black!65}(20.6)}\end{tabular} & \begin{tabular}[t]{@{}c@{}}\scorecell{0.65113602}{51.90}{\textsuperscript{A}}\\[-2pt]{\fontsize{7.5}{9}\selectfont\color{black!65}(20.6)}\end{tabular} & \begin{tabular}[t]{@{}c@{}}\scorecell{0.95791910}{77.20}{\textsuperscript{A}}\\[-2pt]{\fontsize{7.5}{9}\selectfont\color{black!65}(70.5)}\end{tabular} & \begin{tabular}[t]{@{}c@{}}\scorecell{0.93590809}{77.20}{\textsuperscript{A}}\\[-2pt]{\fontsize{7.5}{9}\selectfont\color{black!65}(70.5)}\end{tabular} \\
Jazbec et al. ($\alpha$ sweep) & $8\times32$ & \emptycell & \emptycell & \begin{tabular}[t]{@{}c@{}}\scorecell{0.54049825}{35.20}{\textsuperscript{A}}\\[-2pt]{\fontsize{7.5}{9}\selectfont\color{black!65}(10.3)}\end{tabular} & \begin{tabular}[t]{@{}c@{}}\scorecell{0.48346546}{35.20}{\textsuperscript{A}}\\[-2pt]{\fontsize{7.5}{9}\selectfont\color{black!65}(10.3)}\end{tabular} & \begin{tabular}[t]{@{}c@{}}\scorecell{0.96102156}{76.60}{\textsuperscript{A}}\\[-2pt]{\fontsize{7.5}{9}\selectfont\color{black!65}(50.7)}\end{tabular} & \begin{tabular}[t]{@{}c@{}}\scorecell{0.99266228}{80.00}{\textsuperscript{A}}\\[-2pt]{\fontsize{7.5}{9}\selectfont\color{black!65}(82.7)}\end{tabular} & \begin{tabular}[t]{@{}c@{}}\scorecell{0.97591452}{80.50}{\textsuperscript{A}}\\[-2pt]{\fontsize{7.5}{9}\selectfont\color{black!65}(135.8)}\end{tabular} \\
WINO & $1\times256$ & \emptycell & \emptycell & \emptycell & \emptycell & \begin{tabular}[t]{@{}c@{}}\scorecell{0.67343056}{53.68}{}\\[-2pt]{\fontsize{7.5}{9}\selectfont\color{black!65}(45.2)}\end{tabular} & \begin{tabular}[t]{@{}c@{}}\scorecell{0.66603951}{53.68}{}\\[-2pt]{\fontsize{7.5}{9}\selectfont\color{black!65}(45.2)}\end{tabular} & \begin{tabular}[t]{@{}c@{}}\scorecell{0.65073529}{53.68}{}\\[-2pt]{\fontsize{7.5}{9}\selectfont\color{black!65}(45.2)}\end{tabular} \\
WINO & $2\times128$ & \emptycell & \emptycell & \emptycell & \emptycell & \begin{tabular}[t]{@{}c@{}}\scorecell{0.94451490}{75.28}{}\\[-2pt]{\fontsize{7.5}{9}\selectfont\color{black!65}(46.6)}\end{tabular} & \begin{tabular}[t]{@{}c@{}}\scorecell{0.93414864}{75.28}{}\\[-2pt]{\fontsize{7.5}{9}\selectfont\color{black!65}(46.6)}\end{tabular} & \begin{tabular}[t]{@{}c@{}}\scorecell{0.91268382}{75.28}{}\\[-2pt]{\fontsize{7.5}{9}\selectfont\color{black!65}(46.6)}\end{tabular} \\
KLASS & $1\times256$ & \emptycell & \emptycell & \emptycell & \emptycell & \emptycell & \begin{tabular}[t]{@{}c@{}}\scorecell{0.54750706}{44.12}{}\\[-2pt]{\fontsize{7.5}{9}\selectfont\color{black!65}(98.7)}\end{tabular} & \begin{tabular}[t]{@{}c@{}}\scorecell{0.53492647}{44.12}{}\\[-2pt]{\fontsize{7.5}{9}\selectfont\color{black!65}(98.7)}\end{tabular} \\
KLASS & $4\times64$ & \emptycell & \emptycell & \emptycell & \emptycell & \emptycell & \begin{tabular}[t]{@{}c@{}}\scorecell{0.99811853}{80.44}{}\\[-2pt]{\fontsize{7.5}{9}\selectfont\color{black!65}(112.0)}\end{tabular} & \begin{tabular}[t]{@{}c@{}}\scorecell{0.99172794}{81.80}{}\\[-2pt]{\fontsize{7.5}{9}\selectfont\color{black!65}(158.2)}\end{tabular} \\
CORE & $1\times256$ & \begin{tabular}[t]{@{}c@{}}\scorecell{0.11538462}{4.17}{}\\[-2pt]{\fontsize{7.5}{9}\selectfont\color{black!65}(4.0)}\end{tabular} & \begin{tabular}[t]{@{}c@{}}\scorecell{0.24469590}{13.12}{}\\[-2pt]{\fontsize{7.5}{9}\selectfont\color{black!65}(8.0)}\end{tabular} & \begin{tabular}[t]{@{}c@{}}\scorecell{0.53084983}{34.57}{}\\[-2pt]{\fontsize{7.5}{9}\selectfont\color{black!65}(16.0)}\end{tabular} & \begin{tabular}[t]{@{}c@{}}\scorecell{0.55918084}{40.71}{}\\[-2pt]{\fontsize{7.5}{9}\selectfont\color{black!65}(32.0)}\end{tabular} & \begin{tabular}[t]{@{}c@{}}\scorecell{0.51077996}{40.71}{}\\[-2pt]{\fontsize{7.5}{9}\selectfont\color{black!65}(32.0)}\end{tabular} & \begin{tabular}[t]{@{}c@{}}\scorecell{0.50517404}{40.71}{}\\[-2pt]{\fontsize{7.5}{9}\selectfont\color{black!65}(32.0)}\end{tabular} & \begin{tabular}[t]{@{}c@{}}\scorecell{0.49356618}{40.71}{}\\[-2pt]{\fontsize{7.5}{9}\selectfont\color{black!65}(32.0)}\end{tabular} \\
DAPD-Direct & $1\times256$ & \emptycell & \emptycell & \emptycell & \begin{tabular}[t]{@{}c@{}}\scorecell{0.75702881}{55.12}{}\\[-2pt]{\fontsize{7.5}{9}\selectfont\color{black!65}(24.5)}\end{tabular} & \begin{tabular}[t]{@{}c@{}}\scorecell{0.69150285}{55.12}{}\\[-2pt]{\fontsize{7.5}{9}\selectfont\color{black!65}(24.5)}\end{tabular} & \begin{tabular}[t]{@{}c@{}}\scorecell{0.68391345}{55.12}{}\\[-2pt]{\fontsize{7.5}{9}\selectfont\color{black!65}(24.5)}\end{tabular} & \begin{tabular}[t]{@{}c@{}}\scorecell{0.66819853}{55.12}{}\\[-2pt]{\fontsize{7.5}{9}\selectfont\color{black!65}(24.5)}\end{tabular} \\
DAPD-Direct & $4\times64$ & \emptycell & \emptycell & \emptycell & \emptycell & \begin{tabular}[t]{@{}c@{}}\scorecell{0.89980977}{71.72}{}\\[-2pt]{\fontsize{7.5}{9}\selectfont\color{black!65}(52.6)}\end{tabular} & \begin{tabular}[t]{@{}c@{}}\scorecell{0.88993415}{71.72}{}\\[-2pt]{\fontsize{7.5}{9}\selectfont\color{black!65}(52.6)}\end{tabular} & \begin{tabular}[t]{@{}c@{}}\scorecell{0.86948529}{71.72}{}\\[-2pt]{\fontsize{7.5}{9}\selectfont\color{black!65}(52.6)}\end{tabular} \\
DAPD-Staged & $1\times256$ & \emptycell & \emptycell & \emptycell & \emptycell & \begin{tabular}[t]{@{}c@{}}\scorecell{0.74571972}{59.44}{}\\[-2pt]{\fontsize{7.5}{9}\selectfont\color{black!65}(32.9)}\end{tabular} & \begin{tabular}[t]{@{}c@{}}\scorecell{0.73753528}{59.44}{}\\[-2pt]{\fontsize{7.5}{9}\selectfont\color{black!65}(32.9)}\end{tabular} & \begin{tabular}[t]{@{}c@{}}\scorecell{0.72058824}{59.44}{}\\[-2pt]{\fontsize{7.5}{9}\selectfont\color{black!65}(32.9)}\end{tabular} \\
DAPD-Staged & $4\times64$ & \emptycell & \emptycell & \emptycell & \emptycell & \emptycell & \begin{tabular}[t]{@{}c@{}}\scorecell{0.90216369}{72.71}{}\\[-2pt]{\fontsize{7.5}{9}\selectfont\color{black!65}(92.2)}\end{tabular} & \begin{tabular}[t]{@{}c@{}}\scorecell{0.88143382}{72.71}{}\\[-2pt]{\fontsize{7.5}{9}\selectfont\color{black!65}(92.2)}\end{tabular} \\
\midrule
\multicolumn{9}{l}{\textit{Published evaluation protocols}}\\
DCoLT (frozen) & NR & \emptycell & \emptycell & \emptycell & \emptycell & \scorecell{0.00000000}{69.24}{\textsuperscript{D}} & \emptycell & \emptycell \\
SAS (5-shot) & $1\times256$ & \emptycell & \emptycell & \emptycell & \emptycell & \emptycell & \emptycell & \scorecell{0.00000000}{76.00}{\textsuperscript{S}} \\
SAS (5-shot) & $4\times64$ & \emptycell & \emptycell & \emptycell & \scorecell{0.00000000}{58.00}{\textsuperscript{S}} & \scorecell{0.00000000}{61.00}{\textsuperscript{S}} & \emptycell & \scorecell{0.00000000}{81.00}{\textsuperscript{S}} \\
SAS (5-shot) & $8\times32$ & \emptycell & \emptycell & \emptycell & \emptycell & \emptycell & \emptycell & \scorecell{0.00000000}{80.00}{\textsuperscript{S}} \\
\midrule
\multicolumn{9}{l}{\textit{Separate reference: shorter answers}}\\
UPO (128-token answers) & $4\times32$ & \emptycell & \emptycell & \emptycell & \emptycell & \emptycell & \scorecell{0.00000000}{70.30}{\textsuperscript{U}} & \emptycell \\
\end{longtable}
\endgroup

\clearpage
\begingroup
\definecolor{scorebar}{HTML}{D9ECF5}
\ifdefined\scorecellwidth\else\newlength{\scorecellwidth}\fi
\setlength{\scorecellwidth}{36pt}
\newcommand{\scorecell}[3]{%
 \makebox[\scorecellwidth][l]{%
  \rlap{\raisebox{-2.2pt}[0pt][0pt]{\color{scorebar}\rule{#1\scorecellwidth}{10pt}}}%
  \makebox[30pt][r]{#2}\makebox[6pt][l]{#3}}}
\newcommand{\emptycell}{\makebox[\scorecellwidth][c]{\textemdash}}

\fontsize{9.5}{10.5}\selectfont
\setlength{\LTcapwidth}{\textwidth}
\setlength{\LTpre}{3pt}
\setlength{\tabcolsep}{2.5pt}
\renewcommand{\arraystretch}{1.0}
\begin{longtable}{@{}>{\raggedright\arraybackslash}p{4.7cm} c *{7}{r}@{}}
\caption{MATH500 comparison. Accuracy (\%) under mean-NFE budgets. Adaptive rows report the best observed test score; actual mean NFE is in parentheses. Bars scale from zero to the column maximum; bold marks the highest score. Superscripts identify results from Jazbec et al.\textsuperscript{A}, UPO\textsuperscript{U}, and SAS\textsuperscript{S}. Dashes denote unmeasured budgets or no eligible operating point.}\label{tab:baseline-math500}\\
\toprule
 & & \multicolumn{7}{c}{Mean-NFE budget}\\
Method & Blocks & \makebox[36pt][c]{4} & \makebox[36pt][c]{8} & \makebox[36pt][c]{16} & \makebox[36pt][c]{32} & \makebox[36pt][c]{64} & \makebox[36pt][c]{128} & \makebox[36pt][c]{256} \\
\midrule
\endfirsthead
\multicolumn{9}{l}{\textit{MATH500 comparison (continued)}}\\
\toprule
 & & \multicolumn{7}{c}{Mean-NFE budget}\\
Method & Blocks & \makebox[36pt][c]{4} & \makebox[36pt][c]{8} & \makebox[36pt][c]{16} & \makebox[36pt][c]{32} & \makebox[36pt][c]{64} & \makebox[36pt][c]{128} & \makebox[36pt][c]{256} \\
\midrule
\endhead
\bottomrule
\endfoot
\multicolumn{9}{l}{\textit{Fixed-budget decoding}}\\
\textbf{ReMaRL (ours)} & $1\times256$ & \scorecell{1.00000000}{\textbf{17.07}}{} & \scorecell{1.00000000}{\textbf{19.00}}{} & \scorecell{1.00000000}{\textbf{24.87}}{} & \scorecell{1.00000000}{\textbf{27.27}}{} & \scorecell{0.98624339}{31.07}{} & \emptycell & \emptycell \\
ReMaRL (ours), trained at NFE64 & $1\times256$ & \emptycell & \emptycell & \emptycell & \emptycell & \scorecell{0.98624339}{31.07}{} & \scorecell{0.94095238}{32.93}{} & \scorecell{0.91391941}{33.27}{} \\
P2$'$ (ours, $\tau=0.25$) & $1\times256$ & \scorecell{0.80453037}{13.73}{} & \scorecell{0.94035088}{17.87}{} & \scorecell{0.84171023}{20.93}{} & \scorecell{0.82874954}{22.60}{} & \scorecell{0.78306878}{24.67}{} & \scorecell{0.75047619}{26.27}{} & \scorecell{0.71611722}{26.07}{} \\
P2$'$ (ours, $\tau=1$) & $1\times256$ & \scorecell{0.64843750}{11.07}{} & \scorecell{0.72280702}{13.73}{} & \scorecell{0.71313673}{17.73}{} & \scorecell{0.77995110}{21.27}{} & \scorecell{0.77460317}{24.40}{} & \scorecell{0.77333333}{27.07}{} & \scorecell{0.74725275}{27.20}{} \\
P2$'$ (ours, greedy) & $1\times256$ & \scorecell{0.63281250}{10.80}{} & \scorecell{0.71578947}{13.60}{} & \scorecell{0.76407507}{19.00}{} & \scorecell{0.76283619}{20.80}{} & \scorecell{0.64126984}{20.20}{} & \scorecell{0.59428571}{20.80}{} & \scorecell{0.64285714}{23.40}{} \\
P2 (original self-planner) & $1\times256$ & \scorecell{0.42187500}{7.20}{} & \scorecell{0.52631579}{10.00}{} & \scorecell{0.69973190}{17.40}{} & \scorecell{0.68948655}{18.80}{} & \scorecell{0.69206349}{21.80}{} & \scorecell{0.56000000}{19.60}{} & \scorecell{0.60989011}{22.20}{} \\
Confidence & $1\times256$ & \scorecell{0.37500000}{6.40}{} & \scorecell{0.44210526}{8.40}{} & \scorecell{0.67560322}{16.80}{} & \scorecell{0.68948655}{18.80}{} & \scorecell{0.60952381}{19.20}{} & \scorecell{0.54857143}{19.20}{} & \scorecell{0.50549451}{18.40}{} \\
Confidence & $8\times32$ & \emptycell & \scorecell{0.40000000}{7.60}{} & \scorecell{0.68364611}{17.00}{} & \scorecell{0.77750611}{21.20}{} & \scorecell{0.92698413}{29.20}{} & \scorecell{0.91428571}{32.00}{} & \scorecell{0.96703297}{35.20}{} \\
Random & $1\times256$ & \scorecell{0.51953125}{8.87}{} & \scorecell{0.61403509}{11.67}{} & \scorecell{0.53887399}{13.40}{} & \scorecell{0.63569682}{17.33}{} & \scorecell{0.51216931}{16.13}{} & \scorecell{0.54285714}{19.00}{} & \scorecell{0.54029304}{19.67}{} \\
Random & $8\times32$ & \emptycell & \scorecell{0.40000000}{7.60}{} & \scorecell{0.42359249}{10.53}{} & \scorecell{0.47921760}{13.07}{} & \scorecell{0.53333333}{16.80}{} & \scorecell{0.66476190}{23.27}{} & \scorecell{0.76007326}{27.67}{} \\
Margin & $1\times256$ & \scorecell{0.55078125}{9.40}{} & \scorecell{0.72631579}{13.80}{} & \scorecell{0.67560322}{16.80}{} & \scorecell{0.74816626}{20.40}{} & \scorecell{0.66666667}{21.00}{} & \scorecell{0.52571429}{18.40}{} & \scorecell{0.57692308}{21.00}{} \\
Margin & $8\times32$ & \emptycell & \scorecell{0.40000000}{7.60}{} & \scorecell{0.52278820}{13.00}{} & \scorecell{0.78484108}{21.40}{} & \scorecell{0.93968254}{29.60}{} & \scorecell{0.88000000}{30.80}{} & \scorecell{0.97802198}{35.60}{} \\
ReMDM (UNCODE) & $1\times256$ & \emptycell & \emptycell & \scorecell{0.69973190}{17.40}{} & \scorecell{0.71882641}{19.60}{} & \scorecell{0.66666667}{21.00}{} & \scorecell{0.58285714}{20.40}{} & \scorecell{0.52747253}{19.20}{} \\
UNCODE/PC-Sampler & $1\times256$ & \scorecell{0.35156250}{6.00}{} & \scorecell{0.33684211}{6.40}{} & \scorecell{0.53083110}{13.20}{} & \scorecell{0.59413203}{16.20}{} & \scorecell{0.79365079}{25.00}{} & \scorecell{0.94857143}{33.20}{} & \scorecell{1.00000000}{\textbf{36.40}}{} \\
RCR & $1\times256$ & \scorecell{0.44531250}{7.60}{} & \scorecell{0.74736842}{14.20}{} & \scorecell{0.68364611}{17.00}{} & \scorecell{0.74083130}{20.20}{} & \scorecell{0.65396825}{20.60}{} & \scorecell{0.56000000}{19.60}{} & \scorecell{0.60989011}{22.20}{} \\
\midrule
\multicolumn{9}{l}{\textit{Adaptive decoding}}\\
Fast-dLLM & $1\times256$ & \begin{tabular}[t]{@{}c@{}}\scorecell{0.35156250}{6.00}{}\\[-2pt]{\fontsize{7.5}{9}\selectfont\color{black!65}(2.9)}\end{tabular} & \begin{tabular}[t]{@{}c@{}}\scorecell{0.31578947}{6.00}{}\\[-2pt]{\fontsize{7.5}{9}\selectfont\color{black!65}(2.9)}\end{tabular} & \begin{tabular}[t]{@{}c@{}}\scorecell{0.24128686}{6.00}{}\\[-2pt]{\fontsize{7.5}{9}\selectfont\color{black!65}(2.9)}\end{tabular} & \begin{tabular}[t]{@{}c@{}}\scorecell{0.50611247}{13.80}{}\\[-2pt]{\fontsize{7.5}{9}\selectfont\color{black!65}(17.1)}\end{tabular} & \begin{tabular}[t]{@{}c@{}}\scorecell{0.56507937}{17.80}{}\\[-2pt]{\fontsize{7.5}{9}\selectfont\color{black!65}(36.8)}\end{tabular} & \begin{tabular}[t]{@{}c@{}}\scorecell{0.57142857}{20.00}{}\\[-2pt]{\fontsize{7.5}{9}\selectfont\color{black!65}(64.2)}\end{tabular} & \begin{tabular}[t]{@{}c@{}}\scorecell{0.54945055}{20.00}{}\\[-2pt]{\fontsize{7.5}{9}\selectfont\color{black!65}(64.2)}\end{tabular} \\
Fast-dLLM & $8\times32$ & \emptycell & \emptycell & \begin{tabular}[t]{@{}c@{}}\scorecell{0.36997319}{9.20}{}\\[-2pt]{\fontsize{7.5}{9}\selectfont\color{black!65}(11.3)}\end{tabular} & \begin{tabular}[t]{@{}c@{}}\scorecell{0.33740831}{9.20}{}\\[-2pt]{\fontsize{7.5}{9}\selectfont\color{black!65}(11.3)}\end{tabular} & \begin{tabular}[t]{@{}c@{}}\scorecell{0.84444444}{26.60}{}\\[-2pt]{\fontsize{7.5}{9}\selectfont\color{black!65}(49.6)}\end{tabular} & \begin{tabular}[t]{@{}c@{}}\scorecell{1.00000000}{\textbf{35.00}}{}\\[-2pt]{\fontsize{7.5}{9}\selectfont\color{black!65}(101.3)}\end{tabular} & \begin{tabular}[t]{@{}c@{}}\scorecell{0.97252747}{35.40}{}\\[-2pt]{\fontsize{7.5}{9}\selectfont\color{black!65}(136.9)}\end{tabular} \\
EB-Sampler & $1\times256$ & \emptycell & \emptycell & \emptycell & \begin{tabular}[t]{@{}c@{}}\scorecell{0.74083130}{20.20}{}\\[-2pt]{\fontsize{7.5}{9}\selectfont\color{black!65}(26.5)}\end{tabular} & \begin{tabular}[t]{@{}c@{}}\scorecell{0.64126984}{20.20}{}\\[-2pt]{\fontsize{7.5}{9}\selectfont\color{black!65}(26.5)}\end{tabular} & \begin{tabular}[t]{@{}c@{}}\scorecell{0.62857143}{22.00}{}\\[-2pt]{\fontsize{7.5}{9}\selectfont\color{black!65}(123.0)}\end{tabular} & \begin{tabular}[t]{@{}c@{}}\scorecell{0.60439560}{22.00}{}\\[-2pt]{\fontsize{7.5}{9}\selectfont\color{black!65}(123.0)}\end{tabular} \\
EB-Sampler & $8\times32$ & \emptycell & \emptycell & \emptycell & \emptycell & \begin{tabular}[t]{@{}c@{}}\scorecell{0.55873016}{17.60}{}\\[-2pt]{\fontsize{7.5}{9}\selectfont\color{black!65}(34.4)}\end{tabular} & \begin{tabular}[t]{@{}c@{}}\scorecell{0.94857143}{33.20}{}\\[-2pt]{\fontsize{7.5}{9}\selectfont\color{black!65}(68.0)}\end{tabular} & \begin{tabular}[t]{@{}c@{}}\scorecell{0.96703297}{35.20}{}\\[-2pt]{\fontsize{7.5}{9}\selectfont\color{black!65}(164.7)}\end{tabular} \\
Jazbec et al. & $1\times256$ & \emptycell & \emptycell & \begin{tabular}[t]{@{}c@{}}\scorecell{0.75603217}{18.80}{\textsuperscript{A}}\\[-2pt]{\fontsize{7.5}{9}\selectfont\color{black!65}(14.5)}\end{tabular} & \begin{tabular}[t]{@{}c@{}}\scorecell{0.72616137}{19.80}{\textsuperscript{A}}\\[-2pt]{\fontsize{7.5}{9}\selectfont\color{black!65}(22.3)}\end{tabular} & \begin{tabular}[t]{@{}c@{}}\scorecell{0.66666667}{21.00}{\textsuperscript{A}}\\[-2pt]{\fontsize{7.5}{9}\selectfont\color{black!65}(61.6)}\end{tabular} & \begin{tabular}[t]{@{}c@{}}\scorecell{0.65428571}{22.90}{\textsuperscript{A}}\\[-2pt]{\fontsize{7.5}{9}\selectfont\color{black!65}(121.5)}\end{tabular} & \begin{tabular}[t]{@{}c@{}}\scorecell{0.62912088}{22.90}{\textsuperscript{A}}\\[-2pt]{\fontsize{7.5}{9}\selectfont\color{black!65}(121.5)}\end{tabular} \\
Jazbec et al. + expert steering & $1\times256$ & \emptycell & \emptycell & \emptycell & \begin{tabular}[t]{@{}c@{}}\scorecell{0.72616137}{19.80}{\textsuperscript{A}}\\[-2pt]{\fontsize{7.5}{9}\selectfont\color{black!65}(23.1)}\end{tabular} & \begin{tabular}[t]{@{}c@{}}\scorecell{0.62857143}{19.80}{\textsuperscript{A}}\\[-2pt]{\fontsize{7.5}{9}\selectfont\color{black!65}(23.1)}\end{tabular} & \begin{tabular}[t]{@{}c@{}}\scorecell{0.89714286}{31.40}{\textsuperscript{A}}\\[-2pt]{\fontsize{7.5}{9}\selectfont\color{black!65}(86.7)}\end{tabular} & \begin{tabular}[t]{@{}c@{}}\scorecell{0.86263736}{31.40}{\textsuperscript{A}}\\[-2pt]{\fontsize{7.5}{9}\selectfont\color{black!65}(86.7)}\end{tabular} \\
Jazbec et al. ($\alpha$ sweep) & $8\times32$ & \emptycell & \emptycell & \begin{tabular}[t]{@{}c@{}}\scorecell{0.61528150}{15.30}{\textsuperscript{A}}\\[-2pt]{\fontsize{7.5}{9}\selectfont\color{black!65}(10.8)}\end{tabular} & \begin{tabular}[t]{@{}c@{}}\scorecell{0.56112469}{15.30}{\textsuperscript{A}}\\[-2pt]{\fontsize{7.5}{9}\selectfont\color{black!65}(10.8)}\end{tabular} & \begin{tabular}[t]{@{}c@{}}\scorecell{1.00000000}{\textbf{31.50}}{\textsuperscript{A}}\\[-2pt]{\fontsize{7.5}{9}\selectfont\color{black!65}(59.1)}\end{tabular} & \begin{tabular}[t]{@{}c@{}}\scorecell{0.97428571}{34.10}{\textsuperscript{A}}\\[-2pt]{\fontsize{7.5}{9}\selectfont\color{black!65}(100.2)}\end{tabular} & \begin{tabular}[t]{@{}c@{}}\scorecell{0.96428571}{35.10}{\textsuperscript{A}}\\[-2pt]{\fontsize{7.5}{9}\selectfont\color{black!65}(153.4)}\end{tabular} \\
WINO & $1\times256$ & \emptycell & \emptycell & \emptycell & \emptycell & \emptycell & \begin{tabular}[t]{@{}c@{}}\scorecell{0.69714286}{24.40}{}\\[-2pt]{\fontsize{7.5}{9}\selectfont\color{black!65}(71.3)}\end{tabular} & \begin{tabular}[t]{@{}c@{}}\scorecell{0.67032967}{24.40}{}\\[-2pt]{\fontsize{7.5}{9}\selectfont\color{black!65}(71.3)}\end{tabular} \\
WINO & $2\times128$ & \emptycell & \emptycell & \emptycell & \emptycell & \begin{tabular}[t]{@{}c@{}}\scorecell{0.99047619}{31.20}{}\\[-2pt]{\fontsize{7.5}{9}\selectfont\color{black!65}(63.1)}\end{tabular} & \begin{tabular}[t]{@{}c@{}}\scorecell{0.89142857}{31.20}{}\\[-2pt]{\fontsize{7.5}{9}\selectfont\color{black!65}(63.1)}\end{tabular} & \begin{tabular}[t]{@{}c@{}}\scorecell{0.85714286}{31.20}{}\\[-2pt]{\fontsize{7.5}{9}\selectfont\color{black!65}(63.1)}\end{tabular} \\
KLASS & $1\times256$ & \emptycell & \emptycell & \emptycell & \emptycell & \emptycell & \begin{tabular}[t]{@{}c@{}}\scorecell{0.62857143}{22.00}{}\\[-2pt]{\fontsize{7.5}{9}\selectfont\color{black!65}(99.4)}\end{tabular} & \begin{tabular}[t]{@{}c@{}}\scorecell{0.60439560}{22.00}{}\\[-2pt]{\fontsize{7.5}{9}\selectfont\color{black!65}(99.4)}\end{tabular} \\
KLASS & $4\times64$ & \emptycell & \emptycell & \emptycell & \emptycell & \emptycell & \begin{tabular}[t]{@{}c@{}}\scorecell{0.98857143}{34.60}{}\\[-2pt]{\fontsize{7.5}{9}\selectfont\color{black!65}(126.9)}\end{tabular} & \begin{tabular}[t]{@{}c@{}}\scorecell{0.96703297}{35.20}{}\\[-2pt]{\fontsize{7.5}{9}\selectfont\color{black!65}(133.2)}\end{tabular} \\
CORE & $1\times256$ & \begin{tabular}[t]{@{}c@{}}\scorecell{0.35156250}{6.00}{}\\[-2pt]{\fontsize{7.5}{9}\selectfont\color{black!65}(4.0)}\end{tabular} & \begin{tabular}[t]{@{}c@{}}\scorecell{0.54736842}{10.40}{}\\[-2pt]{\fontsize{7.5}{9}\selectfont\color{black!65}(8.0)}\end{tabular} & \begin{tabular}[t]{@{}c@{}}\scorecell{0.67560322}{16.80}{}\\[-2pt]{\fontsize{7.5}{9}\selectfont\color{black!65}(16.0)}\end{tabular} & \begin{tabular}[t]{@{}c@{}}\scorecell{0.72616137}{19.80}{}\\[-2pt]{\fontsize{7.5}{9}\selectfont\color{black!65}(32.0)}\end{tabular} & \begin{tabular}[t]{@{}c@{}}\scorecell{0.64126984}{20.20}{}\\[-2pt]{\fontsize{7.5}{9}\selectfont\color{black!65}(64.0)}\end{tabular} & \begin{tabular}[t]{@{}c@{}}\scorecell{0.57714286}{20.20}{}\\[-2pt]{\fontsize{7.5}{9}\selectfont\color{black!65}(64.0)}\end{tabular} & \begin{tabular}[t]{@{}c@{}}\scorecell{0.55494505}{20.20}{}\\[-2pt]{\fontsize{7.5}{9}\selectfont\color{black!65}(64.0)}\end{tabular} \\
DAPD-Direct & $1\times256$ & \emptycell & \emptycell & \emptycell & \begin{tabular}[t]{@{}c@{}}\scorecell{0.83618582}{22.80}{}\\[-2pt]{\fontsize{7.5}{9}\selectfont\color{black!65}(30.3)}\end{tabular} & \begin{tabular}[t]{@{}c@{}}\scorecell{0.72380952}{22.80}{}\\[-2pt]{\fontsize{7.5}{9}\selectfont\color{black!65}(30.3)}\end{tabular} & \begin{tabular}[t]{@{}c@{}}\scorecell{0.65142857}{22.80}{}\\[-2pt]{\fontsize{7.5}{9}\selectfont\color{black!65}(30.3)}\end{tabular} & \begin{tabular}[t]{@{}c@{}}\scorecell{0.62637363}{22.80}{}\\[-2pt]{\fontsize{7.5}{9}\selectfont\color{black!65}(30.3)}\end{tabular} \\
DAPD-Direct & $4\times64$ & \emptycell & \emptycell & \emptycell & \emptycell & \begin{tabular}[t]{@{}c@{}}\scorecell{0.96507937}{30.40}{}\\[-2pt]{\fontsize{7.5}{9}\selectfont\color{black!65}(63.9)}\end{tabular} & \begin{tabular}[t]{@{}c@{}}\scorecell{0.86857143}{30.40}{}\\[-2pt]{\fontsize{7.5}{9}\selectfont\color{black!65}(63.9)}\end{tabular} & \begin{tabular}[t]{@{}c@{}}\scorecell{0.83516484}{30.40}{}\\[-2pt]{\fontsize{7.5}{9}\selectfont\color{black!65}(63.9)}\end{tabular} \\
DAPD-Staged & $1\times256$ & \emptycell & \emptycell & \emptycell & \emptycell & \begin{tabular}[t]{@{}c@{}}\scorecell{0.76825397}{24.20}{}\\[-2pt]{\fontsize{7.5}{9}\selectfont\color{black!65}(36.8)}\end{tabular} & \begin{tabular}[t]{@{}c@{}}\scorecell{0.69142857}{24.20}{}\\[-2pt]{\fontsize{7.5}{9}\selectfont\color{black!65}(36.8)}\end{tabular} & \begin{tabular}[t]{@{}c@{}}\scorecell{0.66483516}{24.20}{}\\[-2pt]{\fontsize{7.5}{9}\selectfont\color{black!65}(36.8)}\end{tabular} \\
DAPD-Staged & $4\times64$ & \emptycell & \emptycell & \emptycell & \emptycell & \emptycell & \begin{tabular}[t]{@{}c@{}}\scorecell{0.84571429}{29.60}{}\\[-2pt]{\fontsize{7.5}{9}\selectfont\color{black!65}(99.1)}\end{tabular} & \begin{tabular}[t]{@{}c@{}}\scorecell{0.81318681}{29.60}{}\\[-2pt]{\fontsize{7.5}{9}\selectfont\color{black!65}(99.1)}\end{tabular} \\
\midrule
\multicolumn{9}{l}{\textit{SAS: published evaluation protocol}}\\
SAS & $1\times256$ & \emptycell & \emptycell & \emptycell & \emptycell & \emptycell & \emptycell & \scorecell{0.00000000}{24.67}{\textsuperscript{S}} \\
SAS & $8\times32$ & \emptycell & \emptycell & \emptycell & \emptycell & \emptycell & \emptycell & \scorecell{0.00000000}{30.00}{\textsuperscript{S}} \\
\midrule
\multicolumn{9}{l}{\textit{Separate reference: shorter answers}}\\
UPO (128-token answers) & $4\times32$ & \emptycell & \emptycell & \emptycell & \emptycell & \emptycell & \scorecell{0.00000000}{28.40}{\textsuperscript{U}} & \emptycell \\
\end{longtable}
\endgroup

\clearpage
\subsection{Code}

Tables~\ref{tab:baseline-humaneval} and~\ref{tab:baseline-mbpp} use
the same ReMaRL checkpoints as Figure~\ref{fig:code}. The row labeled “ReMaRL (ours), trained at NFE32” reuses one checkpoint at higher inference budgets. Jazbec et al.'s code-trained
and math-to-code transfer policies are reported separately.
For the ReMDM rows, we use UNCODE’s LLaDA adaptation~\citep{uncode} and count both model calls in each revision loop.

\begingroup
\definecolor{scorebar}{HTML}{D9ECF5}
\ifdefined\scorecellwidth\else\newlength{\scorecellwidth}\fi
\setlength{\scorecellwidth}{36pt}
\newcommand{\scorecell}[3]{%
 \makebox[\scorecellwidth][l]{%
  \rlap{\raisebox{-2.2pt}[0pt][0pt]{\color{scorebar}\rule{#1\scorecellwidth}{10pt}}}%
  \makebox[30pt][r]{#2}\makebox[6pt][l]{#3}}}
\newcommand{\emptycell}{\makebox[\scorecellwidth][c]{\textemdash}}

\fontsize{9.5}{10.5}\selectfont
\setlength{\LTcapwidth}{\textwidth}
\setlength{\LTpre}{3pt}
\setlength{\tabcolsep}{2.5pt}
\renewcommand{\arraystretch}{1.0}
\begin{longtable}{@{}>{\raggedright\arraybackslash}p{4.7cm} c *{6}{r}@{}}
\caption{HumanEval comparison. Pass@1 (\%) under mean-NFE budgets. Adaptive rows report the best observed test score; actual mean NFE is in parentheses. Bars scale from zero to the column maximum; bold marks the highest score. Scores marked \textsuperscript{A} are from Jazbec et al. Dashes denote unmeasured budgets or no eligible operating point.}\label{tab:baseline-humaneval}\\
\toprule
 & & \multicolumn{6}{c}{Mean-NFE budget}\\
Method & Blocks & \makebox[36pt][c]{8} & \makebox[36pt][c]{16} & \makebox[36pt][c]{32} & \makebox[36pt][c]{64} & \makebox[36pt][c]{128} & \makebox[36pt][c]{256} \\
\midrule
\endfirsthead
\multicolumn{8}{l}{\textit{HumanEval comparison (continued)}}\\
\toprule
 & & \multicolumn{6}{c}{Mean-NFE budget}\\
Method & Blocks & \makebox[36pt][c]{8} & \makebox[36pt][c]{16} & \makebox[36pt][c]{32} & \makebox[36pt][c]{64} & \makebox[36pt][c]{128} & \makebox[36pt][c]{256} \\
\midrule
\endhead
\bottomrule
\endfoot
\multicolumn{8}{l}{\textit{Fixed-budget decoding}}\\
\textbf{ReMaRL (ours)} & $1\times256$ & \scorecell{1.00000000}{\textbf{20.73}}{} & \scorecell{1.00000000}{\textbf{29.47}}{} & \scorecell{1.00000000}{\textbf{31.71}}{} & \emptycell & \emptycell & \emptycell \\
ReMaRL (ours), trained at NFE32 & $1\times256$ & \emptycell & \emptycell & \scorecell{1.00000000}{\textbf{31.71}}{} & \scorecell{1.00000000}{\textbf{38.41}}{} & \scorecell{0.87962963}{38.62}{} & \scorecell{0.88127854}{39.23}{} \\
P2$'$ (ours, $\tau=0.25$) & $1\times256$ & \scorecell{0.68633103}{14.23}{} & \scorecell{0.76555737}{22.56}{} & \scorecell{0.76275597}{24.19}{} & \scorecell{0.60324738}{23.17}{} & \scorecell{0.57410597}{25.20}{} & \scorecell{0.51600718}{22.97}{} \\
P2$'$ (ours, $\tau=1$) & $1\times256$ & \scorecell{0.55882353}{11.59}{} & \scorecell{0.50344828}{14.84}{} & \scorecell{0.62820513}{19.92}{} & \scorecell{0.68253968}{26.22}{} & \scorecell{0.50925926}{22.36}{} & \scorecell{0.55707763}{24.80}{} \\
P2$'$ (ours, greedy) & $1\times256$ & \scorecell{0.67647059}{14.02}{} & \scorecell{0.60000000}{17.68}{} & \scorecell{0.69230769}{21.95}{} & \scorecell{0.80952381}{31.10}{} & \scorecell{0.79166667}{34.76}{} & \scorecell{0.95890411}{42.68}{} \\
P2 (original self-planner) & $1\times256$ & \scorecell{0.73529412}{15.24}{} & \scorecell{0.64137931}{18.90}{} & \scorecell{0.69230769}{21.95}{} & \scorecell{0.82539683}{31.71}{} & \scorecell{0.87500000}{38.41}{} & \scorecell{0.98630137}{43.90}{} \\
Confidence & $1\times256$ & \scorecell{0.64705882}{13.41}{} & \scorecell{0.57931034}{17.07}{} & \scorecell{0.48076923}{15.24}{} & \scorecell{0.61904762}{23.78}{} & \scorecell{0.77777778}{34.15}{} & \scorecell{1.00000000}{\textbf{44.51}}{} \\
Confidence & $8\times32$ & \scorecell{0.14705882}{3.05}{} & \scorecell{0.12413793}{3.66}{} & \scorecell{0.32692308}{10.37}{} & \scorecell{0.53968254}{20.73}{} & \scorecell{0.84722222}{37.20}{} & \scorecell{0.98630137}{43.90}{} \\
Random & $8\times32$ & \scorecell{0.14705882}{3.05}{} & \scorecell{0.17241379}{5.08}{} & \scorecell{0.20512821}{6.50}{} & \scorecell{0.28571429}{10.98}{} & \scorecell{0.36111111}{15.85}{} & \scorecell{0.48858447}{21.75}{} \\
ReMDM (UNCODE) & $1\times256$ & \emptycell & \scorecell{0.47586207}{14.02}{} & \scorecell{0.59615385}{18.90}{} & \scorecell{0.44444444}{17.07}{} & \scorecell{0.56944444}{25.00}{} & \scorecell{0.75342466}{33.54}{} \\
UNCODE/PC-Sampler & $1\times256$ & \scorecell{0.14705882}{3.05}{} & \scorecell{0.12413793}{3.66}{} & \scorecell{0.23076923}{7.32}{} & \scorecell{0.44444444}{17.07}{} & \scorecell{0.76388889}{33.54}{} & \scorecell{0.94520548}{42.07}{} \\
RCR & $1\times256$ & \scorecell{0.76470588}{15.85}{} & \scorecell{0.66206897}{19.51}{} & \scorecell{0.80769231}{25.61}{} & \scorecell{0.82539683}{31.71}{} & \scorecell{0.94444444}{41.46}{} & \scorecell{0.93150685}{41.46}{} \\
\midrule
\multicolumn{8}{l}{\textit{Adaptive decoding}}\\
Fast-dLLM & $1\times256$ & \emptycell & \emptycell & \begin{tabular}[t]{@{}c@{}}\scorecell{0.28846154}{9.15}{}\\[-2pt]{\fontsize{7.5}{9}\selectfont\color{black!65}(19.8)}\end{tabular} & \begin{tabular}[t]{@{}c@{}}\scorecell{0.61904762}{23.78}{}\\[-2pt]{\fontsize{7.5}{9}\selectfont\color{black!65}(47.3)}\end{tabular} & \begin{tabular}[t]{@{}c@{}}\scorecell{1.00000000}{\textbf{43.90}}{}\\[-2pt]{\fontsize{7.5}{9}\selectfont\color{black!65}(117.7)}\end{tabular} & \begin{tabular}[t]{@{}c@{}}\scorecell{1.00000000}{\textbf{44.51}}{}\\[-2pt]{\fontsize{7.5}{9}\selectfont\color{black!65}(134.9)}\end{tabular} \\
Fast-dLLM & $8\times32$ & \emptycell & \begin{tabular}[t]{@{}c@{}}\scorecell{0.10344828}{3.05}{}\\[-2pt]{\fontsize{7.5}{9}\selectfont\color{black!65}(8.9)}\end{tabular} & \begin{tabular}[t]{@{}c@{}}\scorecell{0.19230769}{6.10}{}\\[-2pt]{\fontsize{7.5}{9}\selectfont\color{black!65}(30.2)}\end{tabular} & \begin{tabular}[t]{@{}c@{}}\scorecell{0.63492063}{24.39}{}\\[-2pt]{\fontsize{7.5}{9}\selectfont\color{black!65}(52.5)}\end{tabular} & \begin{tabular}[t]{@{}c@{}}\scorecell{1.00000000}{\textbf{43.90}}{}\\[-2pt]{\fontsize{7.5}{9}\selectfont\color{black!65}(118.2)}\end{tabular} & \begin{tabular}[t]{@{}c@{}}\scorecell{0.98630137}{43.90}{}\\[-2pt]{\fontsize{7.5}{9}\selectfont\color{black!65}(118.2)}\end{tabular} \\
Jazbec et al. (math-trained) & $8\times32$ & \emptycell & \emptycell & \emptycell & \emptycell & \begin{tabular}[t]{@{}c@{}}\scorecell{0.91666666}{40.24}{\textsuperscript{A}}\\[-2pt]{\fontsize{7.5}{9}\selectfont\color{black!65}(113.5)}\end{tabular} & \begin{tabular}[t]{@{}c@{}}\scorecell{0.93150685}{41.46}{\textsuperscript{A}}\\[-2pt]{\fontsize{7.5}{9}\selectfont\color{black!65}(163.7)}\end{tabular} \\
Jazbec et al. (code-trained) & $8\times32$ & \emptycell & \emptycell & \emptycell & \begin{tabular}[t]{@{}c@{}}\scorecell{0.69841270}{26.83}{\textsuperscript{A}}\\[-2pt]{\fontsize{7.5}{9}\selectfont\color{black!65}(36.6)}\end{tabular} & \begin{tabular}[t]{@{}c@{}}\scorecell{0.95833333}{42.07}{\textsuperscript{A}}\\[-2pt]{\fontsize{7.5}{9}\selectfont\color{black!65}(76.1)}\end{tabular} & \begin{tabular}[t]{@{}c@{}}\scorecell{0.94520548}{42.07}{\textsuperscript{A}}\\[-2pt]{\fontsize{7.5}{9}\selectfont\color{black!65}(76.1)}\end{tabular} \\
WINO & $1\times256$ & \emptycell & \emptycell & \emptycell & \emptycell & \begin{tabular}[t]{@{}c@{}}\scorecell{0.81944444}{35.98}{}\\[-2pt]{\fontsize{7.5}{9}\selectfont\color{black!65}(88.0)}\end{tabular} & \begin{tabular}[t]{@{}c@{}}\scorecell{0.80821918}{35.98}{}\\[-2pt]{\fontsize{7.5}{9}\selectfont\color{black!65}(88.0)}\end{tabular} \\
WINO & $2\times128$ & \emptycell & \emptycell & \emptycell & \emptycell & \begin{tabular}[t]{@{}c@{}}\scorecell{0.95833333}{42.07}{}\\[-2pt]{\fontsize{7.5}{9}\selectfont\color{black!65}(79.9)}\end{tabular} & \begin{tabular}[t]{@{}c@{}}\scorecell{0.94520548}{42.07}{}\\[-2pt]{\fontsize{7.5}{9}\selectfont\color{black!65}(79.9)}\end{tabular} \\
CORE & $1\times256$ & \begin{tabular}[t]{@{}c@{}}\scorecell{0.67647059}{14.02}{}\\[-2pt]{\fontsize{7.5}{9}\selectfont\color{black!65}(8.0)}\end{tabular} & \begin{tabular}[t]{@{}c@{}}\scorecell{0.47586207}{14.02}{}\\[-2pt]{\fontsize{7.5}{9}\selectfont\color{black!65}(8.0)}\end{tabular} & \begin{tabular}[t]{@{}c@{}}\scorecell{0.50000000}{15.85}{}\\[-2pt]{\fontsize{7.5}{9}\selectfont\color{black!65}(32.0)}\end{tabular} & \begin{tabular}[t]{@{}c@{}}\scorecell{0.69841270}{26.83}{}\\[-2pt]{\fontsize{7.5}{9}\selectfont\color{black!65}(64.0)}\end{tabular} & \begin{tabular}[t]{@{}c@{}}\scorecell{0.77777778}{34.15}{}\\[-2pt]{\fontsize{7.5}{9}\selectfont\color{black!65}(128.0)}\end{tabular} & \begin{tabular}[t]{@{}c@{}}\scorecell{0.87671233}{39.02}{}\\[-2pt]{\fontsize{7.5}{9}\selectfont\color{black!65}(256.0)}\end{tabular} \\
DAPD-Direct & $1\times256$ & \emptycell & \emptycell & \emptycell & \begin{tabular}[t]{@{}c@{}}\scorecell{0.85714286}{32.93}{}\\[-2pt]{\fontsize{7.5}{9}\selectfont\color{black!65}(32.01)}\end{tabular} & \begin{tabular}[t]{@{}c@{}}\scorecell{0.75000000}{32.93}{}\\[-2pt]{\fontsize{7.5}{9}\selectfont\color{black!65}(32.01)}\end{tabular} & \begin{tabular}[t]{@{}c@{}}\scorecell{0.73972603}{32.93}{}\\[-2pt]{\fontsize{7.5}{9}\selectfont\color{black!65}(32.01)}\end{tabular} \\
DAPD-Direct & $4\times64$ & \emptycell & \emptycell & \emptycell & \emptycell & \begin{tabular}[t]{@{}c@{}}\scorecell{0.80555556}{35.37}{}\\[-2pt]{\fontsize{7.5}{9}\selectfont\color{black!65}(66.5)}\end{tabular} & \begin{tabular}[t]{@{}c@{}}\scorecell{0.79452055}{35.37}{}\\[-2pt]{\fontsize{7.5}{9}\selectfont\color{black!65}(66.5)}\end{tabular} \\
DAPD-Staged & $1\times256$ & \emptycell & \emptycell & \emptycell & \begin{tabular}[t]{@{}c@{}}\scorecell{0.90476190}{34.76}{}\\[-2pt]{\fontsize{7.5}{9}\selectfont\color{black!65}(37.4)}\end{tabular} & \begin{tabular}[t]{@{}c@{}}\scorecell{0.79166667}{34.76}{}\\[-2pt]{\fontsize{7.5}{9}\selectfont\color{black!65}(37.4)}\end{tabular} & \begin{tabular}[t]{@{}c@{}}\scorecell{0.78082192}{34.76}{}\\[-2pt]{\fontsize{7.5}{9}\selectfont\color{black!65}(37.4)}\end{tabular} \\
DAPD-Staged & $4\times64$ & \emptycell & \emptycell & \emptycell & \emptycell & \begin{tabular}[t]{@{}c@{}}\scorecell{0.83333333}{36.59}{}\\[-2pt]{\fontsize{7.5}{9}\selectfont\color{black!65}(99.5)}\end{tabular} & \begin{tabular}[t]{@{}c@{}}\scorecell{0.82191781}{36.59}{}\\[-2pt]{\fontsize{7.5}{9}\selectfont\color{black!65}(99.5)}\end{tabular} \\
\end{longtable}
\endgroup

\clearpage
\begingroup
\definecolor{scorebar}{HTML}{D9ECF5}
\ifdefined\scorecellwidth\else\newlength{\scorecellwidth}\fi
\setlength{\scorecellwidth}{36pt}
\newcommand{\scorecell}[3]{%
 \makebox[\scorecellwidth][l]{%
  \rlap{\raisebox{-2.2pt}[0pt][0pt]{\color{scorebar}\rule{#1\scorecellwidth}{10pt}}}%
  \makebox[30pt][r]{#2}\makebox[6pt][l]{#3}}}
\newcommand{\emptycell}{\makebox[\scorecellwidth][c]{\textemdash}}

\fontsize{9.5}{10.5}\selectfont
\setlength{\LTcapwidth}{\textwidth}
\setlength{\LTpre}{3pt}
\setlength{\tabcolsep}{2.5pt}
\renewcommand{\arraystretch}{1.0}
\begin{longtable}{@{}>{\raggedright\arraybackslash}p{4.7cm} c *{6}{r}@{}}
\caption{MBPP comparison. Pass@1 (\%) under mean-NFE budgets. Adaptive rows report the best observed test score; actual mean NFE is in parentheses. Bars scale from zero to the column maximum; bold marks the highest score. Superscripts identify results from Jazbec et al.\textsuperscript{A} and SAS\textsuperscript{S}. Dashes denote unmeasured budgets or no eligible operating point.}\label{tab:baseline-mbpp}\\
\toprule
 & & \multicolumn{6}{c}{Mean-NFE budget}\\
Method & Blocks & \makebox[36pt][c]{8} & \makebox[36pt][c]{16} & \makebox[36pt][c]{32} & \makebox[36pt][c]{64} & \makebox[36pt][c]{128} & \makebox[36pt][c]{256} \\
\midrule
\endfirsthead
\multicolumn{8}{l}{\textit{MBPP comparison (continued)}}\\
\toprule
 & & \multicolumn{6}{c}{Mean-NFE budget}\\
Method & Blocks & \makebox[36pt][c]{8} & \makebox[36pt][c]{16} & \makebox[36pt][c]{32} & \makebox[36pt][c]{64} & \makebox[36pt][c]{128} & \makebox[36pt][c]{256} \\
\midrule
\endhead
\bottomrule
\endfoot
\multicolumn{8}{l}{\textit{Fixed-budget decoding}}\\
\textbf{ReMaRL (ours)} & $1\times256$ & \scorecell{1.00000000}{\textbf{25.40}}{} & \scorecell{1.00000000}{\textbf{32.40}}{} & \scorecell{1.00000000}{\textbf{36.47}}{} & \emptycell & \emptycell & \emptycell \\
ReMaRL (ours), trained at NFE32 & $1\times256$ & \emptycell & \emptycell & \scorecell{1.00000000}{\textbf{36.47}}{} & \scorecell{0.99329983}{39.53}{} & \scorecell{1.00000000}{\textbf{41.33}}{} & \scorecell{1.00000000}{\textbf{43.53}}{} \\
P2$'$ (ours, $\tau=0.25$) & $1\times256$ & \scorecell{0.84514436}{21.47}{} & \scorecell{0.90534979}{29.33}{} & \scorecell{0.92313317}{33.67}{} & \scorecell{0.84757119}{33.73}{} & \scorecell{0.89523349}{37.00}{} & \scorecell{0.82088981}{35.73}{} \\
P2$'$ (ours, $\tau=1$) & $1\times256$ & \scorecell{0.61942257}{15.73}{} & \scorecell{0.72016461}{23.33}{} & \scorecell{0.76965265}{28.07}{} & \scorecell{0.81072027}{32.27}{} & \scorecell{0.78548387}{32.47}{} & \scorecell{0.82542113}{35.93}{} \\
P2$'$ (ours, greedy) & $1\times256$ & \scorecell{0.11811024}{3.00}{} & \scorecell{0.35185185}{11.40}{} & \scorecell{0.54296161}{19.80}{} & \scorecell{0.64321608}{25.60}{} & \scorecell{0.64838710}{26.80}{} & \scorecell{0.66156202}{28.80}{} \\
P2 (original self-planner) & $1\times256$ & \scorecell{0.11811024}{3.00}{} & \scorecell{0.31481481}{10.20}{} & \scorecell{0.48811700}{17.80}{} & \scorecell{0.64824121}{25.80}{} & \scorecell{0.71129032}{29.40}{} & \scorecell{0.66615620}{29.00}{} \\
Confidence & $1\times256$ & \scorecell{0.17322835}{4.40}{} & \scorecell{0.27160494}{8.80}{} & \scorecell{0.47714808}{17.40}{} & \scorecell{0.67336683}{26.80}{} & \scorecell{0.66290323}{27.40}{} & \scorecell{0.68912711}{30.00}{} \\
Confidence & $8\times32$ & \scorecell{0.03149606}{0.80}{} & \scorecell{0.17283951}{5.60}{} & \scorecell{0.37294333}{13.60}{} & \scorecell{0.50251256}{20.00}{} & \scorecell{0.71129032}{29.40}{} & \scorecell{0.92343032}{40.20}{} \\
Random & $8\times32$ & \scorecell{0.03149606}{0.80}{} & \scorecell{0.08436214}{2.73}{} & \scorecell{0.14990859}{5.47}{} & \scorecell{0.31658291}{12.60}{} & \scorecell{0.46451613}{19.20}{} & \scorecell{0.60183767}{26.20}{} \\
ReMDM (UNCODE) & $1\times256$ & \emptycell & \scorecell{0.15432099}{5.00}{} & \scorecell{0.27422303}{10.00}{} & \scorecell{0.42211055}{16.80}{} & \scorecell{0.61935484}{25.60}{} & \scorecell{0.61562021}{26.80}{} \\
UNCODE/PC-Sampler & $1\times256$ & \scorecell{0.03149606}{0.80}{} & \scorecell{0.08641975}{2.80}{} & \scorecell{0.27422303}{10.00}{} & \scorecell{0.52763819}{21.00}{} & \scorecell{0.74032258}{30.60}{} & \scorecell{0.93721286}{40.80}{} \\
RCR & $1\times256$ & \scorecell{0.14960630}{3.80}{} & \scorecell{0.31481481}{10.20}{} & \scorecell{0.59232176}{21.60}{} & \scorecell{0.70854271}{28.20}{} & \scorecell{0.75000000}{31.00}{} & \scorecell{0.69372129}{30.20}{} \\
\midrule
\multicolumn{8}{l}{\textit{Adaptive decoding}}\\
Fast-dLLM & $1\times256$ & \emptycell & \emptycell & \begin{tabular}[t]{@{}c@{}}\scorecell{0.66910420}{24.40}{}\\[-2pt]{\fontsize{7.5}{9}\selectfont\color{black!65}(22.4)}\end{tabular} & \begin{tabular}[t]{@{}c@{}}\scorecell{0.73366834}{29.20}{}\\[-2pt]{\fontsize{7.5}{9}\selectfont\color{black!65}(43.6)}\end{tabular} & \begin{tabular}[t]{@{}c@{}}\scorecell{0.73064516}{30.20}{}\\[-2pt]{\fontsize{7.5}{9}\selectfont\color{black!65}(110.9)}\end{tabular} & \begin{tabular}[t]{@{}c@{}}\scorecell{0.69372129}{30.20}{}\\[-2pt]{\fontsize{7.5}{9}\selectfont\color{black!65}(110.9)}\end{tabular} \\
Fast-dLLM & $8\times32$ & \emptycell & \begin{tabular}[t]{@{}c@{}}\scorecell{0.02469136}{0.80}{}\\[-2pt]{\fontsize{7.5}{9}\selectfont\color{black!65}(8.9)}\end{tabular} & \begin{tabular}[t]{@{}c@{}}\scorecell{0.30164534}{11.00}{}\\[-2pt]{\fontsize{7.5}{9}\selectfont\color{black!65}(24.0)}\end{tabular} & \begin{tabular}[t]{@{}c@{}}\scorecell{1.00000000}{\textbf{39.80}}{}\\[-2pt]{\fontsize{7.5}{9}\selectfont\color{black!65}(62.5)}\end{tabular} & \begin{tabular}[t]{@{}c@{}}\scorecell{0.97258065}{40.20}{}\\[-2pt]{\fontsize{7.5}{9}\selectfont\color{black!65}(71.5)}\end{tabular} & \begin{tabular}[t]{@{}c@{}}\scorecell{0.92343032}{40.20}{}\\[-2pt]{\fontsize{7.5}{9}\selectfont\color{black!65}(71.5)}\end{tabular} \\
Jazbec et al. (math-trained) & $8\times32$ & \emptycell & \emptycell & \emptycell & \begin{tabular}[t]{@{}c@{}}\scorecell{0.69514238}{27.67}{\textsuperscript{A}}\\[-2pt]{\fontsize{7.5}{9}\selectfont\color{black!65}(56.8)}\end{tabular} & \begin{tabular}[t]{@{}c@{}}\scorecell{0.92419355}{38.20}{\textsuperscript{A}}\\[-2pt]{\fontsize{7.5}{9}\selectfont\color{black!65}(116.1)}\end{tabular} & \begin{tabular}[t]{@{}c@{}}\scorecell{0.87748851}{38.20}{\textsuperscript{A}}\\[-2pt]{\fontsize{7.5}{9}\selectfont\color{black!65}(116.1)}\end{tabular} \\
Jazbec et al. (code-trained) & $8\times32$ & \emptycell & \emptycell & \begin{tabular}[t]{@{}c@{}}\scorecell{0.65265082}{23.80}{\textsuperscript{A}}\\[-2pt]{\fontsize{7.5}{9}\selectfont\color{black!65}(29.2)}\end{tabular} & \begin{tabular}[t]{@{}c@{}}\scorecell{0.78894472}{31.40}{\textsuperscript{A}}\\[-2pt]{\fontsize{7.5}{9}\selectfont\color{black!65}(55.2)}\end{tabular} & \begin{tabular}[t]{@{}c@{}}\scorecell{0.86129032}{35.60}{\textsuperscript{A}}\\[-2pt]{\fontsize{7.5}{9}\selectfont\color{black!65}(91.3)}\end{tabular} & \begin{tabular}[t]{@{}c@{}}\scorecell{0.81776417}{35.60}{\textsuperscript{A}}\\[-2pt]{\fontsize{7.5}{9}\selectfont\color{black!65}(91.3)}\end{tabular} \\
WINO & $1\times256$ & \emptycell & \emptycell & \emptycell & \begin{tabular}[t]{@{}c@{}}\scorecell{0.70351759}{28.00}{}\\[-2pt]{\fontsize{7.5}{9}\selectfont\color{black!65}(49.7)}\end{tabular} & \begin{tabular}[t]{@{}c@{}}\scorecell{0.67741935}{28.00}{}\\[-2pt]{\fontsize{7.5}{9}\selectfont\color{black!65}(49.7)}\end{tabular} & \begin{tabular}[t]{@{}c@{}}\scorecell{0.64318530}{28.00}{}\\[-2pt]{\fontsize{7.5}{9}\selectfont\color{black!65}(49.7)}\end{tabular} \\
WINO & $2\times128$ & \emptycell & \emptycell & \emptycell & \begin{tabular}[t]{@{}c@{}}\scorecell{0.79396985}{31.60}{}\\[-2pt]{\fontsize{7.5}{9}\selectfont\color{black!65}(52.3)}\end{tabular} & \begin{tabular}[t]{@{}c@{}}\scorecell{0.76451613}{31.60}{}\\[-2pt]{\fontsize{7.5}{9}\selectfont\color{black!65}(52.3)}\end{tabular} & \begin{tabular}[t]{@{}c@{}}\scorecell{0.72588055}{31.60}{}\\[-2pt]{\fontsize{7.5}{9}\selectfont\color{black!65}(52.3)}\end{tabular} \\
CORE & $1\times256$ & \begin{tabular}[t]{@{}c@{}}\scorecell{0.18110236}{4.60}{}\\[-2pt]{\fontsize{7.5}{9}\selectfont\color{black!65}(8.0)}\end{tabular} & \begin{tabular}[t]{@{}c@{}}\scorecell{0.27777778}{9.00}{}\\[-2pt]{\fontsize{7.5}{9}\selectfont\color{black!65}(16.0)}\end{tabular} & \begin{tabular}[t]{@{}c@{}}\scorecell{0.44424132}{16.20}{}\\[-2pt]{\fontsize{7.5}{9}\selectfont\color{black!65}(32.0)}\end{tabular} & \begin{tabular}[t]{@{}c@{}}\scorecell{0.61306533}{24.40}{}\\[-2pt]{\fontsize{7.5}{9}\selectfont\color{black!65}(64.0)}\end{tabular} & \begin{tabular}[t]{@{}c@{}}\scorecell{0.63870968}{26.40}{}\\[-2pt]{\fontsize{7.5}{9}\selectfont\color{black!65}(128.0)}\end{tabular} & \begin{tabular}[t]{@{}c@{}}\scorecell{0.63399694}{27.60}{}\\[-2pt]{\fontsize{7.5}{9}\selectfont\color{black!65}(256.0)}\end{tabular} \\
DAPD-Direct & $1\times256$ & \emptycell & \emptycell & \begin{tabular}[t]{@{}c@{}}\scorecell{0.81170018}{29.60}{}\\[-2pt]{\fontsize{7.5}{9}\selectfont\color{black!65}(23.3)}\end{tabular} & \begin{tabular}[t]{@{}c@{}}\scorecell{0.74371859}{29.60}{}\\[-2pt]{\fontsize{7.5}{9}\selectfont\color{black!65}(23.3)}\end{tabular} & \begin{tabular}[t]{@{}c@{}}\scorecell{0.71612903}{29.60}{}\\[-2pt]{\fontsize{7.5}{9}\selectfont\color{black!65}(23.3)}\end{tabular} & \begin{tabular}[t]{@{}c@{}}\scorecell{0.67993874}{29.60}{}\\[-2pt]{\fontsize{7.5}{9}\selectfont\color{black!65}(23.3)}\end{tabular} \\
DAPD-Direct & $4\times64$ & \emptycell & \emptycell & \emptycell & \begin{tabular}[t]{@{}c@{}}\scorecell{0.96482412}{38.40}{}\\[-2pt]{\fontsize{7.5}{9}\selectfont\color{black!65}(38.3)}\end{tabular} & \begin{tabular}[t]{@{}c@{}}\scorecell{0.92903226}{38.40}{}\\[-2pt]{\fontsize{7.5}{9}\selectfont\color{black!65}(38.3)}\end{tabular} & \begin{tabular}[t]{@{}c@{}}\scorecell{0.88208270}{38.40}{}\\[-2pt]{\fontsize{7.5}{9}\selectfont\color{black!65}(38.3)}\end{tabular} \\
DAPD-Staged & $1\times256$ & \emptycell & \emptycell & \emptycell & \begin{tabular}[t]{@{}c@{}}\scorecell{0.76884422}{30.60}{}\\[-2pt]{\fontsize{7.5}{9}\selectfont\color{black!65}(40.4)}\end{tabular} & \begin{tabular}[t]{@{}c@{}}\scorecell{0.74032258}{30.60}{}\\[-2pt]{\fontsize{7.5}{9}\selectfont\color{black!65}(40.4)}\end{tabular} & \begin{tabular}[t]{@{}c@{}}\scorecell{0.70290965}{30.60}{}\\[-2pt]{\fontsize{7.5}{9}\selectfont\color{black!65}(40.4)}\end{tabular} \\
DAPD-Staged & $4\times64$ & \emptycell & \emptycell & \emptycell & \emptycell & \begin{tabular}[t]{@{}c@{}}\scorecell{0.87096774}{36.00}{}\\[-2pt]{\fontsize{7.5}{9}\selectfont\color{black!65}(96.8)}\end{tabular} & \begin{tabular}[t]{@{}c@{}}\scorecell{0.82695253}{36.00}{}\\[-2pt]{\fontsize{7.5}{9}\selectfont\color{black!65}(96.8)}\end{tabular} \\
\midrule
\multicolumn{8}{l}{\textit{SAS: published evaluation protocol}}\\
SAS (3-shot) & $1\times256$ & \emptycell & \emptycell & \emptycell & \emptycell & \emptycell & \scorecell{0.00000000}{41.00}{\textsuperscript{S}} \\
SAS (3-shot) & $4\times64$ & \emptycell & \emptycell & \scorecell{0.00000000}{16.67}{\textsuperscript{S}} & \scorecell{0.00000000}{14.00}{\textsuperscript{S}} & \emptycell & \scorecell{0.00000000}{40.70}{\textsuperscript{S}} \\
SAS (3-shot) & $8\times32$ & \emptycell & \emptycell & \emptycell & \emptycell & \emptycell & \scorecell{0.00000000}{40.00}{\textsuperscript{S}} \\
\end{longtable}
\endgroup

Tables~\ref{tab:baseline-gsm8k}--\ref{tab:baseline-mbpp} use
32k-group ReMaRL checkpoints. The Blocks column reports the number of
blocks times the number of tokens per block.
Fixed-budget rows use exact NFE. Each adaptive cell reports the highest observed test score among configurations whose measured mean NFE does not exceed the column budget, with that mean shown below the score. When scores tie, we report the configuration with the lower mean NFE. The supplementary data retain every operating point.
UPO appears separately with 128-token answers and 32-token blocks.
SAS scores come from \citet{sas}, Tables~5, 7, and~9, with the reported
block sizes and prompting: 5-shot GSM8K and 3-shot MBPP; MATH500
prompting is unspecified.

The RCR~\citep{rcr} and CORE~\citep{core} comparisons use full-canvas
decoding. CORE's budget includes its context-verification calls;
the tables report the measured mean NFE. DAPD~\citep{dapd} uses
the released Direct and Staged configurations, each with one
256-token block or four 64-token blocks. 

\clearpage
\subsection{P2 confidence scoring}
\label{app:p2-scoring}

\begin{figure}[!ht]
 \centering
 \includegraphics[width=\linewidth]{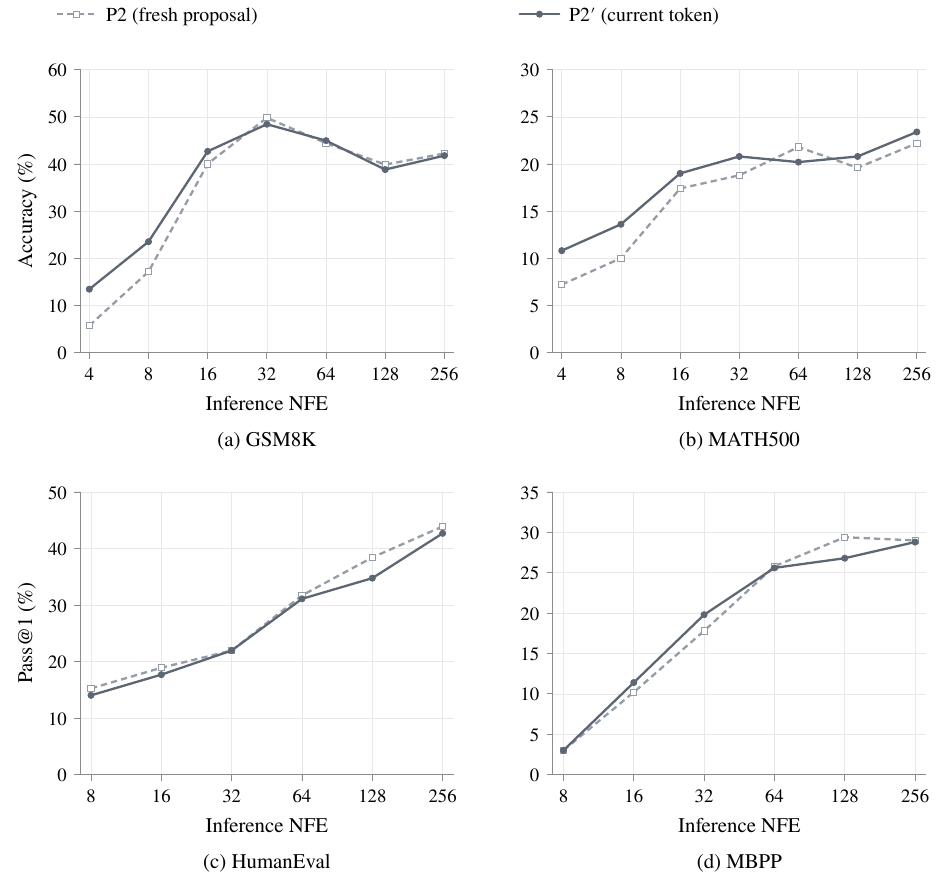}
 \caption{Original P2 and P2$'$ with greedy selection on the full test
 sets. P2 scores each fresh proposal; P2$'$ scores the current token
 at unmasked positions.}
 \label{fig:p2-scoring}
\end{figure}

Figure~\ref{fig:p2-scoring} compares P2's original self-planner~\citep{p2}
with greedy P2$'$ to isolate the effect of current-token scoring. Both use greedy token proposals
and greedy position selection, the same frozen LLaDA-8B-Instruct
checkpoint, 256-token answers, and our task-specific prompts and scorers.
The original self-planner uses equal weighting for masked and unmasked
positions and a linear visibility schedule.

Greedy P2$'$  improves accuracy over the original self-planner on both math
benchmarks at 4, 8, and 16 calls.
The differences are smaller at higher math budgets and on code,
where either scoring rule can give higher accuracy. We use P2$'$ because it scores each unmasked position using the token that selection would retain.

\clearpage
\section{Evaluation on Dream}
\label{app:dream}

We train separate math and code policies on frozen
Dream-v0-Instruct-7B~\citep{dream} at 8, 16, and 32 calls.
Each policy uses 32k accepted training groups, the same training
data and optimization recipe as the LLaDA policies, and 256-token
answers. We adapt the policy's hidden-state projection to Dream's
3,584-dimensional hidden states. Decoding uses Dream's tokenizer. For each answer token, we use
the logits and hidden state from the preceding sequence position.
Task prompts and answer scorers follow the same evaluation
pipeline as the LLaDA experiments.

All scores in this appendix are computed with the same frozen
Dream checkpoint and evaluation pipeline. We evaluate confidence decoding,
Fast-dLLM, EB-Sampler, KLASS, and both DAPD variants with full-answer
and 32-token blocks, and P2$'$ with full-answer blocks.
For adaptive methods, each table reports the highest observed test score
among evaluated settings within the column's mean-NFE budget,
following Appendix~\ref{app:full}. The Blocks column reports the number of blocks times the number of tokens per block.

\subsection{Math}
\begingroup
\definecolor{scorebar}{HTML}{D9ECF5}
\ifdefined\scorecellwidth\else\newlength{\scorecellwidth}\fi
\setlength{\scorecellwidth}{36pt}
\newcommand{\scorecell}[3]{%
 \makebox[\scorecellwidth][l]{%
  \rlap{\raisebox{-2.2pt}[0pt][0pt]{\color{scorebar}\rule{#1\scorecellwidth}{10pt}}}%
  \makebox[30pt][r]{#2}\makebox[6pt][l]{#3}}}
\newcommand{\emptycell}{\makebox[\scorecellwidth][c]{\textemdash}}

\fontsize{9.5}{10.5}\selectfont
\setlength{\LTcapwidth}{\textwidth}
\setlength{\LTpre}{3pt}
\setlength{\tabcolsep}{3.5pt}
\renewcommand{\arraystretch}{1.0}
\begin{longtable}{@{}>{\raggedright\arraybackslash}p{4.1cm} c *{6}{r}@{}}
\caption{Dream GSM8K accuracy (\%). Adaptive cells show the best observed test score within each mean-NFE budget, with actual mean NFE below. All scores are computed here. Bars scale from zero to the column maximum; bold marks the highest score. Dashes indicate an unmeasured budget or no eligible operating point.}\label{tab:dream-gsm8k}\\
\toprule
 & & \multicolumn{6}{c}{Mean-NFE budget}\\
Method & Blocks & \makebox[36pt][c]{8} & \makebox[36pt][c]{16} & \makebox[36pt][c]{32} & \makebox[36pt][c]{64} & \makebox[36pt][c]{128} & \makebox[36pt][c]{256} \\
\midrule
\endfirsthead
\multicolumn{8}{l}{\textit{Dream GSM8K comparison (continued)}}\\
\toprule
 & & \multicolumn{6}{c}{Mean-NFE budget}\\
Method & Blocks & \makebox[36pt][c]{8} & \makebox[36pt][c]{16} & \makebox[36pt][c]{32} & \makebox[36pt][c]{64} & \makebox[36pt][c]{128} & \makebox[36pt][c]{256} \\
\midrule
\endhead
\bottomrule
\endfoot
\multicolumn{8}{l}{\textit{Fixed-budget decoding}}\\
\textbf{ReMaRL (ours)} & $1\times256$ & \scorecell{1.00000000}{\textbf{47.86}}{} & \scorecell{1.00000000}{\textbf{58.15}}{} & \scorecell{1.00000000}{\textbf{65.48}}{} & \emptycell & \emptycell & \emptycell \\
P2$'$ (ours, $\tau=1$) & $1\times256$ & \scorecell{0.42397043}{20.29}{} & \scorecell{0.66232073}{38.51}{} & \scorecell{0.77460440}{50.72}{} & \scorecell{0.73172303}{57.42}{} & \scorecell{0.74859813}{60.73}{} & \scorecell{0.74256506}{60.58}{} \\
P2$'$ (ours, greedy) & $1\times256$ & \scorecell{0.04593453}{2.20}{} & \scorecell{0.16558018}{9.63}{} & \scorecell{0.41682748}{27.29}{} & \scorecell{0.49468599}{38.82}{} & \scorecell{0.48785047}{39.58}{} & \scorecell{0.43773234}{35.71}{} \\
Confidence & $1\times256$ & \scorecell{0.03326294}{1.59}{} & \scorecell{0.14732725}{8.57}{} & \scorecell{0.49324585}{32.30}{} & \scorecell{0.54782609}{42.99}{} & \scorecell{0.41401869}{33.59}{} & \scorecell{0.43866171}{35.78}{} \\
Confidence & $8\times32$ & \scorecell{0.00000000}{0.00}{} & \scorecell{0.01564537}{0.91}{} & \scorecell{0.20957159}{13.72}{} & \scorecell{0.70048309}{54.97}{} & \scorecell{0.93084112}{75.51}{} & \scorecell{1.00000000}{\textbf{81.58}}{} \\
\midrule
\multicolumn{8}{l}{\textit{Adaptive decoding}}\\
Fast-dLLM & $1\times256$ & \begin{tabular}[t]{@{}c@{}}\scorecell{0.06494192}{3.11}{}\\[-2pt]{\fontsize{7.5}{9}\selectfont\color{black!65}(6.9)}\end{tabular} & \begin{tabular}[t]{@{}c@{}}\scorecell{0.05345502}{3.11}{}\\[-2pt]{\fontsize{7.5}{9}\selectfont\color{black!65}(6.9)}\end{tabular} & \begin{tabular}[t]{@{}c@{}}\scorecell{0.42377460}{27.75}{}\\[-2pt]{\fontsize{7.5}{9}\selectfont\color{black!65}(19.0)}\end{tabular} & \begin{tabular}[t]{@{}c@{}}\scorecell{0.35362319}{27.75}{}\\[-2pt]{\fontsize{7.5}{9}\selectfont\color{black!65}(19.0)}\end{tabular} & \begin{tabular}[t]{@{}c@{}}\scorecell{0.44953271}{36.47}{}\\[-2pt]{\fontsize{7.5}{9}\selectfont\color{black!65}(118.0)}\end{tabular} & \begin{tabular}[t]{@{}c@{}}\scorecell{0.44702602}{36.47}{}\\[-2pt]{\fontsize{7.5}{9}\selectfont\color{black!65}(118.0)}\end{tabular} \\
Fast-dLLM & $8\times32$ & \begin{tabular}[t]{@{}c@{}}\scorecell{0.00000000}{0.00}{}\\[-2pt]{\fontsize{7.5}{9}\selectfont\color{black!65}(8.0)}\end{tabular} & \begin{tabular}[t]{@{}c@{}}\scorecell{0.00000000}{0.00}{}\\[-2pt]{\fontsize{7.5}{9}\selectfont\color{black!65}(8.0)}\end{tabular} & \begin{tabular}[t]{@{}c@{}}\scorecell{0.05094558}{3.34}{}\\[-2pt]{\fontsize{7.5}{9}\selectfont\color{black!65}(19.5)}\end{tabular} & \begin{tabular}[t]{@{}c@{}}\scorecell{1.00000000}{\textbf{78.47}}{}\\[-2pt]{\fontsize{7.5}{9}\selectfont\color{black!65}(62.8)}\end{tabular} & \begin{tabular}[t]{@{}c@{}}\scorecell{1.00000000}{\textbf{81.12}}{}\\[-2pt]{\fontsize{7.5}{9}\selectfont\color{black!65}(117.2)}\end{tabular} & \begin{tabular}[t]{@{}c@{}}\scorecell{0.99442379}{81.12}{}\\[-2pt]{\fontsize{7.5}{9}\selectfont\color{black!65}(117.2)}\end{tabular} \\
EB-Sampler & $1\times256$ & \begin{tabular}[t]{@{}c@{}}\scorecell{0.09662091}{4.62}{}\\[-2pt]{\fontsize{7.5}{9}\selectfont\color{black!65}(8.0)}\end{tabular} & \begin{tabular}[t]{@{}c@{}}\scorecell{0.29335072}{17.06}{}\\[-2pt]{\fontsize{7.5}{9}\selectfont\color{black!65}(13.1)}\end{tabular} & \begin{tabular}[t]{@{}c@{}}\scorecell{0.26051717}{17.06}{}\\[-2pt]{\fontsize{7.5}{9}\selectfont\color{black!65}(13.1)}\end{tabular} & \begin{tabular}[t]{@{}c@{}}\scorecell{0.49275362}{38.67}{}\\[-2pt]{\fontsize{7.5}{9}\selectfont\color{black!65}(32.8)}\end{tabular} & \begin{tabular}[t]{@{}c@{}}\scorecell{0.47663551}{38.67}{}\\[-2pt]{\fontsize{7.5}{9}\selectfont\color{black!65}(32.8)}\end{tabular} & \begin{tabular}[t]{@{}c@{}}\scorecell{0.47397770}{38.67}{}\\[-2pt]{\fontsize{7.5}{9}\selectfont\color{black!65}(32.8)}\end{tabular} \\
EB-Sampler & $8\times32$ & \emptycell & \begin{tabular}[t]{@{}c@{}}\scorecell{0.00000000}{0.00}{}\\[-2pt]{\fontsize{7.5}{9}\selectfont\color{black!65}(9.3)}\end{tabular} & \begin{tabular}[t]{@{}c@{}}\scorecell{0.00000000}{0.00}{}\\[-2pt]{\fontsize{7.5}{9}\selectfont\color{black!65}(9.3)}\end{tabular} & \begin{tabular}[t]{@{}c@{}}\scorecell{0.78743961}{61.79}{}\\[-2pt]{\fontsize{7.5}{9}\selectfont\color{black!65}(35.7)}\end{tabular} & \begin{tabular}[t]{@{}c@{}}\scorecell{0.99345794}{80.59}{}\\[-2pt]{\fontsize{7.5}{9}\selectfont\color{black!65}(72.2)}\end{tabular} & \begin{tabular}[t]{@{}c@{}}\scorecell{0.98977695}{80.74}{}\\[-2pt]{\fontsize{7.5}{9}\selectfont\color{black!65}(256.0)}\end{tabular} \\
KLASS & $1\times256$ & \emptycell & \emptycell & \emptycell & \emptycell & \begin{tabular}[t]{@{}c@{}}\scorecell{0.49345794}{40.03}{}\\[-2pt]{\fontsize{7.5}{9}\selectfont\color{black!65}(115.7)}\end{tabular} & \begin{tabular}[t]{@{}c@{}}\scorecell{0.49070632}{40.03}{}\\[-2pt]{\fontsize{7.5}{9}\selectfont\color{black!65}(115.7)}\end{tabular} \\
KLASS & $8\times32$ & \emptycell & \emptycell & \emptycell & \emptycell & \emptycell & \begin{tabular}[t]{@{}c@{}}\scorecell{0.99163569}{80.89}{}\\[-2pt]{\fontsize{7.5}{9}\selectfont\color{black!65}(256.0)}\end{tabular} \\
DAPD-Direct & $1\times256$ & \emptycell & \emptycell & \emptycell & \begin{tabular}[t]{@{}c@{}}\scorecell{0.65507246}{51.40}{}\\[-2pt]{\fontsize{7.5}{9}\selectfont\color{black!65}(48.0)}\end{tabular} & \begin{tabular}[t]{@{}c@{}}\scorecell{0.63364486}{51.40}{}\\[-2pt]{\fontsize{7.5}{9}\selectfont\color{black!65}(48.0)}\end{tabular} & \begin{tabular}[t]{@{}c@{}}\scorecell{0.63011152}{51.40}{}\\[-2pt]{\fontsize{7.5}{9}\selectfont\color{black!65}(48.0)}\end{tabular} \\
DAPD-Direct & $8\times32$ & \emptycell & \emptycell & \emptycell & \emptycell & \begin{tabular}[t]{@{}c@{}}\scorecell{0.97009346}{78.70}{}\\[-2pt]{\fontsize{7.5}{9}\selectfont\color{black!65}(93.5)}\end{tabular} & \begin{tabular}[t]{@{}c@{}}\scorecell{0.96468401}{78.70}{}\\[-2pt]{\fontsize{7.5}{9}\selectfont\color{black!65}(93.5)}\end{tabular} \\
DAPD-Staged & $1\times256$ & \emptycell & \emptycell & \emptycell & \begin{tabular}[t]{@{}c@{}}\scorecell{0.64444444}{50.57}{}\\[-2pt]{\fontsize{7.5}{9}\selectfont\color{black!65}(56.8)}\end{tabular} & \begin{tabular}[t]{@{}c@{}}\scorecell{0.62336449}{50.57}{}\\[-2pt]{\fontsize{7.5}{9}\selectfont\color{black!65}(56.8)}\end{tabular} & \begin{tabular}[t]{@{}c@{}}\scorecell{0.61988848}{50.57}{}\\[-2pt]{\fontsize{7.5}{9}\selectfont\color{black!65}(56.8)}\end{tabular} \\
DAPD-Staged & $8\times32$ & \emptycell & \emptycell & \emptycell & \emptycell & \emptycell & \begin{tabular}[t]{@{}c@{}}\scorecell{0.96282528}{78.54}{}\\[-2pt]{\fontsize{7.5}{9}\selectfont\color{black!65}(135.3)}\end{tabular} \\
\end{longtable}
\endgroup

\clearpage
\begingroup
\definecolor{scorebar}{HTML}{D9ECF5}
\ifdefined\scorecellwidth\else\newlength{\scorecellwidth}\fi
\setlength{\scorecellwidth}{36pt}
\newcommand{\scorecell}[3]{%
 \makebox[\scorecellwidth][l]{%
  \rlap{\raisebox{-2.2pt}[0pt][0pt]{\color{scorebar}\rule{#1\scorecellwidth}{10pt}}}%
  \makebox[30pt][r]{#2}\makebox[6pt][l]{#3}}}
\newcommand{\emptycell}{\makebox[\scorecellwidth][c]{\textemdash}}

\fontsize{9.5}{10.5}\selectfont
\setlength{\LTcapwidth}{\textwidth}
\setlength{\LTpre}{3pt}
\setlength{\tabcolsep}{3.5pt}
\renewcommand{\arraystretch}{1.0}
\begin{longtable}{@{}>{\raggedright\arraybackslash}p{4.1cm} c *{6}{r}@{}}
\caption{Dream MATH500 accuracy (\%). Adaptive cells show the best observed test score within each mean-NFE budget, with actual mean NFE below. All scores are computed here. Bars scale from zero to the column maximum; bold marks the highest score. Dashes indicate an unmeasured budget or no eligible operating point.}\label{tab:dream-math500}\\
\toprule
 & & \multicolumn{6}{c}{Mean-NFE budget}\\
Method & Blocks & \makebox[36pt][c]{8} & \makebox[36pt][c]{16} & \makebox[36pt][c]{32} & \makebox[36pt][c]{64} & \makebox[36pt][c]{128} & \makebox[36pt][c]{256} \\
\midrule
\endfirsthead
\multicolumn{8}{l}{\textit{Dream MATH500 comparison (continued)}}\\
\toprule
 & & \multicolumn{6}{c}{Mean-NFE budget}\\
Method & Blocks & \makebox[36pt][c]{8} & \makebox[36pt][c]{16} & \makebox[36pt][c]{32} & \makebox[36pt][c]{64} & \makebox[36pt][c]{128} & \makebox[36pt][c]{256} \\
\midrule
\endhead
\bottomrule
\endfoot
\multicolumn{8}{l}{\textit{Fixed-budget decoding}}\\
\textbf{ReMaRL (ours)} & $1\times256$ & \scorecell{1.00000000}{\textbf{18.60}}{} & \scorecell{1.00000000}{\textbf{24.33}}{} & \scorecell{1.00000000}{\textbf{27.40}}{} & \emptycell & \emptycell & \emptycell \\
P2$'$ (ours, $\tau=1$) & $1\times256$ & \scorecell{0.46953405}{8.73}{} & \scorecell{0.65753425}{16.00}{} & \scorecell{0.75425791}{20.67}{} & \scorecell{0.77551020}{22.80}{} & \scorecell{0.54308943}{22.27}{} & \scorecell{0.49388379}{21.53}{} \\
P2$'$ (ours, greedy) & $1\times256$ & \scorecell{0.03225806}{0.60}{} & \scorecell{0.15616438}{3.80}{} & \scorecell{0.46715328}{12.80}{} & \scorecell{0.67346939}{19.80}{} & \scorecell{0.46829268}{19.20}{} & \scorecell{0.49541284}{21.60}{} \\
Confidence & $1\times256$ & \scorecell{0.05376344}{1.00}{} & \scorecell{0.15616438}{3.80}{} & \scorecell{0.54744526}{15.00}{} & \scorecell{0.71428571}{21.00}{} & \scorecell{0.41951220}{17.20}{} & \scorecell{0.48623853}{21.20}{} \\
Confidence & $8\times32$ & \scorecell{0.00000000}{0.00}{} & \scorecell{0.04931507}{1.20}{} & \scorecell{0.33576642}{9.20}{} & \scorecell{0.91836735}{27.00}{} & \scorecell{0.94634146}{38.80}{} & \scorecell{1.00000000}{\textbf{43.60}}{} \\
\midrule
\multicolumn{8}{l}{\textit{Adaptive decoding}}\\
Fast-dLLM & $1\times256$ & \begin{tabular}[t]{@{}c@{}}\scorecell{0.12903226}{2.40}{}\\[-2pt]{\fontsize{7.5}{9}\selectfont\color{black!65}(7.9)}\end{tabular} & \begin{tabular}[t]{@{}c@{}}\scorecell{0.09863014}{2.40}{}\\[-2pt]{\fontsize{7.5}{9}\selectfont\color{black!65}(7.9)}\end{tabular} & \begin{tabular}[t]{@{}c@{}}\scorecell{0.68613139}{18.80}{}\\[-2pt]{\fontsize{7.5}{9}\selectfont\color{black!65}(20.6)}\end{tabular} & \begin{tabular}[t]{@{}c@{}}\scorecell{0.63945578}{18.80}{}\\[-2pt]{\fontsize{7.5}{9}\selectfont\color{black!65}(20.6)}\end{tabular} & \begin{tabular}[t]{@{}c@{}}\scorecell{0.46341463}{19.00}{}\\[-2pt]{\fontsize{7.5}{9}\selectfont\color{black!65}(64.4)}\end{tabular} & \begin{tabular}[t]{@{}c@{}}\scorecell{0.49082569}{21.40}{}\\[-2pt]{\fontsize{7.5}{9}\selectfont\color{black!65}(134.7)}\end{tabular} \\
Fast-dLLM & $8\times32$ & \begin{tabular}[t]{@{}c@{}}\scorecell{0.00000000}{0.00}{}\\[-2pt]{\fontsize{7.5}{9}\selectfont\color{black!65}(8.0)}\end{tabular} & \begin{tabular}[t]{@{}c@{}}\scorecell{0.00000000}{0.00}{}\\[-2pt]{\fontsize{7.5}{9}\selectfont\color{black!65}(8.0)}\end{tabular} & \begin{tabular}[t]{@{}c@{}}\scorecell{0.09489051}{2.60}{}\\[-2pt]{\fontsize{7.5}{9}\selectfont\color{black!65}(25.5)}\end{tabular} & \begin{tabular}[t]{@{}c@{}}\scorecell{0.74149660}{21.80}{}\\[-2pt]{\fontsize{7.5}{9}\selectfont\color{black!65}(45.1)}\end{tabular} & \begin{tabular}[t]{@{}c@{}}\scorecell{0.96585366}{39.60}{}\\[-2pt]{\fontsize{7.5}{9}\selectfont\color{black!65}(74.9)}\end{tabular} & \begin{tabular}[t]{@{}c@{}}\scorecell{0.98165138}{42.80}{}\\[-2pt]{\fontsize{7.5}{9}\selectfont\color{black!65}(256.0)}\end{tabular} \\
EB-Sampler & $1\times256$ & \emptycell & \begin{tabular}[t]{@{}c@{}}\scorecell{0.36986301}{9.00}{}\\[-2pt]{\fontsize{7.5}{9}\selectfont\color{black!65}(14.6)}\end{tabular} & \begin{tabular}[t]{@{}c@{}}\scorecell{0.32846715}{9.00}{}\\[-2pt]{\fontsize{7.5}{9}\selectfont\color{black!65}(14.6)}\end{tabular} & \begin{tabular}[t]{@{}c@{}}\scorecell{0.68707483}{20.20}{}\\[-2pt]{\fontsize{7.5}{9}\selectfont\color{black!65}(41.9)}\end{tabular} & \begin{tabular}[t]{@{}c@{}}\scorecell{0.49268293}{20.20}{}\\[-2pt]{\fontsize{7.5}{9}\selectfont\color{black!65}(41.9)}\end{tabular} & \begin{tabular}[t]{@{}c@{}}\scorecell{0.46788991}{20.40}{}\\[-2pt]{\fontsize{7.5}{9}\selectfont\color{black!65}(256.0)}\end{tabular} \\
EB-Sampler & $8\times32$ & \emptycell & \begin{tabular}[t]{@{}c@{}}\scorecell{0.00821918}{0.20}{}\\[-2pt]{\fontsize{7.5}{9}\selectfont\color{black!65}(11.4)}\end{tabular} & \begin{tabular}[t]{@{}c@{}}\scorecell{0.00729927}{0.20}{}\\[-2pt]{\fontsize{7.5}{9}\selectfont\color{black!65}(11.4)}\end{tabular} & \begin{tabular}[t]{@{}c@{}}\scorecell{0.99319728}{29.20}{}\\[-2pt]{\fontsize{7.5}{9}\selectfont\color{black!65}(41.3)}\end{tabular} & \begin{tabular}[t]{@{}c@{}}\scorecell{1.00000000}{\textbf{41.00}}{}\\[-2pt]{\fontsize{7.5}{9}\selectfont\color{black!65}(84.6)}\end{tabular} & \begin{tabular}[t]{@{}c@{}}\scorecell{1.00000000}{\textbf{43.60}}{}\\[-2pt]{\fontsize{7.5}{9}\selectfont\color{black!65}(164.0)}\end{tabular} \\
KLASS & $1\times256$ & \emptycell & \emptycell & \emptycell & \emptycell & \emptycell & \begin{tabular}[t]{@{}c@{}}\scorecell{0.49082569}{21.40}{}\\[-2pt]{\fontsize{7.5}{9}\selectfont\color{black!65}(133.8)}\end{tabular} \\
KLASS & $8\times32$ & \emptycell & \emptycell & \emptycell & \emptycell & \emptycell & \begin{tabular}[t]{@{}c@{}}\scorecell{0.98165138}{42.80}{}\\[-2pt]{\fontsize{7.5}{9}\selectfont\color{black!65}(256.0)}\end{tabular} \\
DAPD-Direct & $1\times256$ & \emptycell & \emptycell & \emptycell & \begin{tabular}[t]{@{}c@{}}\scorecell{1.00000000}{\textbf{29.40}}{}\\[-2pt]{\fontsize{7.5}{9}\selectfont\color{black!65}(53.7)}\end{tabular} & \begin{tabular}[t]{@{}c@{}}\scorecell{0.71707317}{29.40}{}\\[-2pt]{\fontsize{7.5}{9}\selectfont\color{black!65}(53.7)}\end{tabular} & \begin{tabular}[t]{@{}c@{}}\scorecell{0.67431193}{29.40}{}\\[-2pt]{\fontsize{7.5}{9}\selectfont\color{black!65}(53.7)}\end{tabular} \\
DAPD-Direct & $8\times32$ & \emptycell & \emptycell & \emptycell & \emptycell & \begin{tabular}[t]{@{}c@{}}\scorecell{0.98536585}{40.40}{}\\[-2pt]{\fontsize{7.5}{9}\selectfont\color{black!65}(108.3)}\end{tabular} & \begin{tabular}[t]{@{}c@{}}\scorecell{0.92660550}{40.40}{}\\[-2pt]{\fontsize{7.5}{9}\selectfont\color{black!65}(108.3)}\end{tabular} \\
DAPD-Staged & $1\times256$ & \emptycell & \emptycell & \emptycell & \begin{tabular}[t]{@{}c@{}}\scorecell{0.89115646}{26.20}{}\\[-2pt]{\fontsize{7.5}{9}\selectfont\color{black!65}(62.5)}\end{tabular} & \begin{tabular}[t]{@{}c@{}}\scorecell{0.63902439}{26.20}{}\\[-2pt]{\fontsize{7.5}{9}\selectfont\color{black!65}(62.5)}\end{tabular} & \begin{tabular}[t]{@{}c@{}}\scorecell{0.60091743}{26.20}{}\\[-2pt]{\fontsize{7.5}{9}\selectfont\color{black!65}(62.5)}\end{tabular} \\
DAPD-Staged & $8\times32$ & \emptycell & \emptycell & \emptycell & \emptycell & \emptycell & \begin{tabular}[t]{@{}c@{}}\scorecell{0.88532110}{38.60}{}\\[-2pt]{\fontsize{7.5}{9}\selectfont\color{black!65}(145.7)}\end{tabular} \\
\end{longtable}
\endgroup

\clearpage
\subsection{Code}
\begingroup
\definecolor{scorebar}{HTML}{D9ECF5}
\ifdefined\scorecellwidth\else\newlength{\scorecellwidth}\fi
\setlength{\scorecellwidth}{36pt}
\newcommand{\scorecell}[3]{%
 \makebox[\scorecellwidth][l]{%
  \rlap{\raisebox{-2.2pt}[0pt][0pt]{\color{scorebar}\rule{#1\scorecellwidth}{10pt}}}%
  \makebox[30pt][r]{#2}\makebox[6pt][l]{#3}}}
\newcommand{\emptycell}{\makebox[\scorecellwidth][c]{\textemdash}}

\fontsize{9.5}{10.5}\selectfont
\setlength{\LTcapwidth}{\textwidth}
\setlength{\LTpre}{3pt}
\setlength{\tabcolsep}{3.5pt}
\renewcommand{\arraystretch}{1.0}
\begin{longtable}{@{}>{\raggedright\arraybackslash}p{4.1cm} c *{6}{r}@{}}
\caption{Dream HumanEval pass@1 (\%). Adaptive cells show the best observed test score within each mean-NFE budget, with actual mean NFE below. All scores are computed here. Bars scale from zero to the column maximum; bold marks the highest score. Dashes indicate an unmeasured budget or no eligible operating point.}\label{tab:dream-humaneval}\\
\toprule
 & & \multicolumn{6}{c}{Mean-NFE budget}\\
Method & Blocks & \makebox[36pt][c]{8} & \makebox[36pt][c]{16} & \makebox[36pt][c]{32} & \makebox[36pt][c]{64} & \makebox[36pt][c]{128} & \makebox[36pt][c]{256} \\
\midrule
\endfirsthead
\multicolumn{8}{l}{\textit{Dream HumanEval comparison (continued)}}\\
\toprule
 & & \multicolumn{6}{c}{Mean-NFE budget}\\
Method & Blocks & \makebox[36pt][c]{8} & \makebox[36pt][c]{16} & \makebox[36pt][c]{32} & \makebox[36pt][c]{64} & \makebox[36pt][c]{128} & \makebox[36pt][c]{256} \\
\midrule
\endhead
\bottomrule
\endfoot
\multicolumn{8}{l}{\textit{Fixed-budget decoding}}\\
\textbf{ReMaRL (ours)} & $1\times256$ & \scorecell{1.00000000}{\textbf{24.39}}{} & \scorecell{1.00000000}{\textbf{38.01}}{} & \scorecell{1.00000000}{\textbf{46.95}}{} & \emptycell & \emptycell & \emptycell \\
P2$'$ (ours, $\tau=1$) & $1\times256$ & \scorecell{0.41666667}{10.16}{} & \scorecell{0.50802139}{19.31}{} & \scorecell{0.55844156}{26.22}{} & \scorecell{0.55925926}{30.69}{} & \scorecell{0.51485149}{31.71}{} & \scorecell{0.54575163}{33.94}{} \\
P2$'$ (ours, greedy) & $1\times256$ & \scorecell{0.30000000}{7.32}{} & \scorecell{0.36898396}{14.02}{} & \scorecell{0.40259740}{18.90}{} & \scorecell{0.51111111}{28.05}{} & \scorecell{0.47524752}{29.27}{} & \scorecell{0.53921569}{33.54}{} \\
Confidence & $1\times256$ & \scorecell{0.37500000}{9.15}{} & \scorecell{0.30481283}{11.59}{} & \scorecell{0.45454545}{21.34}{} & \scorecell{0.52222222}{28.66}{} & \scorecell{0.46534653}{28.66}{} & \scorecell{0.42156863}{26.22}{} \\
Confidence & $8\times32$ & \scorecell{0.02500000}{0.61}{} & \scorecell{0.04812834}{1.83}{} & \scorecell{0.14285714}{6.71}{} & \scorecell{0.46666667}{25.61}{} & \scorecell{0.79207921}{48.78}{} & \scorecell{0.99019608}{61.59}{} \\
\midrule
\multicolumn{8}{l}{\textit{Adaptive decoding}}\\
Fast-dLLM & $1\times256$ & \begin{tabular}[t]{@{}c@{}}\scorecell{0.35000000}{8.54}{}\\[-2pt]{\fontsize{7.5}{9}\selectfont\color{black!65}(7.3)}\end{tabular} & \begin{tabular}[t]{@{}c@{}}\scorecell{0.35294118}{13.41}{}\\[-2pt]{\fontsize{7.5}{9}\selectfont\color{black!65}(9.2)}\end{tabular} & \begin{tabular}[t]{@{}c@{}}\scorecell{0.28571429}{13.41}{}\\[-2pt]{\fontsize{7.5}{9}\selectfont\color{black!65}(9.2)}\end{tabular} & \begin{tabular}[t]{@{}c@{}}\scorecell{0.47777778}{26.22}{}\\[-2pt]{\fontsize{7.5}{9}\selectfont\color{black!65}(41.3)}\end{tabular} & \begin{tabular}[t]{@{}c@{}}\scorecell{0.43564356}{26.83}{}\\[-2pt]{\fontsize{7.5}{9}\selectfont\color{black!65}(107.9)}\end{tabular} & \begin{tabular}[t]{@{}c@{}}\scorecell{0.44117647}{27.44}{}\\[-2pt]{\fontsize{7.5}{9}\selectfont\color{black!65}(256.0)}\end{tabular} \\
Fast-dLLM & $8\times32$ & \begin{tabular}[t]{@{}c@{}}\scorecell{0.02500000}{0.61}{}\\[-2pt]{\fontsize{7.5}{9}\selectfont\color{black!65}(8.0)}\end{tabular} & \begin{tabular}[t]{@{}c@{}}\scorecell{0.01604278}{0.61}{}\\[-2pt]{\fontsize{7.5}{9}\selectfont\color{black!65}(8.0)}\end{tabular} & \begin{tabular}[t]{@{}c@{}}\scorecell{0.45454545}{21.34}{}\\[-2pt]{\fontsize{7.5}{9}\selectfont\color{black!65}(29.1)}\end{tabular} & \begin{tabular}[t]{@{}c@{}}\scorecell{0.94444444}{51.83}{}\\[-2pt]{\fontsize{7.5}{9}\selectfont\color{black!65}(54.5)}\end{tabular} & \begin{tabular}[t]{@{}c@{}}\scorecell{1.00000000}{\textbf{61.59}}{}\\[-2pt]{\fontsize{7.5}{9}\selectfont\color{black!65}(104.4)}\end{tabular} & \begin{tabular}[t]{@{}c@{}}\scorecell{0.99019608}{61.59}{}\\[-2pt]{\fontsize{7.5}{9}\selectfont\color{black!65}(104.4)}\end{tabular} \\
EB-Sampler & $1\times256$ & \begin{tabular}[t]{@{}c@{}}\scorecell{0.25000000}{6.10}{}\\[-2pt]{\fontsize{7.5}{9}\selectfont\color{black!65}(7.0)}\end{tabular} & \begin{tabular}[t]{@{}c@{}}\scorecell{0.16042781}{6.10}{}\\[-2pt]{\fontsize{7.5}{9}\selectfont\color{black!65}(7.0)}\end{tabular} & \begin{tabular}[t]{@{}c@{}}\scorecell{0.50649351}{23.78}{}\\[-2pt]{\fontsize{7.5}{9}\selectfont\color{black!65}(28.0)}\end{tabular} & \begin{tabular}[t]{@{}c@{}}\scorecell{0.43333333}{23.78}{}\\[-2pt]{\fontsize{7.5}{9}\selectfont\color{black!65}(28.0)}\end{tabular} & \begin{tabular}[t]{@{}c@{}}\scorecell{0.47524752}{29.27}{}\\[-2pt]{\fontsize{7.5}{9}\selectfont\color{black!65}(118.9)}\end{tabular} & \begin{tabular}[t]{@{}c@{}}\scorecell{0.47058824}{29.27}{}\\[-2pt]{\fontsize{7.5}{9}\selectfont\color{black!65}(118.9)}\end{tabular} \\
EB-Sampler & $8\times32$ & \emptycell & \begin{tabular}[t]{@{}c@{}}\scorecell{0.09625668}{3.66}{}\\[-2pt]{\fontsize{7.5}{9}\selectfont\color{black!65}(15.5)}\end{tabular} & \begin{tabular}[t]{@{}c@{}}\scorecell{0.59740260}{28.05}{}\\[-2pt]{\fontsize{7.5}{9}\selectfont\color{black!65}(27.4)}\end{tabular} & \begin{tabular}[t]{@{}c@{}}\scorecell{1.00000000}{\textbf{54.88}}{}\\[-2pt]{\fontsize{7.5}{9}\selectfont\color{black!65}(57.8)}\end{tabular} & \begin{tabular}[t]{@{}c@{}}\scorecell{0.98019802}{60.37}{}\\[-2pt]{\fontsize{7.5}{9}\selectfont\color{black!65}(108.0)}\end{tabular} & \begin{tabular}[t]{@{}c@{}}\scorecell{1.00000000}{\textbf{62.20}}{}\\[-2pt]{\fontsize{7.5}{9}\selectfont\color{black!65}(256.0)}\end{tabular} \\
KLASS & $1\times256$ & \emptycell & \emptycell & \emptycell & \begin{tabular}[t]{@{}c@{}}\scorecell{0.40000000}{21.95}{}\\[-2pt]{\fontsize{7.5}{9}\selectfont\color{black!65}(52.4)}\end{tabular} & \begin{tabular}[t]{@{}c@{}}\scorecell{0.46534653}{28.66}{}\\[-2pt]{\fontsize{7.5}{9}\selectfont\color{black!65}(92.2)}\end{tabular} & \begin{tabular}[t]{@{}c@{}}\scorecell{0.46078431}{28.66}{}\\[-2pt]{\fontsize{7.5}{9}\selectfont\color{black!65}(92.2)}\end{tabular} \\
KLASS & $8\times32$ & \emptycell & \emptycell & \emptycell & \emptycell & \begin{tabular}[t]{@{}c@{}}\scorecell{0.95049505}{58.54}{}\\[-2pt]{\fontsize{7.5}{9}\selectfont\color{black!65}(99.5)}\end{tabular} & \begin{tabular}[t]{@{}c@{}}\scorecell{0.99019608}{61.59}{}\\[-2pt]{\fontsize{7.5}{9}\selectfont\color{black!65}(256.0)}\end{tabular} \\
DAPD-Direct & $1\times256$ & \emptycell & \emptycell & \emptycell & \emptycell & \emptycell & \begin{tabular}[t]{@{}c@{}}\scorecell{0.41176471}{25.61}{}\\[-2pt]{\fontsize{7.5}{9}\selectfont\color{black!65}(132.9)}\end{tabular} \\
DAPD-Direct & $8\times32$ & \emptycell & \emptycell & \emptycell & \emptycell & \begin{tabular}[t]{@{}c@{}}\scorecell{0.93069307}{57.32}{}\\[-2pt]{\fontsize{7.5}{9}\selectfont\color{black!65}(73.5)}\end{tabular} & \begin{tabular}[t]{@{}c@{}}\scorecell{0.92156863}{57.32}{}\\[-2pt]{\fontsize{7.5}{9}\selectfont\color{black!65}(73.5)}\end{tabular} \\
DAPD-Staged & $1\times256$ & \emptycell & \emptycell & \emptycell & \emptycell & \emptycell & \begin{tabular}[t]{@{}c@{}}\scorecell{0.44117647}{27.44}{}\\[-2pt]{\fontsize{7.5}{9}\selectfont\color{black!65}(154.8)}\end{tabular} \\
DAPD-Staged & $8\times32$ & \emptycell & \emptycell & \emptycell & \emptycell & \begin{tabular}[t]{@{}c@{}}\scorecell{0.89108911}{54.88}{}\\[-2pt]{\fontsize{7.5}{9}\selectfont\color{black!65}(127.4)}\end{tabular} & \begin{tabular}[t]{@{}c@{}}\scorecell{0.88235294}{54.88}{}\\[-2pt]{\fontsize{7.5}{9}\selectfont\color{black!65}(127.4)}\end{tabular} \\
\end{longtable}
\endgroup

\clearpage
\begingroup
\definecolor{scorebar}{HTML}{D9ECF5}
\ifdefined\scorecellwidth\else\newlength{\scorecellwidth}\fi
\setlength{\scorecellwidth}{36pt}
\newcommand{\scorecell}[3]{%
 \makebox[\scorecellwidth][l]{%
  \rlap{\raisebox{-2.2pt}[0pt][0pt]{\color{scorebar}\rule{#1\scorecellwidth}{10pt}}}%
  \makebox[30pt][r]{#2}\makebox[6pt][l]{#3}}}
\newcommand{\emptycell}{\makebox[\scorecellwidth][c]{\textemdash}}

\fontsize{9.5}{10.5}\selectfont
\setlength{\LTcapwidth}{\textwidth}
\setlength{\LTpre}{3pt}
\setlength{\tabcolsep}{3.5pt}
\renewcommand{\arraystretch}{1.0}
\begin{longtable}{@{}>{\raggedright\arraybackslash}p{4.1cm} c *{6}{r}@{}}
\caption{Dream MBPP pass@1 (\%). Adaptive cells show the best observed test score within each mean-NFE budget, with actual mean NFE below. All scores are computed here. Bars scale from zero to the column maximum; bold marks the highest score. Dashes indicate an unmeasured budget or no eligible operating point.}\label{tab:dream-mbpp}\\
\toprule
 & & \multicolumn{6}{c}{Mean-NFE budget}\\
Method & Blocks & \makebox[36pt][c]{8} & \makebox[36pt][c]{16} & \makebox[36pt][c]{32} & \makebox[36pt][c]{64} & \makebox[36pt][c]{128} & \makebox[36pt][c]{256} \\
\midrule
\endfirsthead
\multicolumn{8}{l}{\textit{Dream MBPP comparison (continued)}}\\
\toprule
 & & \multicolumn{6}{c}{Mean-NFE budget}\\
Method & Blocks & \makebox[36pt][c]{8} & \makebox[36pt][c]{16} & \makebox[36pt][c]{32} & \makebox[36pt][c]{64} & \makebox[36pt][c]{128} & \makebox[36pt][c]{256} \\
\midrule
\endhead
\bottomrule
\endfoot
\multicolumn{8}{l}{\textit{Fixed-budget decoding}}\\
\textbf{ReMaRL (ours)} & $1\times256$ & \scorecell{1.00000000}{\textbf{29.27}}{} & \scorecell{1.00000000}{\textbf{40.13}}{} & \scorecell{0.98232323}{51.87}{} & \emptycell & \emptycell & \emptycell \\
P2$'$ (ours, $\tau=1$) & $1\times256$ & \scorecell{0.80637813}{23.60}{} & \scorecell{0.84883721}{34.07}{} & \scorecell{0.79292929}{41.87}{} & \scorecell{0.82613909}{45.93}{} & \scorecell{0.84223013}{47.33}{} & \scorecell{0.88137604}{49.53}{} \\
P2$'$ (ours, greedy) & $1\times256$ & \scorecell{0.02050114}{0.60}{} & \scorecell{0.22923588}{9.20}{} & \scorecell{0.44318182}{23.40}{} & \scorecell{0.71582734}{39.80}{} & \scorecell{0.80782918}{45.40}{} & \scorecell{0.81138790}{45.60}{} \\
Confidence & $1\times256$ & \scorecell{0.04783599}{1.40}{} & \scorecell{0.16445183}{6.60}{} & \scorecell{0.43181818}{22.80}{} & \scorecell{0.79856115}{44.40}{} & \scorecell{0.81494662}{45.80}{} & \scorecell{0.84697509}{47.60}{} \\
Confidence & $8\times32$ & \scorecell{0.02050114}{0.60}{} & \scorecell{0.09966777}{4.00}{} & \scorecell{0.31818182}{16.80}{} & \scorecell{0.56115108}{31.20}{} & \scorecell{0.79359431}{44.60}{} & \scorecell{0.98932384}{55.60}{} \\
\midrule
\multicolumn{8}{l}{\textit{Adaptive decoding}}\\
Fast-dLLM & $1\times256$ & \begin{tabular}[t]{@{}c@{}}\scorecell{0.62186788}{18.20}{}\\[-2pt]{\fontsize{7.5}{9}\selectfont\color{black!65}(6.8)}\end{tabular} & \begin{tabular}[t]{@{}c@{}}\scorecell{0.45348837}{18.20}{}\\[-2pt]{\fontsize{7.5}{9}\selectfont\color{black!65}(6.8)}\end{tabular} & \begin{tabular}[t]{@{}c@{}}\scorecell{0.85606061}{45.20}{}\\[-2pt]{\fontsize{7.5}{9}\selectfont\color{black!65}(28.0)}\end{tabular} & \begin{tabular}[t]{@{}c@{}}\scorecell{0.87769784}{48.80}{}\\[-2pt]{\fontsize{7.5}{9}\selectfont\color{black!65}(57.6)}\end{tabular} & \begin{tabular}[t]{@{}c@{}}\scorecell{0.86832740}{48.80}{}\\[-2pt]{\fontsize{7.5}{9}\selectfont\color{black!65}(57.6)}\end{tabular} & \begin{tabular}[t]{@{}c@{}}\scorecell{0.86832740}{48.80}{}\\[-2pt]{\fontsize{7.5}{9}\selectfont\color{black!65}(57.6)}\end{tabular} \\
Fast-dLLM & $8\times32$ & \begin{tabular}[t]{@{}c@{}}\scorecell{0.02733485}{0.80}{}\\[-2pt]{\fontsize{7.5}{9}\selectfont\color{black!65}(8.0)}\end{tabular} & \begin{tabular}[t]{@{}c@{}}\scorecell{0.49833887}{20.00}{}\\[-2pt]{\fontsize{7.5}{9}\selectfont\color{black!65}(15.3)}\end{tabular} & \begin{tabular}[t]{@{}c@{}}\scorecell{0.95075758}{50.20}{}\\[-2pt]{\fontsize{7.5}{9}\selectfont\color{black!65}(22.5)}\end{tabular} & \begin{tabular}[t]{@{}c@{}}\scorecell{1.00000000}{\textbf{55.60}}{}\\[-2pt]{\fontsize{7.5}{9}\selectfont\color{black!65}(47.7)}\end{tabular} & \begin{tabular}[t]{@{}c@{}}\scorecell{0.98932384}{55.60}{}\\[-2pt]{\fontsize{7.5}{9}\selectfont\color{black!65}(47.7)}\end{tabular} & \begin{tabular}[t]{@{}c@{}}\scorecell{0.98932384}{55.60}{}\\[-2pt]{\fontsize{7.5}{9}\selectfont\color{black!65}(47.7)}\end{tabular} \\
EB-Sampler & $1\times256$ & \begin{tabular}[t]{@{}c@{}}\scorecell{0.12984055}{3.80}{}\\[-2pt]{\fontsize{7.5}{9}\selectfont\color{black!65}(4.8)}\end{tabular} & \begin{tabular}[t]{@{}c@{}}\scorecell{0.93687708}{37.60}{}\\[-2pt]{\fontsize{7.5}{9}\selectfont\color{black!65}(15.2)}\end{tabular} & \begin{tabular}[t]{@{}c@{}}\scorecell{0.71212121}{37.60}{}\\[-2pt]{\fontsize{7.5}{9}\selectfont\color{black!65}(15.2)}\end{tabular} & \begin{tabular}[t]{@{}c@{}}\scorecell{0.86690647}{48.20}{}\\[-2pt]{\fontsize{7.5}{9}\selectfont\color{black!65}(63.8)}\end{tabular} & \begin{tabular}[t]{@{}c@{}}\scorecell{0.85765125}{48.20}{}\\[-2pt]{\fontsize{7.5}{9}\selectfont\color{black!65}(63.8)}\end{tabular} & \begin{tabular}[t]{@{}c@{}}\scorecell{0.86476868}{48.60}{}\\[-2pt]{\fontsize{7.5}{9}\selectfont\color{black!65}(256.0)}\end{tabular} \\
EB-Sampler & $8\times32$ & \emptycell & \begin{tabular}[t]{@{}c@{}}\scorecell{0.22923588}{9.20}{}\\[-2pt]{\fontsize{7.5}{9}\selectfont\color{black!65}(12.7)}\end{tabular} & \begin{tabular}[t]{@{}c@{}}\scorecell{1.00000000}{\textbf{52.80}}{}\\[-2pt]{\fontsize{7.5}{9}\selectfont\color{black!65}(28.1)}\end{tabular} & \begin{tabular}[t]{@{}c@{}}\scorecell{0.99640288}{55.40}{}\\[-2pt]{\fontsize{7.5}{9}\selectfont\color{black!65}(51.0)}\end{tabular} & \begin{tabular}[t]{@{}c@{}}\scorecell{0.98576512}{55.40}{}\\[-2pt]{\fontsize{7.5}{9}\selectfont\color{black!65}(51.0)}\end{tabular} & \begin{tabular}[t]{@{}c@{}}\scorecell{0.98576512}{55.40}{}\\[-2pt]{\fontsize{7.5}{9}\selectfont\color{black!65}(51.0)}\end{tabular} \\
KLASS & $1\times256$ & \emptycell & \emptycell & \emptycell & \begin{tabular}[t]{@{}c@{}}\scorecell{0.84172662}{46.80}{}\\[-2pt]{\fontsize{7.5}{9}\selectfont\color{black!65}(53.7)}\end{tabular} & \begin{tabular}[t]{@{}c@{}}\scorecell{0.83274021}{46.80}{}\\[-2pt]{\fontsize{7.5}{9}\selectfont\color{black!65}(53.7)}\end{tabular} & \begin{tabular}[t]{@{}c@{}}\scorecell{0.86120996}{48.40}{}\\[-2pt]{\fontsize{7.5}{9}\selectfont\color{black!65}(256.0)}\end{tabular} \\
KLASS & $8\times32$ & \emptycell & \emptycell & \emptycell & \begin{tabular}[t]{@{}c@{}}\scorecell{0.89568345}{49.80}{}\\[-2pt]{\fontsize{7.5}{9}\selectfont\color{black!65}(54.5)}\end{tabular} & \begin{tabular}[t]{@{}c@{}}\scorecell{0.88612100}{49.80}{}\\[-2pt]{\fontsize{7.5}{9}\selectfont\color{black!65}(54.5)}\end{tabular} & \begin{tabular}[t]{@{}c@{}}\scorecell{0.98932384}{55.60}{}\\[-2pt]{\fontsize{7.5}{9}\selectfont\color{black!65}(256.0)}\end{tabular} \\
DAPD-Direct & $1\times256$ & \emptycell & \emptycell & \begin{tabular}[t]{@{}c@{}}\scorecell{0.97727273}{51.60}{}\\[-2pt]{\fontsize{7.5}{9}\selectfont\color{black!65}(26.7)}\end{tabular} & \begin{tabular}[t]{@{}c@{}}\scorecell{0.92805755}{51.60}{}\\[-2pt]{\fontsize{7.5}{9}\selectfont\color{black!65}(26.7)}\end{tabular} & \begin{tabular}[t]{@{}c@{}}\scorecell{0.91814947}{51.60}{}\\[-2pt]{\fontsize{7.5}{9}\selectfont\color{black!65}(26.7)}\end{tabular} & \begin{tabular}[t]{@{}c@{}}\scorecell{0.91814947}{51.60}{}\\[-2pt]{\fontsize{7.5}{9}\selectfont\color{black!65}(26.7)}\end{tabular} \\
DAPD-Direct & $8\times32$ & \emptycell & \emptycell & \emptycell & \begin{tabular}[t]{@{}c@{}}\scorecell{0.99640288}{55.40}{}\\[-2pt]{\fontsize{7.5}{9}\selectfont\color{black!65}(34.1)}\end{tabular} & \begin{tabular}[t]{@{}c@{}}\scorecell{0.98576512}{55.40}{}\\[-2pt]{\fontsize{7.5}{9}\selectfont\color{black!65}(34.1)}\end{tabular} & \begin{tabular}[t]{@{}c@{}}\scorecell{0.98576512}{55.40}{}\\[-2pt]{\fontsize{7.5}{9}\selectfont\color{black!65}(34.1)}\end{tabular} \\
DAPD-Staged & $1\times256$ & \emptycell & \emptycell & \emptycell & \begin{tabular}[t]{@{}c@{}}\scorecell{0.92086331}{51.20}{}\\[-2pt]{\fontsize{7.5}{9}\selectfont\color{black!65}(47.3)}\end{tabular} & \begin{tabular}[t]{@{}c@{}}\scorecell{0.91103203}{51.20}{}\\[-2pt]{\fontsize{7.5}{9}\selectfont\color{black!65}(47.3)}\end{tabular} & \begin{tabular}[t]{@{}c@{}}\scorecell{0.91103203}{51.20}{}\\[-2pt]{\fontsize{7.5}{9}\selectfont\color{black!65}(47.3)}\end{tabular} \\
DAPD-Staged & $8\times32$ & \emptycell & \emptycell & \emptycell & \emptycell & \begin{tabular}[t]{@{}c@{}}\scorecell{1.00000000}{\textbf{56.20}}{}\\[-2pt]{\fontsize{7.5}{9}\selectfont\color{black!65}(119.4)}\end{tabular} & \begin{tabular}[t]{@{}c@{}}\scorecell{1.00000000}{\textbf{56.20}}{}\\[-2pt]{\fontsize{7.5}{9}\selectfont\color{black!65}(119.4)}\end{tabular} \\
\end{longtable}
\endgroup

\clearpage

\section{Generation trajectory analysis}
\label{app:trajectories}

We analyze generation trajectories on all 1,319 GSM8K test
questions using the NFE32 ReMaRL policy trained for 32k
groups. We compare it with P2$'$ at $\tau\in\{0.25,1\}$
and confidence-based decoding. All methods use the same
frozen LLaDA model, prompts, 256 answer positions, and
greedy token proposals. We record every intermediate
answer, using three inference seeds for stochastic methods
and one run for deterministic confidence decoding.

\begin{figure}[H]
  \centering
  \includegraphics[width=\linewidth]
    {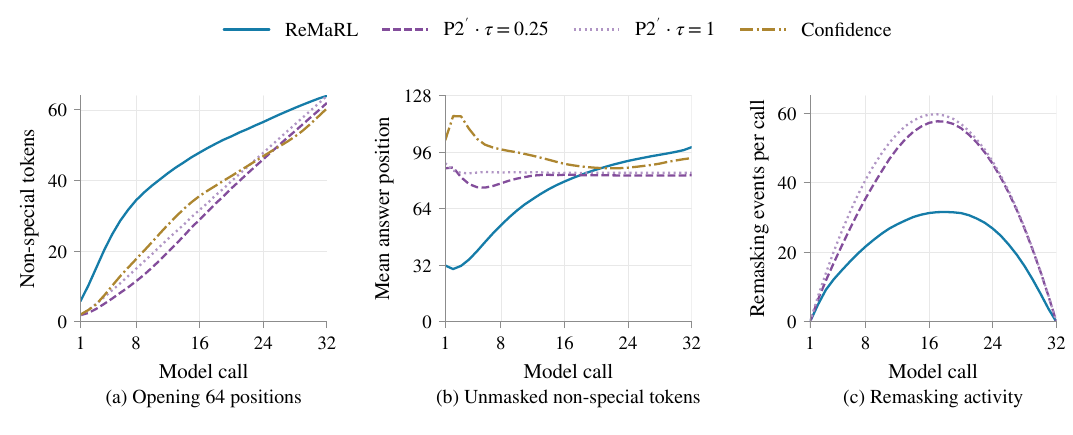}
  \caption{Generation trajectories on GSM8K at NFE32.
  (a) Visible non-special tokens in the first 64 positions;
  (b) mean position of all currently unmasked tokens,
  excluding special tokens;
  (c) remasking events per call. Curves show averages
  over questions and inference seeds.}
  \label{fig:trajectory-analysis}
\end{figure}

\paragraph{The opening takes shape earlier.}
ReMaRL prioritizes the beginning of the answer
(Figure~\ref{fig:trajectory-analysis}a). After eight calls, ReMaRL has roughly two to three times
as many visible non-special tokens in the first 64 positions
as the other methods.

We also measure the mean position of all currently unmasked
tokens
after excluding special tokens.
For ReMaRL, this mean rises during generation, while both P2$'$ curves remain roughly flat
(Figure~\ref{fig:trajectory-analysis}b). Together, these
patterns support an autoregressive-like ordering preference
that emerges without an explicit left-to-right constraint.

\paragraph{Fewer remasking events.}
ReMaRL consistently remasks less often than P2$'$ at
$\tau=0.25$ and $\tau=1$ throughout generation
(Figure~\ref{fig:trajectory-analysis}c).

A token replacement occurs when a remasked position
receives a different token at its next reveal.
ReMaRL makes fewer replacements than P2$'$ at $\tau=1$
and more than calibrated P2$'$
(Table~\ref{tab:trajectory-remasking}).
One possible explanation is that selection at $\tau=1$
is too random, causing unnecessary replacements.
Lowering the temperature reduces this randomness but may
also suppress useful replacements. ReMaRL may learn to
recover these useful changes while avoiding repeated
remasking that only restores the same token.

\begin{table}[H]
  \centering
  \small
  \caption{Mean remasking events and token replacements
  per answer, rounded to whole counts. The last column
  gives the percentage of remasking events that lead
  to a token replacement.}
  \label{tab:trajectory-remasking}
  \begin{tabular}{@{}lrrr@{}}
    \toprule
    Method & Remasking events & Replacements & Replaced (\%) \\
    \midrule
    ReMaRL & 673 & 33 & 4.9 \\
    P2$'$, $\tau=0.25$ & 1,158 & 27 & 2.3 \\
    P2$'$, $\tau=1$ & 1,227 & 84 & 6.8 \\
    \bottomrule
  \end{tabular}
\end{table}

\clearpage
\section{Code-data overlap}
\label{app:data}

We check that each KodCode training problem has executable tests that its reference solution passes, and we screen the problems for overlap with the evaluation benchmarks.
In the audited KodCode-Light-RL-10K snapshot, task
\texttt{Filter\_25989\_I} is a high-confidence near-duplicate of
HumanEval/117.\footnote{Dataset revision
\href{https://huggingface.co/datasets/KodCode/KodCode-Light-RL-10K/tree/dcf78a8bbba9a613b596ce993c4921a38687dfcc}{dcf78a8b}.}
Both specify \texttt{select\_words(s, n)}, select words by consonant
count, preserve word order, and share the example
\texttt{select\_words('Mary had a little lamb', 4) = ['little']}.
Their implementations differ. We exclude the training problem and
leave HumanEval unchanged.

We conservatively quarantined 156 candidate training rows. This set
includes uncertain matches and common elementary exercises, so it
should not be read as 156 confirmed benchmark leaks. The original
KodCode analysis~\citep{kodcode} reports 94 potentially contaminated questions in its larger corpus and describes the exclusions used for its own evaluation.
The residual overlap in Light-RL-10K supports screening each training
subset against the benchmarks used for evaluation.

We also inspected released training code for methods using this
dataset. The loader released by \citet{apple} formats all input rows without a
benchmark-overlap filter.\footnote{\href{https://github.com/apple-aiml-research/ml-rl-dllm/blob/1309c06470deeea529645efcb79bd0744a16c445/data/data_utils.py\#L134-L167}{Jazbec et al. data loader, revision 1309c064}.}
Its local-data fallback permits preprocessing outside the released
loader, so the exact historical training rows remain unverified.
GDPO~\citep{gdpo} retains rows with at least three extracted
assertions; the near-duplicate above has six and passes that
filter.\footnote{\href{https://github.com/KevinRojas1499/GDPO/blob/1918683043ce92b8ec628fceb39b7f29a614a35c/gdpo/data_utils.py\#L174-L197}{GDPO data loader, revision 19186830}.}
We found no benchmark-overlap removal in that released path.

MAPG and SPG report KodCode training, but the inspected releases
provide no coding-data loader.\footnote{
\href{https://github.com/Haran71/mask-aware-policy-gradients/blob/5a10e472c861cec863a702538904845e650ea0c3/map/train.py}{MAPG revision 5a10e472};
\href{https://github.com/facebookresearch/SPG/blob/bc8f521c9f16ce8c69f397d7ca7f5ffc57f9f248/spg/diffu_grpo_train.py}{SPG revision bc8f521c}.}
Their coding-data filtering remains unverified. This audit describes
the public loaders; it does not establish which rows were used for
published runs or quantify their effect on scores.

\clearpage
\section{Training histories}
\label{app:training}

\begin{figure}[!ht]
 \centering
 \includegraphics[width=\linewidth]{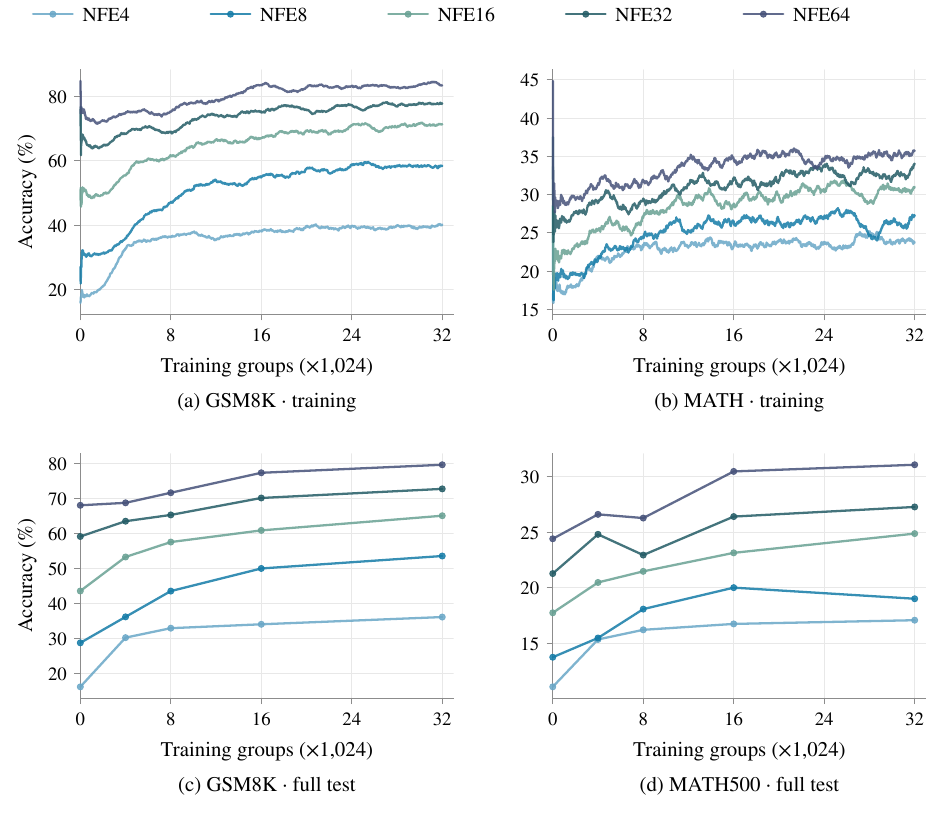}
 \caption{Math training and full-test histories. Each color denotes a policy trained at a different inference budget. The top panels show smoothed
 training accuracy; the bottom panels show test evaluations.}
 \label{fig:math-history}
\end{figure}

\begin{figure}[!ht]
 \centering
 \includegraphics[width=\linewidth]{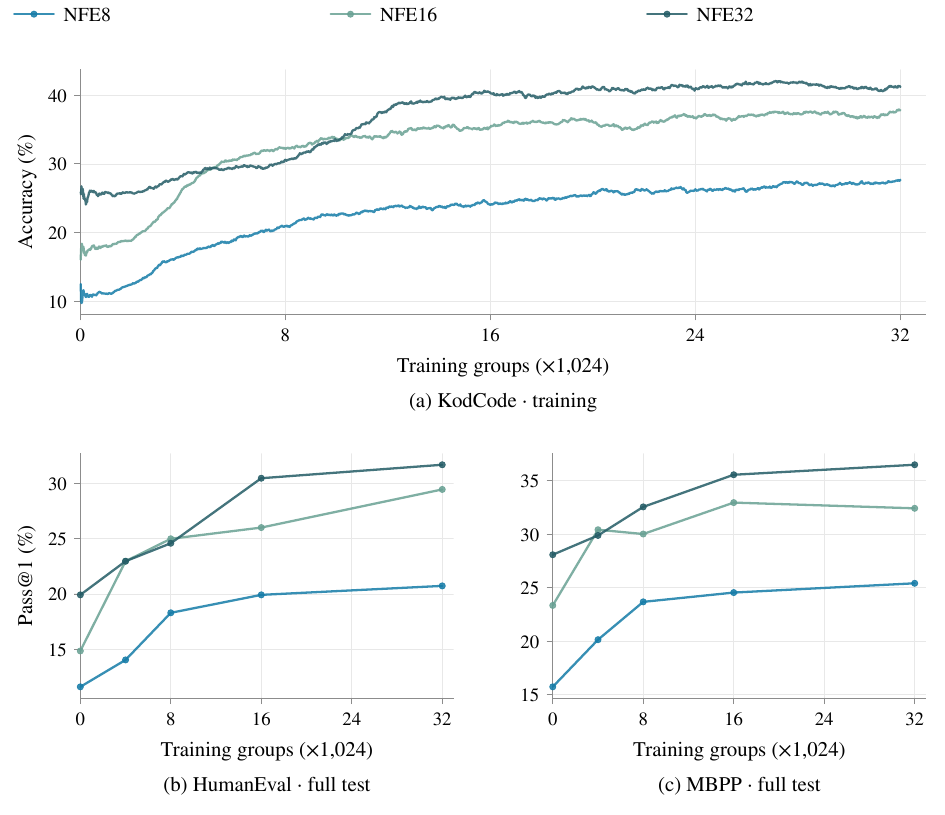}
 \caption{Code training and full-test histories for NFE8, NFE16, and
 NFE32. Main comparisons use each run's 32k-group checkpoint.}
 \label{fig:code-history}
\end{figure}

Figures~\ref{fig:math-history} and~\ref{fig:code-history} show training
accuracy and full-test evaluations through the 32k-group
endpoint. Training curves show trailing averages over up to 128 collections, weighted by the number of attempted trajectories and including trajectories from rejected groups.

\clearpage

\clearpage
\section{Ablation learning curves and remasking}
\label{app:ablations}

\begin{figure}[ht]
 \centering
 \includegraphics[width=\linewidth]{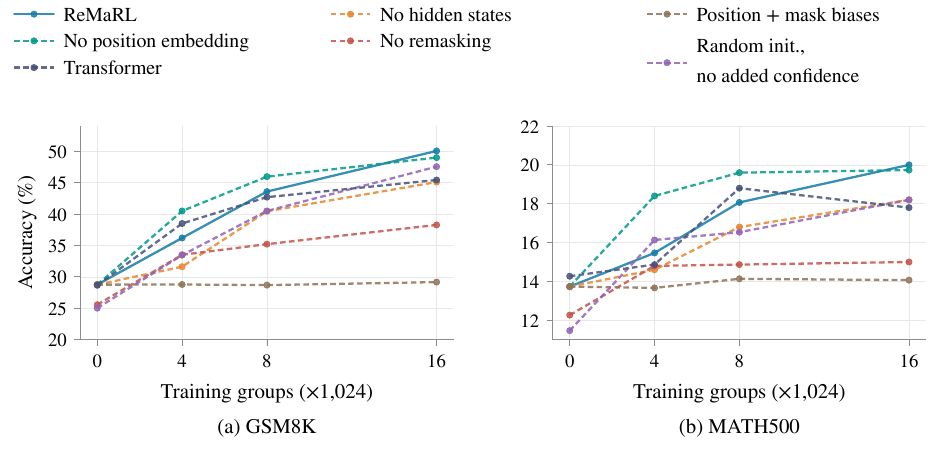}
 \caption{Full-test learning curves for the NFE8 ablation variants.}
 \label{fig:ablation-learning}
\end{figure}

Figure~\ref{fig:ablation-learning} follows the NFE8 ablations through
16k groups. The variant without added position embeddings leads the full policy early in
training, then falls slightly behind at the endpoint. Removing hidden
states slows learning and lowers final accuracy. Bias-only training
remains close to the untrained reference. The randomly initialized variant without the fixed confidence score improves substantially during training but finishes
below standard ReMaRL on both benchmarks. The transformer policy finishes below the MLP on both benchmarks
despite having roughly twice as many parameters.

\begin{figure}[ht]
 \centering
 \includegraphics[width=\linewidth]{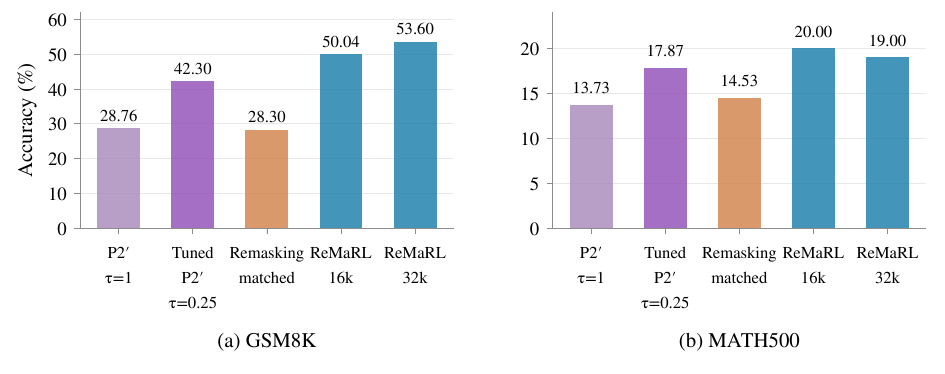}
 \caption{Full-test accuracy at NFE8 for ReMaRL and three P2$'$
settings: the default position temperature, a tuned position
temperature, and a retention bonus tuned to match the remasking frequency of the 32k-group ReMaRL policy.}
 \label{fig:revision-control}
\end{figure}

\begin{figure}[!ht]
 \centering
 \includegraphics[width=\linewidth]{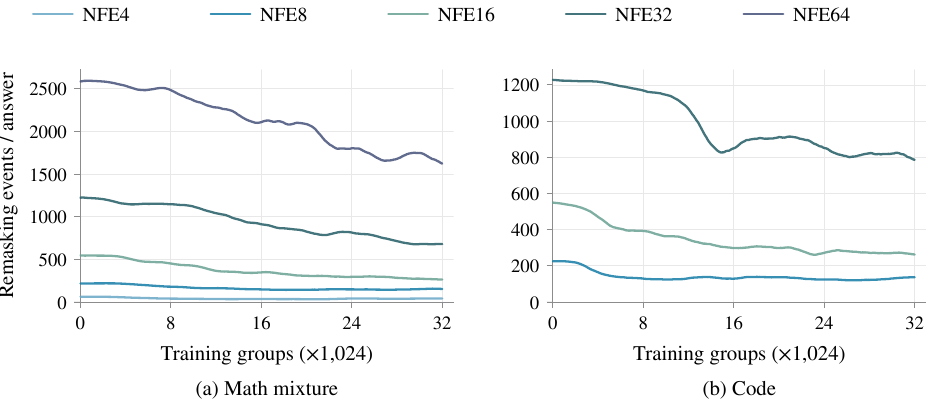}
 \caption{Remasking frequency through 32k training groups. Curves are
 trailing means over up to 128 collections, weighted by the number of attempted trajectories and including trajectories from rejected groups.}
 \label{fig:revision-dynamics}
\end{figure}

\clearpage
\paragraph{Remasking frequency.}
ReMaRL generally remasks fewer tokens as training progresses
(Figure~\ref{fig:revision-dynamics}). We count each time an
unmasked token is masked again, so a position can contribute
more than once per answer.

To test whether a retention bonus can recover ReMaRL's gains,
we added a fixed bonus to the scores of unmasked positions
in P2$'$ at $\tau=1$.
We tuned the bonus on 256 training questions to match the mean
number of remasking events per answer for the 32k-group NFE8
ReMaRL policy. ReMaRL remained more accurate on both GSM8K and
MATH500 (Figure~\ref{fig:revision-control}).

\begin{table}[t]
\centering
\small
\begin{tabular}{rc}
\toprule
Position temperature $\tau$ & GSM8K accuracy (\%) \\
\midrule
0    & 33.36 \\
0.10 & 34.47 \\
0.25 & 39.20 \\
0.50 & 48.19 \\
0.75 & 51.98 \\
0.85 & 53.53 \\
0.90 & 53.50 \\
0.95 & 54.38 \\
1.00 & 53.96 \\
1.05 & 53.47 \\
1.10 & 53.55 \\
1.15 & 53.17 \\
1.25 & 53.42 \\
1.50 & 51.58 \\
2.00 & 45.69 \\
\bottomrule
\end{tabular}
\caption{Independent position-temperature reevaluation of the fixed NFE8 ReMaRL policy trained for 32k groups. At $\tau=0$, position selection is greedy. Token predictions are greedy throughout.}
\label{tab:remarl-temperature-sensitivity}
\end{table}

\clearpage
\section{Data-size learning curves}
\label{app:data-size}

\begin{figure}[!ht]
 \centering
 \includegraphics[width=\linewidth]{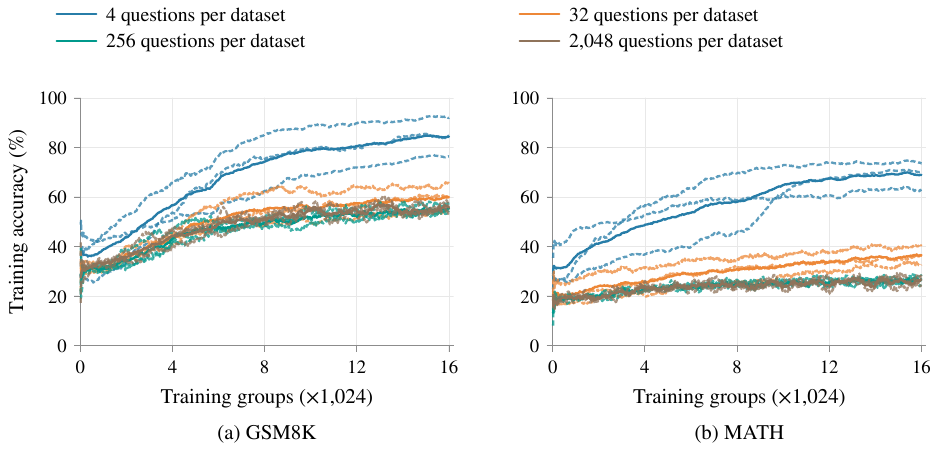}
 \caption{Training accuracy for the NFE8 data-size study. Each run uses
 16k accepted groups. Curves use 32-collection trailing averages, weighted by the number of attempted trajectories and including trajectories from rejected groups.}
 \label{fig:data-size-training}
\end{figure}

\begin{figure}[!ht]
 \centering
 \includegraphics[width=\linewidth]{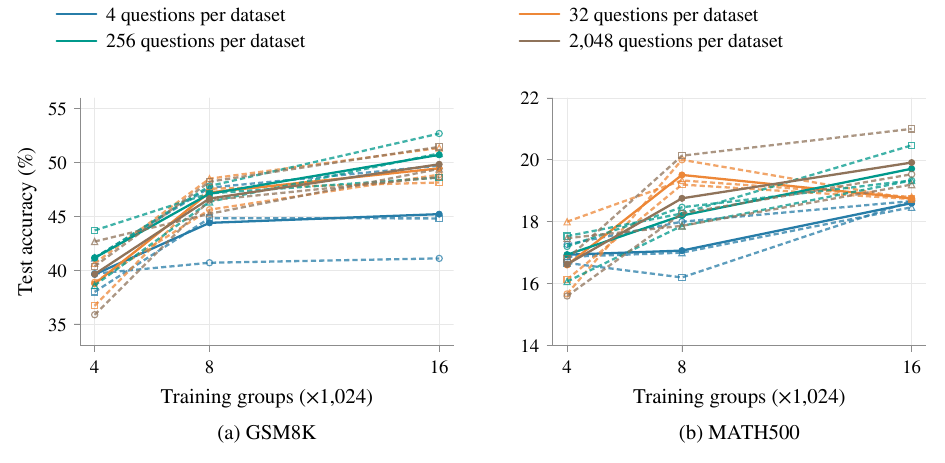}
 \caption{Full-test accuracy for the runs in Figure~\ref{fig:data-size-training}. For each training seed, smaller question subsets are contained in larger ones.}
 \label{fig:data-size-learning}
\end{figure}

Figures~\ref{fig:data-size-training} and~\ref{fig:data-size-learning}
follow all twelve data-size runs. A color identifies the number of
questions per dataset. Dashed lines show the three runs at each size; the solid
line shows their mean.
The smallest subsets attain high training accuracy quickly. On GSM8K,
their full-test learning curves show a large gap between seeds.

\clearpage
\section{Comparison with concurrent TIPS}
\label{app:tips}

\subsection{Policy formulation and training}

TIPS~\citep{tips} and ReMaRL independently learn to reveal,
retain, or remask positions using terminal correctness rewards and
a frozen diffusion model. TIPS uses a roughly 0.6M-parameter transformer that takes confidence, mask state, and the diffusion timestep as inputs. It samples independent
Bernoulli visibility decisions, with a shared logit shift calibrated
to an expected occupancy schedule. ReMaRL's roughly 0.3M-parameter policy projects the diffusion model's hidden states and scores each position with a shared MLP. It samples a fixed number of positions without replacement from a Plackett--Luce distribution.

TIPS's main recipe initializes RL with reference-supervised warmup. Both methods score unmasked positions
using the probability of the retained token. TIPS's policy
omits hidden states and can transfer to Dream-7B. Our zero-initialized
score correction starts from stochastic P2$'$. TIPS trains its main
policies at 32 calls; we train budget-specific policies and also
evaluate fixed-checkpoint transfer. Training compute differs across studies, as do the math answer checkers and the code evaluation pipelines.

\subsection{Math}

\begin{figure}[ht]
 \centering
 \includegraphics[width=\linewidth]{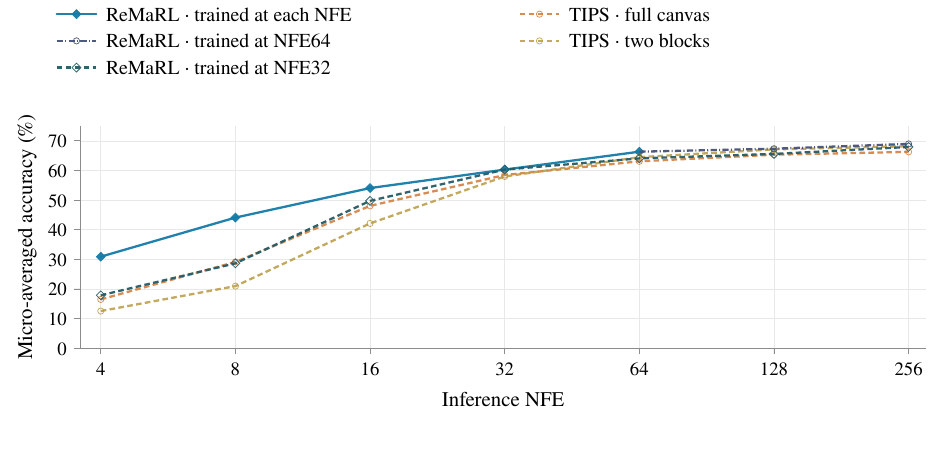}
 \caption{Math comparison with concurrent TIPS. Solid blue denotes
 separate policies trained at each budget; the dashed and dash-dot ReMaRL curves reuse policies trained at 32 and 64 calls, respectively, across inference budgets. Our checkpoints use 32k groups. TIPS values are
 reported in its Table 4.}
 \label{fig:tips-math}
\end{figure}

Both studies use frozen LLaDA-8B-Instruct, 256-token answers, the same
documented math prompt templates, and the full GSM8K and MATH500 test
sets. Figure~\ref{fig:tips-math} reports micro-averaged accuracy across the 1,319 GSM8K and 500 MATH500 questions.

At 4, 8, and 16 calls, our budget-specific policies score about 6--15 percentage points higher than the better TIPS variant on the combined math benchmark. Our fixed policy trained at 32 calls stays within two percentage points of the better TIPS variant from 8 to 256 inference calls.

\clearpage
\subsection{Code}

\begin{figure}[ht]
 \centering
 \includegraphics[width=\linewidth]{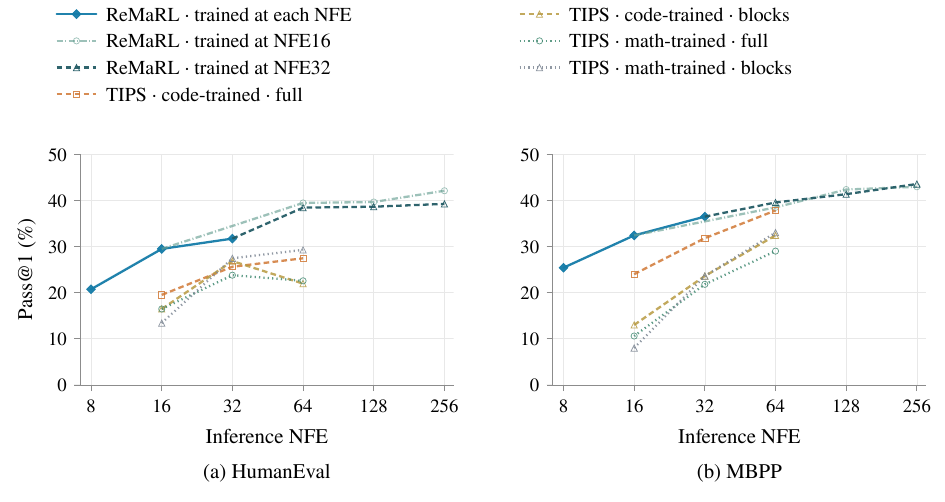}
 \caption{Code comparison with all four TIPS policy variants reported in Tables 24–25 of \citet{tips}. Solid blue denotes separate policies at their training
 NFE; dash-dot and dashed curves reuse the NFE16 and NFE32 checkpoints.
 Our policies use 32k groups.}
 \label{fig:tips-code}
\end{figure}

\begin{figure}[ht]
 \centering
 \includegraphics[width=\linewidth]{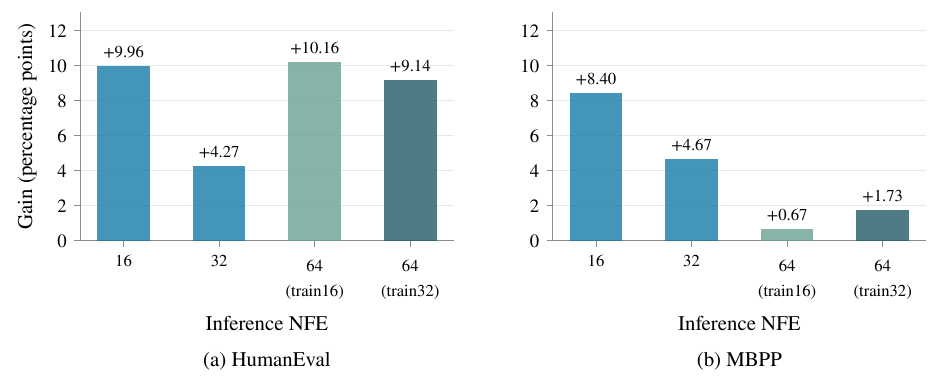}
 \caption{Gains over the highest-scoring TIPS policy variant at each
 budget and benchmark. The reference can select a different variant
 at each point. At 64 calls, both of our fixed-checkpoint results are
 shown separately.}
 \label{fig:tips-gaps}
\end{figure}

Figure~\ref{fig:tips-code} includes all four reported TIPS policy
variants: code-trained and math-to-code transfer, each with
full-canvas or two-block decoding. At 64 calls, their best reported scores are 29.27\% on HumanEval and 37.80\% on MBPP. When evaluated at 64 calls, our policy trained at 16 calls reaches 39.43\% and 38.47\%, respectively. Figure~\ref{fig:tips-gaps}
shows gaps to the best TIPS policy variant at each budget and
benchmark.

\clearpage
\section{Implementation and reproducibility}
\label{app:implementation}

\subsection{Code and checkpoints}

The \href{https://github.com/EgorShibaev/remarl}{ReMaRL repository}
contains the policy, GRPO trainer, inference code, benchmark evaluators,
and experiment configurations. Its command-line entry points are
\href{https://github.com/EgorShibaev/remarl/blob/main/docs/training.md}{\texttt{remarl train}}
for training,
\href{https://github.com/EgorShibaev/remarl/blob/main/docs/inference.md}{\texttt{remarl generate}}
for inference, and
\href{https://github.com/EgorShibaev/remarl/blob/main/docs/evaluation.md}{\texttt{remarl evaluate}}
and \texttt{remarl baseline} for policy and baseline evaluation.
The linked guides give complete commands; \texttt{configs/} supplies
recipes for math, code, and ablations.

The final LLaDA and Dream policies are available in the
\href{https://drive.google.com/drive/u/0/folders/1UFagh30vsvA-3jK-3sAf_iREaiQ3mxki}{public checkpoint folder}.
Running \texttt{remarl download checkpoints -{}-output checkpoints}
downloads them and verifies their checksums. The frozen LLaDA and Dream backbones are downloaded separately.

\subsection{Model, policy, and optimization}

We use \texttt{GSAI-ML/LLaDA-8B-Instruct}, snapshot
\texttt{08b83a6f}, for the LLaDA experiments. The backbone runs in
bfloat16 with bidirectional attention over the prompt and answer;
no key--value cache approximation is used. Vocabulary probabilities
are normalized in float32. The policy and its optimization use
float32. Hidden states are detached from the diffusion model before the
policy's LayerNorm and projection. The pointwise MLP has no dropout.
Except in the random-initialization ablation, the output layer and
skip weights start at zero. Added position embeddings use a
zero-mean normal initialization with standard deviation 0.02.

We use AdamW with a learning rate of $10^{-4}$, $(\beta_1,\beta_2)=(0.9,0.999)$,
an optimizer epsilon of $10^{-8}$, and zero weight decay. The learning
rate is constant. Group advantages use population standard deviation
and $\epsilon_A=10^{-8}$. Each update averages the clipped surrogate over all non-final
decisions in the batch. For each decision, we compute the log
probability of the sampled order. Gradient accumulation changes memory use while preserving normalization by the total number of non-final decisions in the batch. Behavior log
probabilities remain fixed during the four optimization passes.

Training uses $\tau=1$. Gumbel noise is generated from independent
uniform draws as $-\log(-\log u)$, clamping $u$ to a minimum of $10^{-10}$.
The implementation computes Plackett--Luce denominators with
\texttt{logcumsumexp}. The final fully unmasked step is excluded from
the objective. For a one-call decoder, only the final greedy
unmasking remains; our training budgets have at least two calls.

\subsection{Question sampling and training budget}

Main runs shuffle the retained training pool into epochs and avoid
repeating a question within a collection. A group is accepted only
if its eight rewards contain both successes and failures. Collection
continues until 16 groups are accepted. If 1,024 attempted groups do not fill the batch, collection aborts without a policy update. This filtering makes the effective training
distribution depend on the current policy's success probabilities.
The accepted-group budget therefore does not equal a fixed number
of generated answers or a fixed amount of computation.

For the data-size study, questions are shuffled within each dataset
and smaller subsets are prefixes of larger subsets for the same
seed. Once every question in a subset has been attempted within a
collection, the exclusion set is cleared. Repeated questions receive
fresh rollout noise and form distinct groups. This permits a
16-group collection even for the eight-question subset. The full-data
reference retains the main run's sampling procedure. Three seeds
jointly vary subset choice, policy initialization, and rollout noise.

Table~\ref{tab:training-costs} reports how many question groups each
main run sampled to collect 32k accepted groups. We report the
32k-group checkpoints for the main comparisons and the 16k-group
checkpoints for the ablations. The supplementary bundle includes
the sampling records for the ablation and data-size runs.

\begin{table}[!ht]
\centering\small
\caption{Question groups sampled to collect 32,768 accepted groups for each main policy.}
\label{tab:training-costs}
\begin{tabular}{@{}lrr@{}}
\toprule
Run & Attempted groups & Accepted (\%) \\
\midrule
Math, NFE4 & 74,845 & 43.8 \\
Math, NFE8 & 60,539 & 54.1 \\
Math, NFE16 & 61,974 & 52.9 \\
Math, NFE32 & 65,803 & 49.8 \\
Math, NFE64 & 75,748 & 43.3 \\
Code, NFE8 & 99,632 & 32.9 \\
Code, NFE16 & 84,817 & 38.6 \\
Code, NFE32 & 81,877 & 40.0 \\
\bottomrule
\end{tabular}
\end{table}

\Needspace{10\baselineskip}
\subsection{Ablation configurations}

The no-hidden-state ablation freezes a zero-valued hidden projection.
The no-position ablation freezes the added position embeddings at
zero; positional information can still be present in the diffusion model's hidden states. The bias-only correction is
$\Delta_i=b_i^{\mathrm{position}}+b_{m_i}^{\mathrm{mask}}$, added to
the same confidence score as the full policy.

For the no-remasking variant, let $V_{t-1}$ be the unmasked positions and
$M_{t-1}$ the masked positions before a call. All of $V_{t-1}$ are
retained, and the policy samples $k_t-|V_{t-1}|$ new positions from
$M_{t-1}$. Its Plackett--Luce likelihood normalizes only over the
remaining masked candidates. Deterministic retention contributes no term to the policy log-likelihood. The final call fills the remaining masks. 

For the random-initialization ablation, we remove the fixed
confidence addition and use
$z_i^t=\mathrm{MLP}_\phi(f_i^t)+w_{\mathrm{skip}}^\top f_i^t$.
The MLP output layer, including its bias, and the linear skip
use PyTorch's default linear-layer initialization. All input
features, including confidence, are retained.

For the transformer ablation, we reuse the RoPE DiT block from
the implementation of~\citet{apple}. We use one block with
hidden width 128, two attention heads, a 512-dimensional
feed-forward layer, GELU activations, and no dropout.
The block attends across answer positions and is conditioned
on each position's mask state. We retain ReMaRL's input features,
fixed confidence score, and linear skip connection. The output
projection, skip weights, and adaptive-conditioning projection
are initialized to zero. The resulting policy has approximately
0.58M trainable parameters, compared with 0.28M for the MLP.

\subsection{Data, evaluation, and supplementary records}

Math uses the zero-shot, task-specific prompts, reasoning
suffix, and answer checkers of \citet{apple}. The pool contains 7,473 GSM8K and 7,500
MATH training questions. Code uses the pinned KodCode-Light-RL-10K
snapshot and the task-specific prompts, sanitization, and tests of \citet{apple}.
Prompts are not truncated, and the answer canvas has 256 tokens.
We retain all test questions: 1,319 from GSM8K, 500 from MATH500, 164 from HumanEval, and 500 from MBPP. Code execution runs in an isolated
sandbox; infrastructure failures abort evaluation rather than
receiving an incorrect-answer reward. For confidence decoding, random unmasking, and Fast-dLLM, we use the implementations of \citet{apple}.

The code filter requires executable tests that the reference
solution passes and screens training problems for benchmark overlap. The supplementary manifest
lists retained and excluded source IDs, exclusion reasons, source
revisions, and hashes. It also records which training problems were quarantined because of runtime-compatibility issues. Appendix~\ref{app:data} describes
the overlap screen. Filtering scripts and the original per-run
cohort manifests preserve those distinctions.

The \href{https://drive.google.com/file/d/1Gh0MaNUTpl5gIWBNtsHfYodrOaYQV86O/view?usp=sharing}{supplementary bundle}
includes training and evaluation code, policy configurations, and
manifests for dependencies and model files. It provides the per-question
outcomes used for paired comparisons, records the evaluated operating
points, and identifies scores taken from prior work. Python scripts
reproduce the tables, figures, and paired intervals.

\clearpage
\section{Paired evaluation uncertainty}
\label{app:uncertainty}

We estimate paired uncertainty from saved binary outcomes for the
reported checkpoints. For each question and method, we average the
binary outcomes across three inference seeds. We then subtract the
comparison method's mean from the ReMaRL mean for that question.
We resample questions with replacement 20,000 times and report the
2.5th and 97.5th percentiles of the mean difference. All seed outcomes
for a question remain together in each resample. Question identities,
prompts, reference answers, and recorded totals are checked before
pairing.

We also report the equally weighted average difference across the two
benchmarks in each domain. For its interval, we independently resample
questions within each benchmark and average the two benchmark
differences in each bootstrap replicate.
Table~\ref{tab:paired-uncertainty} reports pointwise 95\%
question-bootstrap intervals conditional on the trained checkpoints.

The average gains over both default and calibrated P2$^{\prime}$
have intervals entirely above zero in math and code. Among the math
component ablations, retaining remasking and hidden-state inputs
improves average accuracy by 8.39 and 3.38 percentage points,
respectively, with both intervals entirely above zero.

\begin{table}[!ht]
\centering\small
\setlength{\tabcolsep}{4pt}
\renewcommand{\arraystretch}{1.15}
\caption{Paired score differences in percentage points, with 95\%
bootstrap intervals in brackets. Average gives equal weight
to the two benchmarks within each domain.}
\label{tab:paired-uncertainty}
\begin{tabular*}{\linewidth}{@{\extracolsep{\fill}}lrrr@{}}
\toprule
Comparison (math, NFE8) & GSM8K & MATH500 & Average\\
\midrule
ReMaRL (32k) vs. P2$^{\prime}$, $\tau=1$
 & +24.84 [22.87, 26.81]
 & +5.27 [3.00, 7.60]
 & +15.05 [13.55, 16.59]\\
ReMaRL (32k) vs. calibrated P2$^{\prime}$
 & +11.30 [9.48, 13.12]
 & +1.13 [-0.93, 3.27]
 & +6.21 [4.88, 7.61]\\
ReMaRL (16k) vs. calibrated P2$^{\prime}$
 & +7.73 [5.86, 9.60]
 & +2.13 [0.00, 4.33]
 & +4.93 [3.53, 6.35]\\
ReMaRL (32k) vs. remasking-matched P2$^{\prime}$
 & +25.30 [23.35, 27.27]
 & +4.47 [2.33, 6.60]
 & +14.88 [13.44, 16.33]\\
ReMaRL (16k) vs. no remasking
 & +11.78 [9.98, 13.57]
 & +5.00 [2.80, 7.20]
 & +8.39 [6.96, 9.83]\\
ReMaRL (16k) vs. no hidden states
 & +4.95 [3.21, 6.72]
 & +1.80 [-0.07, 3.73]
 & +3.38 [2.08, 4.66]\\
ReMaRL (16k) vs. no added position embeddings
 & +1.06 [-0.63, 2.73]
 & +0.27 [-1.67, 2.20]
 & +0.66 [-0.62, 1.93]\\
ReMaRL (16k) vs. position and mask biases
 & +20.87 [18.95, 22.77]
 & +5.93 [3.53, 8.33]
 & +13.40 [11.86, 14.93]\\
\midrule
Comparison (code, NFE16) & HumanEval & MBPP & Average\\
\midrule
ReMaRL (32k) vs. P2$^{\prime}$, $\tau=1$
 & +14.63 [10.16, 19.51]
 & +9.07 [6.40, 11.80]
 & +11.85 [9.19, 14.62]\\
ReMaRL (32k) vs. calibrated P2$^{\prime}$
& +6.91 [2.44, 11.38]
& +3.07 [0.47, 5.67]
& +4.99 [2.40, 7.64] \\
\bottomrule
\end{tabular*}
\end{table}
\end{document}